\documentclass[11pt,twoside]{article}
\usepackage{fullpage}

\usepackage{epsf}
\usepackage{fancyhdr}
\usepackage{graphics}
\usepackage{graphicx}
\usepackage{psfrag}
\usepackage{microtype}
\usepackage{subfigure}
\usepackage{algorithmic}
\usepackage{color,xcolor}
\usepackage{caption}
\usepackage{subcaption}
\usepackage{subfigure}

\usepackage[linesnumbered,ruled]{algorithm2e}
\DontPrintSemicolon

\usepackage{color}

\usepackage{amsthm}
\usepackage{amsfonts}
\usepackage{amsmath}
\usepackage{amssymb,bbm}

\usepackage{mathrsfs}

\usepackage{url}
\usepackage[colorlinks,linkcolor=magenta,citecolor=blue, pagebackref=true,backref=true]{hyperref}
\renewcommand*{\backrefalt}[4]{%
    \ifcase #1 \footnotesize{(Not cited.)}%
    \or        \footnotesize{(Cited on page~#2.)}%
    \else      \footnotesize{(Cited on pages~#2.)}%
    \fi}

\usepackage{nicefrac}
\usepackage{comment}

\usepackage{chngpage}

 \usepackage{tabularx}%
\usepackage{enumitem}
\usepackage{booktabs}
\usepackage{caption}

\usepackage{bm,bbm}
\usepackage{mathtools}
\usepackage{cleveref}

\usepackage{eucal}

\usepackage{scalerel}

\usepackage{final_macros}

\newcommand{\lpspace}[1]{\mathbb{L}^{#1}}

\newcommand{\state}{x}

\newcommand{\action}{a}

\newcommand{\MyState}{X}
\newcommand{\State}{\MyState}

\newcommand{\Action}{A}

\newcommand{\actionspace}{\mathbb{A}}

\newcommand{\policy}{\pi}

\newcommand{\Delhat}{\widehat{\Delta}}

\newcommand{\dudley}{\mathcal{J}}

\newtheorem{assumption}{Assumption}
\newcommand{\numobs}{\ensuremath{n}}

\newcommand{\usedim}{\ensuremath{d}}

\newcommand{\reward}{r}

\newcommand{\valueaux}{{\valuefunc}_\stepsize^*}

\newcommand{\projectto}[1]{\Pi_{#1}}

\newcommand{\ltwospace}{\mathbb{L}^2}
\newcommand{\statespace}{\mathbb{X}}

\newcommand{\randreward}{R}

\newcommand{\behavpolicy}{{\pi_0}}
\newcommand{\initDistr}{\rho_0}

\newcommand{\Ball}{\ensuremath{\mathbb{B}}}

\newcommand{\stationary}{\rho}
\newcommand{\Stationary}{\nu}

\newcommand{\Statnorm}[1]{\vecnorm{#1}{\Stationary}}

\newcommand{\Qfunc}{Q}
\newcommand{\fakerefassumelip}[1]{\hyperref[assume:smooth-high-order]{{\color{magenta} {\upshape\textbf (}{\upshape{\textbf{Lip}}#1}{\upshape\textbf )}}} }

\newcommand{\discount}{\beta}

\newcommand{\valuefunc}{v}

\newcommand{\soboinprod}[2]{\inprod{#1}{#2}_{\mathbb{H}^1}}

\newcommand{\ValueFunc}{\valuefunc}

\newcommand{\ValFun}{\valuefunc}
\newcommand{\ValFunc}{\valuefunc}
\newcommand{\valfun}{\valuefunc}

\DeclareFontFamily{U}{mathx}{}
\DeclareFontShape{U}{mathx}{m}{n}{<-> mathx10}{}
\DeclareSymbolFont{mathx}{U}{mathx}{m}{n}

\DeclareMathAccent{\widecheck}{0}{mathx}{"71}
\newcommand{\BellOp}{\ensuremath{\mathcal{T}}}

\newcommand{\valuebar}{\widebar{\valuefunc}}

\newcommand{\valuehat}{\widehat{\valuefunc}}

\newcommand{\lammin}{\lambda_{\min}}
\newcommand{\lammax}{\lambda_{\max}}

\newcommand{\Event}{\mathcal{E}}
\newcommand{\Term}{T}

\usepackage{mathbbol}

\long\def\comment#1{}

\newcommand{\funcClass}{\mathcal{C}}

\newcommand{\learningrate}{\alpha}

\newcommand{\statnorm}[1]{\|#1\|_{\stationary}}
\newcommand{\statinprod}[2]{\inprod{#1}{#2}_{\stationary}}
\newcommand{\Statinprod}[2]{\inprod{#1}{#2}_{\Stationary}}

\newcommand{\ValPol}{\ensuremath{\ValFun^\star}}
\newcommand{\ValTrue}{\ValPol}

\newcommand{\ValBar}{\ensuremath{\bar{\ValFun}}}

\newcommand{\empsoboinprod}[2]{\widehat{\mathcal{E}}_\numobs \big(#1, #2 \big)}

\newcommand{\soboone}{{\mathbb{H}^1 (\stationary)}}

\newcommand{\Fclass}{\ensuremath{\mathcal{F}}}

\newcommand{\cmalliavin}{c_{\mathrm{Mall}}}

\newcommand{\ctwofour}{c_{24}}

\newcommand{\ErrTermSq}{Z^{\mathrm{(sq)}}}
\newcommand{\ErrTermMain}{Z^{\mathrm{(Bel)}}}
\newcommand{\ErrTermSqVal}{\ErrTermSq_{\valfun}}
\newcommand{\ErrTermSqAdv}{\ErrTermSq_{\advFunc}}
\newcommand{\ErrTermMainVal}{\ErrTermMain_{\valfun}}
\newcommand{\ErrTermMainAdv}{\ErrTermMain_{\advFunc}}

\newcommand{\eucnorm}{\abss}

\usepackage[framemethod=TikZ]{mdframed}

\newcommand{\valproj}{\widetilde{\valfun}}
\newcommand{\advfuncproj}{\widetilde{\advFunc}}

\newcommand{\diameter}{\mathrm{diam}}

\newenvironment{narrowpara}
  {\par\addvspace{\smallskipamount}\narrower\noindent\ignorespaces}
  {\par\addvspace{\smallskipamount}}

\newcommand{\Tmax}{T_{\max}}

\newcommand{\LinSpace}{\mathbb{K}}

\newcommand{\IdMat}{\mathcal{I}}

\newcommand{\Dset}{\mathcal{D}}
\newcommand{\creg}{c_{\mathrm{reg}}}

\newcommand{\valfuncspace}{\Fclass_\ValFunc}

\newcommand{\drift}{b}
\newcommand{\covMat}{\Lambda}

\newcommand{\BM}{B}

\newcommand{\lpnorm}[3]{\|#1\|_{\mathbb{L}^{#2}(#3)}}
\newcommand{\lpstatnorm}[2]{\|#1\|_{\mathbb{L}^{#2}(\stationary)}}
\newcommand{\orlicznorm}[2]{\|#1\|_{\psi_{#2}}}

\crefname{assumption}{assumption}{assumptions}

\newcommand{\constFclass}{c_{\Fclass}}

\newcommand{\advFuncBar}{\widebar{\advFunc}}

\newcommand{\traj}{\tau}

\newcommand{\marginal}{\mu}

\newcommand{\semigroup}{\mathcal{P}}
\newcommand{\generator}{\mathcal{A}}

\newcommand{\constbd}{B}

\newcommand{\sobonorm}[1]{\vecnorm{#1}{\mathbb{H}^1 (\stationary)}}
\newcommand{\advFunc}{q}

\newcommand{\advFuncaux}{{\advFunc}^*_\stepsize}
\newcommand{\advFuncHat}{\widehat{\advFunc}}

\newcommand{\constcmp}{c_{\actionspace}}
\newcommand{\empbelladv}{\widehat{\mathcal{T}}_\numobs^{(q)}}

\newcommand{\hjboperator}{\mathcal{T}}

\newcommand{\obstransition}{\widebar{\semigroup}_\stepsize^{\behavpolicy}}
\newcommand{\Clin}[1]{C_{\mathrm{lin}}^{#1}}
\newcommand{\couyang}{c_{\mathrm{discr}}}

\newcommand{\empbilinear}[3]{\widehat{\mathcal{B}}_\numobs \big(#1, #2; #3 \big)}

\begin{document}

\begin{center}
{\bf{\LARGE{Continuous-time reinforcement learning: ellipticity enables model-free value function approximation}}}

\vspace*{.2in}

{\large{
 \begin{tabular}{cc}
  Wenlong Mou$^{ \diamond}$ 
 \end{tabular}

}

\vspace*{.2in}

 \begin{tabular}{c}
 Department of Statistical Sciences, University of Toronto$^{\diamond}$
 \end{tabular}
}

\begin{abstract}
We study off-policy reinforcement learning for controlling continuous-time Markov diffusion processes with discrete-time observations and actions. We consider model-free algorithms with function approximation that learn value and advantage functions directly from data, without unrealistic structural assumptions on the dynamics.

Leveraging the ellipticity of the diffusions, we establish a new class of Hilbert-space positive definiteness and boundedness properties for the Bellman operators. Based on these properties, we propose the Sobolev-prox fitted $q$-learning algorithm, which learns value and advantage functions by iteratively solving least-squares regression problems. We derive oracle inequalities for the estimation error, governed by (i) the best approximation error of the function classes, (ii) their localized complexity, (iii) exponentially decaying optimization error, and (iv) numerical discretization error. These results identify ellipticity as a key structural property that renders reinforcement learning with function approximation for Markov diffusions no harder than supervised learning.
\end{abstract}
\end{center}

\section{Introduction}\label{sec:intro}
A long-standing open question in reinforcement learning (RL) theory is whether RL with function approximation is harder than supervised learning --- it is well-known that supervised learning methods enjoy distribution-free theoretical guarantees such as PAC learning and oracle inequalities (see e.g.~\cite{bartlett2005local,koltchinskii2011oracle}), with the prediction error depending on the approximation and intrinsic complexities of the function class. This raises the fundamental question of whether comparable, model-free guarantees are possible for reinforcement learning with function approximation. In particular, by using offline data generated from a behavior policy, and by employing
 a function class $\Fclass$ to approximate the optimal value function, can we learn the optimal value function efficiently without requiring a model of the environment?

The answer is negative in general. Standard value-based RL algorithms such as fitted Q-iteration and temporal difference learning can be unstable with function approximation, even in linear settings~\cite{tsitsiklis1996feature,baird1995residual,wang2021instabilities}. Recent works~\cite{wang2020statistical,foster2021offline} have also established information-theoretic lower bounds, showing the impossibility of model-free learning with function approximation without further assumptions. These negative results suggest that RL with function approximation is strictly harder than supervised learning in general.

Mathematically, the difficulty of model-free RL with function approximation stems from a basic fact: the Bellman operator is a contraction only in the supremum norm, but not in the $\ltwospace$-norm induced by the data distribution. This mismatch between the geometry of the problem lies at the heart of the instability of model-free RL with function approximation. As a result, Bellman backup operations can amplify the statistical error in an uncontrollable manner. To resolve this issue, existing literature either impose additional structural assumptions like Bellman completeness~\cite{munos2008finite,xie2021bellman}, or require strong data coverage conditions~\cite{xie2021batch,zhan2022offline}. Despite their popularity in theory, these assumptions are often difficult to verify or enforce in practice, and the resulting algorithms can often be computationally inefficient. This motivates a central question in RL theory:
\begin{quote}
\emph{ What mathematical structures enable model-free RL with function approximation?}
\end{quote}
In this paper, we provide an alternative and more realistic answer to this question, by focusing on continuous-time processes that often arise in engineering applications. Concretely, we consider a controlled diffusion process
\begin{align}
  d \MyState_t = \drift^{A_t} (\MyState_t) dt + \covMat (\MyState_t)^{1/2} d \BM_t,\label{eq:cts-time-process}
\end{align}
where $\MyState_t \in \statespace = \real^\usedim$ is the state variable, $\Action_t \in \actionspace$ is the control variable, and $\{\BM_t\}_{t \geq 0}$ is a standard $\usedim$-dimensional Brownian motion. The drift function $\drift: \statespace \times \actionspace \rightarrow \real^\usedim$ and the diffusion matrix function $\covMat: \statespace \rightarrow \real^{\usedim \times \usedim}$ are unknown to the learner. We focus on an infinite-horizon discounted reward setting, where the goal is to find a control policy $\policy$ that maximizes the expected cumulative reward.
\begin{align*}
  \Exs^\policy \Big[ \int_0^{\infty} e^{- \discount t} \reward (\MyState_t, \Action_t) dt \Big].
\end{align*}
Here $\discount > 0$ is the discount rate, and $\reward: \statespace \times \actionspace \rightarrow \real$ is the reward function. This problem formulation finds natural applications in finance~\cite{dai2025data,wang2020continuous}, robotics~\cite{theodorou2010generalized,recht2019tour}, queuing systems~\cite{ata2025drift}, and reward-guided fine-tuning of diffusion generative models~\cite{han2024stochastic,mou2025rl,gao2024reward}, among others.

Despite the continuous-time formulation, in a realistic setting, the states and rewards can only be observed at discrete time steps, and the actions can only change at these discrete time steps. By observing the process~\eqref{eq:cts-time-process} at discrete time steps with step size $\stepsize > 0$, the problem can be cast as a discrete-time Markov decision process, and the optimal policy, optimal value function, $Q$-function, and advantage functions are well-defined (see Section~\ref{sec:setup} for details). In addition to aforementioned difficulties for RL with function approximation, this MDP also suffers from additional issues: (1) the effective horizon of the discrete-time MDP scales as $\stepsize^{-1}$, which can be very large when $\stepsize$ is small; (2) the fluctuations of the Brownian motion introduces large variances in the observations, which can further amplify the statistical error. These challenges have been addressed in the context of policy evaluation~\cite{mou2024bellman,mou2025statistical} using a similar framework, and we extend these results to the control setting in this paper.

\paragraph{Contributions:} In this paper, we solve the discrete-time RL problem induced by the continuous-time diffusion process~\eqref{eq:cts-time-process} under mild assumptions with value function approximations. Assuming access to offline data generated from a behavior policy, we seek to learn the optimal value function and optimal advantage function within pre-specified function classes $\Fclass_\valfun$ and $\Fclass_\advFunc$, respectively. Our main contributions are as follows:
\begin{itemize}
\item Assuming the diffusion matrix function $\covMat$ is uniformly elliptic (i.e., it has eigenvalues bounded away from zero) along with mild regularity conditions, we establish monotonicity and boundedness properties for a class of operators induced by the Bellman operators in appropriate Sobolev spaces. These properties guarantee the approximation error of the projected Bellman solution can be controlled by the best approximation error in the function classes $\Fclass_\valfun$ and $\Fclass_\advFunc$.
\item Based on these properties, we propose the Sobolev-prox fitted $q$-learning algorithm. The algorithm updates the advantage function estimate by solving a least-squares regression problem in the function class $\Fclass_\advFunc$, followed by a value function update via a proximal step under Sobolev norm in the function class $\Fclass_\valfun$.
\item We derive oracle inequalities for the output of the Sobolev-prox fitted $q$-learning algorithm, showing that the estimation error is governed by (i) the best approximation errors of the function classes $\Fclass_\valfun$ and $\Fclass_\advFunc$, measured under same norms (ii) critical radii defined by localized complexities of these classes, (iii) exponentially decaying optimization error, and (iv) numerical discretization error that scales as $\stepsize^{1/2}$. Both the statistical guarantees and the computational complexity of the algorithm match those of regression problems in the respective function classes, up to logarithmic factors.
\end{itemize}
The rest of the paper is organized as follows. We first introduce notations and discuss related work. In \Cref{sec:setup}, we formalize the continuous-time and discrete-time RL problems. In \Cref{sec:algorithm}, we derive the Sobolev-prox fitted $q$-learning algorithm from projected Bellman equations. In \Cref{sec:theory}, we present the main theoretical results, followed by proofs in \Cref{sec:proofs}. We conclude the paper with a discussion in \Cref{sec:discussion}.

\paragraph{Notation:} For symmetric matrices $A, B \in \real^{\usedim \times \usedim}$, we write $A \preceq B$ if $B - A$ is positive semidefinite. For scalars, vectors, and matrices we use $\abss{\cdot}$ to denote the absolute value, Euclidean norm, and operator norm, respectively. We use $\mathcal{P} (S)$ to denote the space of probability measures on a set $S$.
For a multi-index $\alpha = (\alpha_1, \alpha_2, \ldots, \alpha_\usedim) \in \mathbb{N}^\usedim$, we define its order as $|\alpha| = \sum_{i=1}^\usedim \alpha_i$. For a smooth function $f: \real^\usedim \rightarrow \real$, we use the notation $\partial^\alpha f$ to denote the mixed partial derivative $\frac{\partial^{|\alpha|} f}{\partial x_1^{\alpha_1} \partial x_2^{\alpha_2} \cdots \partial x_\usedim^{\alpha_\usedim}}$. Given a function $f: \real^\usedim \rightarrow \real$, we define the norms
\begin{align*}
  \vecnorm{f}{C^k} \mydefn \sum_{0 \leq |\alpha| \leq k} \sup_{x \in \real^\usedim} \eucnorm{\partial^\alpha f (x)}, \quad \mbox{and} \quad
  \vecnorm{f}{\Clin{k}} \mydefn \sum_{1 \leq |\alpha| \leq k} \sup_{x \in \real^\usedim} \eucnorm{\partial^\alpha f (x)} + \sup_{x \in \real^\usedim} \frac{\abss{f (x)}}{1 + \eucnorm{x}},
\end{align*}
where the latter norm $\Clin{k}$ allows linear growth of the function $f$. Given a probability measure $\rho$, we use $\lpstatnorm{f}{p}$ to denote the $\mathbb{L}^p (\rho)$-norm of $f$, defined as $\lpstatnorm{f}{p} \mydefn \big( \int_{\real^\usedim} \abss{f (x)}^p d \rho (x) \big)^{1/p}$ for $1 \leq p < \infty$, and $\lpstatnorm{f}{\infty} \mydefn \inf \{ C \geq 0: \mu (\{ x: \abss{f (x)} > C \}) = 0 \}$. For $\ltwospace$ norms, we use the shorthand notation $\statnorm{f} \mydefn \lpstatnorm{f}{2}$ and $\statinprod{f}{g} \mydefn \int_{\real^\usedim} f (x) g (x) d \rho (x)$. When the measure $\rho$ is defined on subset of $\real^\usedim$, we further define the Sobolev norm
\begin{align*}
  \sobonorm{f}^2 \mydefn \statnorm{f}^2 + \statnorm{\nabla f}^2 = \int_{\real^\usedim} f (x)^2 + \eucnorm{\nabla f (x)}^2 d \rho (x).
\end{align*}
Given a metric space $(T, d)$, we use $\Ball (x; r) \mydefn \{ y \in T: d (x, y) \leq r \}$ to denote the closed ball of radius $r$ centered at $x \in T$, and we use $\Ball (r)$ to denote the closed ball centered at the origin. We use $\diameter_d (T)$ to denote the diameter of $T$ under metric $d$. Given a subset $S \subseteq T$, we define the projection operator
\begin{align*}
  \projectto{S}^d (x) \mydefn \arg\min_{y \in S} d (x, y).
\end{align*}
Note that the projection operator may not be unique in general, and we assume that a measurable selection is chosen whenever necessary.

For $q \geq 1$, we use use $\orlicznorm{X}{q}$ to denote the Orlicz norm of a random variable $X$, defined as $\orlicznorm{X}{q} \mydefn \inf \{ t > 0: \Exs [\exp (|X|^q / t^q)] \leq 2 \}$. We use $N(\epsilon; \Fclass, d)$ to denote the covering number of a function class $\Fclass$ under metric $d$ at scale $\epsilon > 0$, and we define the Dudley integral as
\begin{align*}
  \dudley_q (\Fclass, d) \mydefn \int_0^{\diameter_d (\Fclass)} \big( \log N (\epsilon; \Fclass, d) \big)^{1/q} d \epsilon.
\end{align*}
We use $(B_t: t \geq 0)$ to denote a standard Brownian motion in $\real^\usedim$.

\subsection{Related work}\label{subsec:related-work}
Since this work involves RL with function approximation, stochastic control, and statistical learning for PDEs, we briefly review related literature in these areas.

\paragraph{Continuous-time RL:} Reinforcement learning in continuous-time Markov diffusions has been studied since the foundational work of \cite{munos1997reinforcement}. More recently, a growing body of work~\cite{wang2020reinforcement,jia2022policyeval,jia2022policygrad,jia2023q} have developed model-free RL algorithms for controlled Markov diffusions based on martingale orthogonality conditions. The advantage function formulation in our paper is inspired by their $q$-learning algorithm~\cite{jia2023q}. Due to its practical significance, continuous-time RL has attracted recent research attention, with many works extending and analyzing these algorithms~\cite{zhao2023policy,jiang2024robust,wang2023reinforcement}. In particular, motivated by recent development of diffusion generative models, several recent works~\cite{han2024stochastic,mou2025rl,gao2024reward} have also studied RL for reward-guided fine-tuning of diffusion models.

Convergence and sample complexity guarantees for continuous-time RL algorithms are relatively scarce. Existing work either relies on model-based assumptions like linear-quadratic systems~\cite{basei2022logarithmic}, or focus on population-level analysis without statistical guarantees~\cite{reisinger2023linear,tang2024regret}. The sequence of work~\cite{mou2024bellman,mou2025statistical,mou2025rl} is closely related to our paper. They establish oracle inequalities for policy evaluation, as well as for control-affine systems with entropy regularization. Our work extends these results to general control problems without structural assumptions on the dynamics, and develops new algorithmic and analytical techniques.

\paragraph{RL with function approximation:} The past decade has witnessed significant progress in understanding RL with function approximation. In addition to aforementioned results, several lines of work attempt to bridge the gap between RL and supervised learning. \cite{chen2022finite,chen2023target,meyn2024projected} study stochastic approximation schemes for solving projected Bellman equations with linear function approximation, showing stability and convergence under certain assumptions. But they do not provide statistical guarantees. By introducing additional approximation function classes, many recent work circumvented the negative results for model-free RL with function approximation. \cite{zanette2022bellman} considers orthogonality from projected Bellman equations, and establishes statistical guarantees by leveraging value function and test function approximations for any policy. \cite{chen2022offline,ozdaglar2023revisiting} uses linear programming formulation and density ratio modelling to achieve learning guarantees only under realizability. In contrast, we focus on the value learning settings without requiring additional function approximations.

\paragraph{Learning for elliptic PDEs:} The value function of the continuous-time control problem satisfies an associated Hamilton--Jacobi--Bellman (HJB) equation, which is a second-order nonlinear elliptic PDE. While we do not directly solve the HJB equation in this paper, our analysis relies on the elliptic structures. There is a growing literature on statistical learning theory for elliptic and parabolic PDEs~\cite{lu2021machine,lu2021priori,yu2018deep}. For HJB equations, several RL-based algorithms have been proposed~\cite{zhou2021actor,zhu2025optimal}. Our approach is different from these works, as we focus on model-free RL algorithms that learn value and advantage functions directly from data, without knowing or learning the dynamics.

\section{Problem setup}\label{sec:setup}
Let us now formalize control problems in this section. We first introduce the MDP induced by the continuous-time controlled diffusion, followed by a description of the observation model.

\subsection{Control protocol and MDP formulation}\label{subsec:mdp-setup}
Given a stepsize $\stepsize > 0$, we assume that the state and reward can only be observed at discrete time steps $0, \stepsize, 2 \stepsize, \ldots$, and the action can only change at these discrete time steps. In particular, given an action space $\actionspace$, the agent can take action $\Action_{k \stepsize} = a \in \actionspace$ at time step $k \stepsize$, and the process evolves as
\begin{align}
  d \MyState_t = \drift^\action (\MyState_t) dt + \covMat (\MyState_t)^{1/2} d \BM_t, \quad \mbox{for } t \in [k \stepsize, (k+1) \stepsize).\label{eq:cts-time-process-action}
\end{align}
 The diffusion process~\eqref{eq:cts-time-process-action} defines a semigroup $(\semigroup_t^\action: t \geq 0)$, such that $\semigroup_t^\action f (x) \mydefn \Exs \big[ f(\MyState_t^\action) \mid \MyState_0^\action = x \big]$, where the generator is given by
\begin{align*}
  \generator^\action: f \mapsto \big(\drift^\action \big)^\top \nabla f + \frac{1}{2} \mathrm{Tr} \big( \covMat \cdot \nabla^2 f \big).
\end{align*}
We also define the adjoint operators $( (\semigroup_t^{\action})^*: t \geq 0)$ and $(\generator^\action)^*$ that acts on measures.

Given a bivariate reward function $\reward: \statespace \times \actionspace \rightarrow \real$, the discounted reward received within time interval $[k \stepsize, (k+1) \stepsize)$ is given by $\stepsize \reward (\MyState_{k \stepsize}, \Action_{k \stepsize})$.\footnote{Here we assume that the reward observed at discrete time points are the actual received reward. Alternatively, one can consider the accumulated reward received within each time interval (see \cite{mou2024bellman}). The difference is controlled by $O (\stepsize)$, and our analysis can be extended to this setting with minor modifications.}
Therefore, the control problem is cast as a discrete-time MDP with state space $\statespace = \real^\usedim$, action space $\actionspace$, transition kernel $\semigroup_\stepsize^\action$, discount factor $e^{- \discount \stepsize}$, and reward $\stepsize \reward (\MyState_{k \stepsize}, \Action_{k \stepsize})$.

A policy $\policy: \statespace \rightarrow \mathcal{P} (\actionspace)$ is a mapping from state space to probability distribution over action space. As the action can only change at discrete time steps, we consider policies that take actions according to $\Action_{k \stepsize} \sim \policy (\State_{k \stepsize})$ at time step $k \stepsize$, sampled independently from previous history. An idealized process that is mathematically more tractable is the aggregated process that evolves according to the averaged drift under the control policy:
\begin{align}
  d \widetilde{\MyState}_t = \drift^{\policy} (\widetilde{\MyState}_t) dt + \covMat (\widetilde{\MyState}_t)^{1/2} d \BM_t, \quad \mbox{where} ~ \drift^\policy (\state) \mydefn \Exs_{\Action \sim \policy(\state)} [\drift (\state, \Action)].\label{eq:cts-time-process-policy}
\end{align}
The prior work~\cite{jia2025accuracy} establishes the approximation error between the law of the actual control process~\eqref{eq:cts-time-process-action} and the aggregated diffusion process~\eqref{eq:cts-time-process-policy}.
\begin{proposition}[Theorem 4.1 of \cite{jia2025accuracy}]\label{prop:ouyang}
  Under suitable regularity conditions on the drift function $\drift$, diffusion matrix function $\covMat$, and policy $\policy$, there exists a constant $\couyang > 0$ depending on these regularity parameters, such that for any test function $g \in \Clin{4}$, we have \footnote{\cite{jia2025accuracy} did not explicitly provide the dependence on the time horizon $t$ in their Theorem 4.1. However, by carefully tracking the dependence on $t$ in their proof, we can obtain the exponential dependence shown here.}
\begin{align*}
  \abss{ \Exs [g (X_t)] - \Exs [g (\widetilde{X}_t)] } \leq \stepsize \exp \big(\couyang \cdot (1 + t) \big)  \vecnorm{g}{\Clin{4}}.
\end{align*}
\end{proposition}
Our work focuses learning control in a discrete-time formulation described above. Nevertheless, the aggregated process~\eqref{eq:cts-time-process-policy} and the approximation result provided by \Cref{prop:ouyang} serve as a mathematical tool for our analysis.

\subsection{Value functions and Bellman equation}\label{subsec:value-bellman}

Under the MDP formulation described in the previous section, optimal value function, optimal $Q$-function, and optimal policy are well-defined following standard approaches for discrete-time Markov decision processes~\cite{bertsekas1996neuro}. In particular, the value function under a policy $\policy$ is
\begin{align*}
  \valfun^\policy (\state) \mydefn \Exs^\policy \Big[ \sum_{k=0}^\infty e^{- \discount k \stepsize} \stepsize \reward (\MyState_{k \stepsize}, \Action_{k \stepsize}) \mid \State_0 = \state \Big].
\end{align*}
The optimal value function is defined as $\ValTrue (\state) \mydefn \sup_{\policy} \valfun^\policy (\state)$, with the maximum achieved by optimal policy $\policy^*$. These functions are well-defined as long as the reward function is bounded.

Playing an important role is the state-action value function (i.e. $Q$-function), defined as
\begin{align*}
  \Qfunc^\policy (\state, \action) \mydefn  \Exs^\policy \Big[ \sum_{k = 0}^\infty e^{- k \discount \stepsize} \stepsize \reward (\State_{k \stepsize}, \Action_{k \stepsize}) \mid \State_0 = \state, \Action_0 = \action \Big],
\end{align*}
and the optimal $Q$-function is  $\Qfunc^* (\state, \action) \mydefn \sup_{\policy} \Qfunc^\policy (\state, \action)$. It is known that the optimal $Q$-function satisfies the Bellman equation
\begin{align}
  \Qfunc^* (\state, \action) = \stepsize \reward (\state, \action) + e^{- \discount \stepsize} \Exs^\action \Big[ \max_{\action' \in \actionspace} \Qfunc^* (\State_\stepsize, \action') \mid \State_0 = \state \Big].\label{eq:big-Q-bellman}
\end{align}
This fixed-point equation serves as a starting point of the projected fixed-point analysis and algorithm derivation in \Cref{sec:algorithm} to follow.

\subsection{Observation model}\label{subsec:action-obs-model}
We consider an off-policy observation model, where the data is collected under a fixed and known behavior policy $\behavpolicy: \statespace \rightarrow \mathcal{P} (\actionspace)$, through the control protocol described in \Cref{subsec:mdp-setup}. At each discrete time step $k \stepsize$, the action is sampled from the behavior policy $\Action_{k \stepsize} \sim \behavpolicy (\State_{k \stepsize})$, and the process evolves according to~\eqref{eq:cts-time-process-action}.

Following standard random-horizon formulation in discounted MDPs~\cite{bertsekas1996neuro}, we consider $\mathrm{i.i.d.}$ trajectories where the length of each trajectory is a geometric random variable independent of the process. Concretely, given an initial distribution $\initDistr$ and the behavior policy $\behavpolicy$, we observe discretely-sampled trajectories
\begin{align*}
  \traj^{(i)} \mydefn \big\{(\State^{(i)}_{k \stepsize}, \Action^{(i)}_{k \stepsize}, \randreward^{(i)}_{k \stepsize}) \big\}_{k = 0}^{\lfloor T_i / \stepsize \rfloor} \quad \mbox{for } i = 1, 2, \ldots, \numobs,
\end{align*}
where the length $T_i$ of each trajectory is an independent exponential random variable with rate $\discount$. We define the set
\begin{align*}
  \Dset \mydefn \big\{ (i, k) \mid i = 1, 2, \ldots, \numobs, \quad k = 0, 1, \ldots, \lfloor T_i / \stepsize \rfloor \big\},
\end{align*}
which indexes all the observed state-action-reward tuples.
 The random reward $\randreward^{(i)}_{k \stepsize}$ is a noisy observation of the instantaneous reward, satisfying the unbiasedness condition $\Exs \big[ {\randreward^{(i)}_{k \stepsize} \mid \State^{(i)}_{k \stepsize}}, \Action^{(i)}_{k \stepsize} \big] = \reward (\State^{(i)}_{k \stepsize}, \Action^{(i)}_{k \stepsize})$. Throughout this paper, we assume that the observed rewards are uniformly bounded by $1$ almost surely.

Under this observation model, the discrete-time process $(\State_{k \stepsize})_{0 \leq k \leq \lfloor T_i / \stepsize \rfloor}$ is a Markov chain stopped at a geometric time with parameter $1 - e^{- \discount \stepsize}$. The transition kernel of this chain is
\begin{align*}
  \obstransition f (\state) \mydefn \int_\actionspace \semigroup_\stepsize^\action f (\state) \behavpolicy (d \action \mid \state).
\end{align*}
Playing a central role in our analysis are the occupancy measures induced by this stopped Markov chain. We define the state occupancy measure $\stationary \in \mathcal{P} (\statespace)$ and state-action occupancy measure $\Stationary \in \mathcal{P} (\statespace \times \actionspace)$ as
\begin{align*}
  \stationary \mydefn (1 - e^{- \discount \stepsize}) \sum_{k = 0}^\infty e^{- k \discount \stepsize} \big[ (\obstransition)^* \big]^k \initDistr, \quad \mbox{and}\quad \Stationary \mydefn \stationary \otimes \behavpolicy.
\end{align*}
In words, $\stationary$ is the probability distribution of a random state sampled from the trajectory, and $\Stationary$ is the joint distribution of a random state-action pair sampled from the trajectory.

\section{From projected fixed points to the algorithm}\label{sec:algorithm}
In this section, we derive the main algorithm, Sobolev-prox fitted $q$-learning algorithm, from the population-level fixed-point equations. We first introduce the population-level projected fixed-point equations in \Cref{subsec:projected-fp}. \Cref{subsec:population-iterates} presents a population-level iterative scheme to solve such equations. Finally, \Cref{subsec:data-driven-algorithm} derives the empirical version of the iterative scheme, leading to the proposed algorithm.

\subsection{Population-level projected fixed-point equation}\label{subsec:projected-fp}
For the RL with discretely-observed continuous-time processes, when the stepsize $\stepsize$ is small, the value function $v^*(x)$ itself scales as $O(1)$, while the $Q$-function $\Qfunc^* (x, a)$ differs from $v^*(x)$ by an amount of order $O(\stepsize)$. So if we directly approximate the $Q$-function, the effect of action would be easily buried in the error terms. Therefore, it is often more convenient to separate the value function and advantage function components of the $Q$-function. To this end, we define the auxiliary value function $\valueaux: \statespace \rightarrow \real$ and advantage function $\advFuncaux: \statespace \times \actionspace \rightarrow \real$ as
\begin{subequations}\label{eq:aux-value-advantage-def}
\begin{align}
  \valueaux (\state) & \mydefn  \Exs_{\Action \sim \behavpolicy (\state)} \Big[ \stepsize \reward (\state, \Action)  + e^{- \discount \stepsize} \ValTrue (\State_\stepsize) \mid \State_0 = \state \Big],\\
  \advFuncaux (\state, \action) & \mydefn \frac{1}{\stepsize} \big( \Qfunc^* (\state, \action) - \valueaux (\state) \big).
\end{align}
\end{subequations}
In other words, $\valueaux (\state)$ is the value at $\state$ by taking the behavior policy $\behavpolicy$ for one step and then following the optimal policy, and $\advFuncaux (\state, \action)$ is the advantage of taking action $\action$ at state $\state$ compared to taking the behavior policy $\behavpolicy$ at state $\state$.

The functions $(\valueaux,\advFuncaux)$ can be seen  as discrete-time version of the value function and advantage function in continuous-time RL literature~\cite{jia2023q, mou2024bellman}.
Moreover, the optimal value function and $Q$-function can be recovered from $\valueaux$ and $\advFuncaux$ via 
\begin{align}
\Qfunc^* = \valueaux + \stepsize \advFuncaux \quad \mbox{and} \quad \ValTrue (\state) = \valueaux (\state) + \stepsize \max_{\action} \advFuncaux (\state, \action).\label{eq:recover-val-Q-from-aux}
\end{align}
Following \Cref{eq:aux-value-advantage-def,eq:recover-val-Q-from-aux}, we can derive the fixed-point equations satisfied by $(\valueaux, \advFuncaux)$. To this end, we define the Bellman operators $\BellOp_\valfun$ and $\BellOp_\advFunc$ that act on pairs of functions $(\valfun, \advFunc)$ as
\begin{subequations}\label{eqs:bellman-operators}
\begin{align}
   \BellOp_\valfun [\valfun, \advFunc] (\state) &= \stepsize \reward^{\behavpolicy} (\state) + e^{- \discount \stepsize} \Exs^{\behavpolicy} \Big[ \valuefunc (\State_\stepsize) + \stepsize  \max_{\action' \in \actionspace} \advFunc (\State_\stepsize, \action') \mid \State_0 = \state \Big], \label{eq:bellman-op-in-valfunc} \\
   \BellOp_\advFunc [\valfun, \advFunc] (\state, \action) &=  \reward (\state, \action) - \reward^{\behavpolicy} (\state) +  \frac{e^{- \discount \stepsize}}{\stepsize} \big(\Exs^{\action} - \Exs^\behavpolicy \big) \big[  \valuefunc (\State_\stepsize) + \stepsize \max_{\action' \in \actionspace} \advFunc (\State_\stepsize, \action') \mid \State_0 = \state \big].\label{eq:bellman-op-in-advfunc}
\end{align}
\end{subequations}
By definition, the operators $\BellOp_\valfun$ and $\BellOp_\advFunc$ are affine in their first argument, but nonlinear in general, due to the presence of the maximization operation.

The pair of functions $(\valueaux, \advFuncaux)$ satisfy the coupled system of Bellman equations
\begin{align}
  \begin{dcases}
    \valueaux = \BellOp_\valfun [\valueaux, \advFuncaux],\\
    \advFuncaux = \BellOp_\advFunc [\valueaux, \advFuncaux].
  \end{dcases}\label{eq:discretized-bellman-eq}
\end{align}

\paragraph{Projected fixed-point equations:} To introduce function approximation, we consider projecting the Bellman operators in \Cref{eqs:bellman-operators} onto pre-specified function classes. This approach is widely used and foundational in RL algorithms~\cite{bertsekas1996neuro,munos2008finite}, and has been recently applied to continuous-time RL problems~\cite{jia2023q}.

We use the function classes $\Fclass_v$ and $\Fclass_q$ to approximate the value function and advantage function, respectively. We assume that both $\Fclass_v$ and $\Fclass_q$ are closed convex subsets of $\ltwospace(\statespace, \stationary)$ and $\ltwospace(\statespace \times \actionspace, \Stationary)$, respectively. Since the class $\Fclass_q$ is used to model the advantage function, it is invariant under adding a constant function in the action variable. Therefore, we assume that for any $\advFunc \in \Fclass_q$ and any state $\state \in \statespace$, we have
\begin{align}
  \int_{\actionspace} \advFunc (\state, \action) d \behavpolicy (\action \mid \state) = 0. \label{eq:advantage-mean-zero}
\end{align}
Since the behavior policy $\behavpolicy$ is known, this constraint can be easily incorporated into the function class $\Fclass_q$.


Recall that $\projectto{\Fclass_v}^{\stationary}$ and $\projectto{\Fclass_q}^{\Stationary}$ are the orthonormal projection operators onto $\Fclass_v$ and $\Fclass_q$ under the norms $\statnorm{\cdot}$ and $\Statnorm{\cdot}$, respectively. A natural way to introduce function approximation is to consider the projected fixed-point equations for $(\valuebar, \advFuncBar)$:
\begin{align}\label{eq:projected-fixed-point}
  \begin{dcases}
    \valuebar = \projectto{\Fclass_v}^{\stationary} \circ \BellOp_\valfun [\valuebar, \advFuncBar],\\
    \advFuncBar = \projectto{\Fclass_q}^{\Stationary} \circ \BellOp_\advFunc [\valuebar, \advFuncBar].
  \end{dcases}
\end{align}
When $\Fclass_v$ and $\Fclass_q$ are convex and compact under $\ltwospace$-norm, the solution $(\valuebar, \advFuncBar)$ is guaranteed to exist, by the Brouwer fixed-point theorem.

\subsection{Population-level iterates}\label{subsec:population-iterates}
Even with access to the exact Bellman operators, solving the projected fixed-point equations~\eqref{eq:projected-fixed-point} directly is challenging due to the coupled and nonlinear nature of the system. Therefore, we consider an iterative scheme to solve the system. Given an initial pair of functions $(\valuefunc^{(0)}, \advFunc^{(0)}) \in \Fclass_v \times \Fclass_q$, we generate a sequence of function pairs $\{(\valuefunc^{(t)}, \advFunc^{(t)})\}_{t = 0}^\infty$ via the updates
\begin{subequations}\label{eq:iterative-scheme-population}
\begin{align}
  \advFunc^{(t + 1)} & = \arg\min_{\advFunc \in \Fclass_q}  \Statnorm{\BellOp_\advFunc [\valuefunc^{(t)}, \advFunc^{(t)}] - \advFunc}^2 .\label{eq:iterative-scheme-advantage}\\
  \valuefunc^{(t + 1)} & = \arg\min_{\valfun \in \Fclass_v} \Big\{ \sobonorm{\valfun - \valfun^{(t)}}^2 + 2 \frac{\learningrate}{\stepsize} \statinprod{v^{(t)} - \BellOp_\ValFunc [\valuefunc^{(t)}, \advFunc^{(t + 1)}]}{\valfun - \ValFunc^{(t)}} \Big\} ,\label{eq:iterative-scheme-value}
\end{align}
\end{subequations}
where $\learningrate > 0$ is a step size parameter. In each iteration, we first update the advantage function and then use the updated advantage function to update the value function. The advantage function update~\eqref{eq:iterative-scheme-advantage} is simply a one-step fixed-point iteration for the projected Bellman equation in \Cref{eq:projected-fixed-point}.

The value function update~\eqref{eq:iterative-scheme-value}, on the other hand, deviates from a direct fixed-point iteration or $Q$-learning updates. Indeed, if we replace the Sobolev norm in~\eqref{eq:iterative-scheme-value} by the standard $\ltwospace$-norm, we will get
\begin{align*}
  \valuefunc^{(t + 1)} = \arg\min_{\valfun \in \Fclass_v} \statnorm{ \big(1 - \frac{\learningrate}{\stepsize} \big)\valfun^{(t)} + \frac{\learningrate}{\stepsize}   \BellOp_\ValFunc [\valuefunc^{(t)}, \advFunc^{(t + 1)}] - \valfun}^2,
\end{align*}
which reduces to a standard projected $Q$-learning update~\cite{bertsekas1996neuro}. However, this update may not converge, as the Bellman operator $\BellOp_\ValFunc$ is not a contraction under the $\ltwospace$-norm in general. By way of contrast, the proximal step~\eqref{eq:iterative-scheme-value} under Sobolev norm aligns the geometry of the update with the hidden positive definiteness structure of the Bellman operator. This structure-aware proximal update has been exploited in recent work on RL fine-tuning for diffusion generative models~\cite{mou2025rl}. Here we adapt this idea to the general model-free RL setting.
As we will see in \Cref{thm:population-level-convergence}, this modification leads to a linearly convergent iterative scheme.

\subsection{Data-driven algorithm}\label{subsec:data-driven-algorithm}
Now we are ready to turn the population-level iterative scheme~\eqref{eq:iterative-scheme-population} into a data-driven algorithm. The idea is to replace the population-level norms and inner products in \Cref{eq:iterative-scheme-population} by their empirical approximations based on the observed data. Given the trajectories $\{\traj^{(i)}\}_{i = 1}^\numobs$, the projected operator step in \Cref{eq:iterative-scheme-advantage} can be solved by simply running a least-squares regression. In particular, we define the projected operator for the advantage function as
\begin{subequations}
  \begin{multline*}
    \empbelladv (\valfun, \advFunc) \mydefn \arg\min_{g \in \Fclass_q}  
   \sum_{(i, k) \in \Dset}  \Big\{  \randreward^{(i)}_{k \stepsize}  + \frac{e^{- \discount \stepsize}}{\stepsize} \big(\valfun (\State_{(k + 1) \stepsize}^{(i)}) - \valfun (\State_{k \stepsize}^{(i)}) \big)\\
    + e^{ - \discount \stepsize} \max_{\action' \in \actionspace} \advFunc(\State_{(k + 1) \stepsize}^{(i)}, \action') - g (\State_{k \stepsize}^{(i)}, \Action_{k \stepsize}^{(i)})  \Big\}^2
  \end{multline*}
  Note that following the definition of the Bellman operator~\eqref{eq:bellman-op-in-advfunc} for the advantage function, the second term in the regression target needs to be $\frac{e^{- \discount \stepsize}}{\stepsize} \big(\valfun (\State_{(k + 1) \stepsize}^{(i)}) - \obstransition\valfun (\State_{k \stepsize}^{(i)}) \big)$, instead of $\frac{e^{- \discount \stepsize}}{\stepsize} \big(\valfun (\State_{(k + 1) \stepsize}^{(i)}) - \valfun (\State_{k \stepsize}^{(i)}) \big)$ used in our empirical approximation. Since the transition operator $\obstransition$ is unknown, we use the value function at the current state $\State_{k \stepsize}^{(i)}$ as a proxy. This approximation does not change the regression target at the population-level, since functions in $\Fclass_q$ have zero mean under the behavior policy by \Cref{eq:advantage-mean-zero}.

  The empirical advantage function update can be written as
    \begin{align}
    \advFuncHat^{(t + 1)} = \empbelladv (\valuehat^{(t)}, \advFuncHat^{(t)}) .\label{eq:empirical-iterative-scheme-advantage}
  \end{align}
  To derive the value function update, we approximate both the Sobolev norm and inner product using empirical counterparts. In particular, we define the empirical Sobolev inner product
  \begin{align*}
    \empsoboinprod{f}{g} \mydefn \frac{1 - e^{- \discount \stepsize}}{\numobs} \sum_{(i, k) \in \Dset} f (\State_{k \stepsize}^{(i)}) g (\State_{k \stepsize}^{(i)}) +  \frac{1 - e^{- \discount \stepsize}}{\numobs} \sum_{(i, k) \in \Dset} \nabla f (\State_{k \stepsize}^{(i)})^\top \nabla g (\State_{k \stepsize}^{(i)}).
  \end{align*}
  As for the term involving the Bellman operator, we use the empirical approximation
  \begin{align*}
    \empbilinear{\valfun_1}{\valfun_2}{\advFunc} \mydefn  \frac{1 - e^{- \discount \stepsize}}{\numobs} \sum_{(i, k) \in \Dset} \Big\{ \valfun_1 (\State_{k \stepsize}^{(i)}) - \stepsize \randreward^{(i)}_{k \stepsize} - e^{- \discount \stepsize} \valfun_1 (\State_{(k + 1) \stepsize}^{(i)}) - \stepsize e^{- \discount \stepsize} \max_{\action' \in \actionspace} \advFunc (\State_{(k + 1) \stepsize}^{(i)}, \action')  \Big\}  \valfun_2 (\State_{k \stepsize}^{(i)}).
  \end{align*}
  With these notations, the empirical value function update can be written as
  \begin{align}
  \valuehat^{(t + 1)}  = \arg\min_{\valfun \in \Fclass_v} \Big\{ \empsoboinprod{\valuefunc - \valuehat^{(t)}}{\ValueFunc - \valuehat^{(t)}}  + \frac{2 \learningrate}{\stepsize} \empbilinear{\valuehat^{(t)}}{\valfun - \valuehat^{(t)}}{\advFuncHat^{(t + 1)}} \Big\}. \label{eq:empirical-iterative-scheme-value}
  \end{align}
  Starting from an initial pair $(\valuehat^{(0)}, \advFuncHat^{(0)})$, we alternate between the updates~\eqref{eq:empirical-iterative-scheme-advantage} and~\eqref{eq:empirical-iterative-scheme-value} for $N$ iterations to obtain the final estimates $(\valuehat^{(N)}, \advFuncHat^{(N)})$. We call this procedure the \emph{Sobolev-prox fitted $q$-learning} algorithm.
\end{subequations}

The dependence introduced by reusing the same dataset across iterations poses technical challenges in the analysis. To address this issue, we employ the technique of \emph{sample-splitting}, which is standard in literature (see e.g.~\cite{chandrasekher2023sharp}). Specifically, given a dataset of $2 N \numobs$ trajectories, we partition the dataset into $2N$ disjoint subsets, each containing $\numobs$ trajectories. In iteration $t$, we utilize the $2t - 1$-th subset to perform the update~\eqref{eq:empirical-iterative-scheme-advantage}, and the $2t$-th subset to perform the update~\eqref{eq:empirical-iterative-scheme-value}. As we will show in \Cref{thm:main}, since the optimization error converges linearly, the sample-splitting technique only results in a logarithmic factor loss in the final error bound. In practice, one can use the entire dataset in each iteration without sample-splitting, and we conjecture that similar performance guarantees would still hold.

\section{Theoretical guarantees}\label{sec:theory}
In this section, we provide theoretical guarantees on the proposed method. We start by introducing technical assumptions in \Cref{subsec:technical-assumptions}, followed by approximation guarantees of the projected fixed points in \Cref{subsec:approximation-guarantees}. \Cref{subsec:population-level-convergence} and \Cref{subsec:data-driven-convergence} present convergence guarantees for the population-level iterates~\eqref{eq:iterative-scheme-population} and the Sobolev-prox fitted $q$-learning algorithm, respectively.

\subsection{Technical assumptions}\label{subsec:technical-assumptions}
We introduce several technical assumptions that will be used in the theoretical analysis.

\begin{assumption}[uniform ellipticity]\label{assume:ellipticity}
  There exist constants $0 < \lammin \leq \lammax < \infty$ such that for any state $\state \in \statespace$,
  \begin{align*}
    \lammin I \preceq \covMat (\state) \preceq \lammax I.
  \end{align*}
\end{assumption}

\begin{assumption}[coefficient regularity]\label{assume:coefficient-regularity}
  There exists constant $\creg > 0$ such that the drift function $\drift$ and diffusion matrix function $\covMat$ satisfies the uniform bounds
  \begin{align*}
    \max_{1 \leq \vecnorm{\alpha}{1} \leq 4}\sup_{\state, \action} \eucnorm{\partial^\alpha \drift (\state, \action)} + \sup_{\state, \action} \frac{\eucnorm{ \drift (\state, \action)}}{1 + \eucnorm{x}} + \max_{0 \leq \vecnorm{\alpha}{1} \leq 4}\sup_{\state} \eucnorm{\partial^\alpha \covMat (\state)} \leq \creg,
  \end{align*}
\end{assumption}
Ellipticity and regularity conditions on the coefficients are standard in the literature of diffusion processes~\cite{pavliotis2016stochastic}, and widely used in continuous-time RL~\cite{mou2024bellman,mou2025statistical}. They also guarantee the discretization error bound in \Cref{prop:ouyang}. Note that here we allow the drift function $\drift$ to be unbounded, as long as it grows at most linearly in the state variable.

Additionally, we need the following stability condition to ensure that the diffusion process does not diverge too fast.
\begin{assumption}[stability]\label{assume:stability}
  There exist constant $\constbd > 0$ such that $\Exs \big[ |X_0|^p \big] \leq (p B)^{p/2}$ for any $p \geq 2$, and for any state $\state \in \statespace$ and action $\action \in \actionspace$,
  \begin{align*}
    \drift (\state, \action)^\top \state \leq \constbd , \quad \mbox{for any } \state \in \real^\usedim. 
  \end{align*}
\end{assumption}
This assumption ensures that the diffusion process $(X_t)_{t \geq 0}$ initialized from $\initDistr$ has bounded moments, and grows at most linearly in time. This assumption is substantially weaker than the dissipativity or ergodicity conditions that are commonly used in the literature~\cite{bakry2008simple}, and is satisfied by a wide range of diffusion processes.

Our next assumption involves regularity conditions on the function classes $\Fclass_v$ and $\Fclass_q$, as well as the true solutions $(\valueaux, \advFuncaux)$.
\begin{assumption}[function class]\label{assume:function-class}
  The function classes $\Fclass_v$ and $\Fclass_q$ are closed convex subsets of $\ltwospace(\statespace, \stationary)$ and $\ltwospace(\statespace \times \actionspace, \Stationary)$, respectively. 
  Furthermore, for any function $f \in \Fclass_v $ or $f \in \Fclass_q $, we have $
    \vecnorm{f}{C^6} \leq \constFclass$.
    Furthermore, there exists a constant $\ctwofour > 0$ such that for any $v \in \Fclass_v - \Fclass_v$ or $q \in \Fclass_q - \Fclass_q$, we have
    \begin{align*}
      \vecnorm{v}{\mathbb{L}^4} \leq \ctwofour \vecnorm{v}{\mathbb{L}^2},\quad \vecnorm{\nabla v}{\mathbb{L}^4} \leq \ctwofour \vecnorm{\nabla v}{\mathbb{L}^2} \quad \mbox{and} \quad \vecnorm{q}{\mathbb{L}^4} \leq \ctwofour \vecnorm{q}{\mathbb{L}^2}.
    \end{align*}
    Moreover, the true solutions $(\valueaux, \advFuncaux)$ satisfy uniform bounds $\vecnorm{\valueaux}{\infty} \leq \constFclass$ and $\vecnorm{\advFuncaux}{\infty} \leq \constFclass$.
\end{assumption}
The smoothness assumptions on the function classes and true solutions are imposed for convenience. They may be relaxed to lower-order smoothness at the cost of more involved technical arguments. If the true solutions $(\valueaux, \advFuncaux)$ do not satisfy the smoothness conditions, we can consider their smooth approximations instead, and the resulting approximation error can be incorporated into the final error bounds.

\begin{assumption}[action coverage]\label{assume:coverage}
   For any function $\advFunc \in \Fclass_q - \Fclass_q$, we have
    \begin{align}
      \max_{\action \in \actionspace} \abss{\advFunc (\state, \action) } \leq \constcmp \sqrt{\int_\actionspace \advFunc (\state, \action)^2 d \behavpolicy (\action \mid \state)}, \quad \mbox{for any } \state \in \statespace.\label{eq:action-space-compare}
    \end{align}
\end{assumption}

The coverage condition~\eqref{eq:action-space-compare} is imposed only on the action space and the function class $\Fclass_q$. It requires that the supremum norm of the advantage function over the action space can be controlled by its variance under the behavior policy. It is easy to see that this condition holds under the following scenarios:
\begin{itemize}
  \item When the action space $\actionspace$ is finite, and the behavior policy $\behavpolicy$ has a uniform lower bound. In this case, we can take $\constcmp$ to be the inverse of the minimum probability mass.
  \item When the action space is compact subset of $\real^m$, and the behavior policy $\behavpolicy$ has a density function bounded from below. Since the functions in the class $\Fclass_q$ are sufficiently smooth, we can use Sobolev embedding theorems to establish~\eqref{eq:action-space-compare}.
\end{itemize}

Finally, we need the following regularity condition on the time marginal densities of the diffusion process.

\begin{assumption}[density regularity]\label{assume:density-regularity}
  The time marginal density $\marginal_t$ of the diffusion process $(X_t)_{t \geq 0}$ initialized from $\initDistr$ satisfies
  \begin{align*}
   \Big\{ \int \eucnorm{\nabla_x \log \marginal_t (x)}^{p}\marginal_t (x) d x \Big\}^{1/p} \leq  \exp \big(\cmalliavin (1 + t) \big)  \sqrt{p}, \quad \mbox{for any } t \geq 0,
  \end{align*}
  for any $p \geq 2$. Furthermore, for the second derivative, we have
  \begin{align*}
    \int \marginal_t (x)^{-1} \eucnorm{\nabla_x^2  \marginal_t (x)}^2 d x 
    \leq  \exp \big(\cmalliavin (1 + t) \big).
  \end{align*}
\end{assumption}
Assumption~\ref{assume:density-regularity} controls the high-order moments of the score function of the time marginal density. Such regularity conditions have been established using Malliavin calculus techniques; see, for example,~\cite{menozzi2021density,li2024estimates}. As the focus of this paper is on the statistical and algorithmic aspects of continuous-time RL with function approximation, we refer the readers to these references for more discussions on sufficient conditions for Assumption~\ref{assume:density-regularity} to hold.

\subsection{Approximation guarantees of the projected-fixed points}\label{subsec:approximation-guarantees}
With the assumptions in place, we are ready to present our first main results. We start with the approximation guarantees of the projected fixed-point equations~\eqref{eq:projected-fixed-point}. To state the result, we need to impose a lower bound on the discount rate $\discount$:
\begin{align}
  \discount \geq c \cdot \max \Big\{ \cmalliavin^2 , \couyang, \constcmp^2  \Big\}, \label{eq:discount-rate-lower-bound}
\end{align}
for a sufficiently large constant $c > 0$ depending on the regularity parameters. This lower bound ensures that the effective horizon of the RL problem is bounded by a constant. It is important direction of future work to extend the boundary on the discount rate to a wider range.

Under this condition, we can bound the approximation error of the projected fixed-point solution $(\valuebar, \advFuncBar)$ by the best approximation errors achievable within the function classes $(\Fclass_v,\Fclass_q)$.
\begin{theorem}\label{lemma:projected-fixed-point-approximation}
  Under \Cref{assume:coefficient-regularity,assume:ellipticity,assume:density-regularity,assume:function-class,assume:stability}, there exists a constant $c > 0$ depending on the regularity parameters, such that when the discount rate $\discount$ satisfies~\eqref{eq:discount-rate-lower-bound}, the solution $(\valuebar, \advFuncBar)$ to the projected fixed-point equation~\eqref{eq:projected-fixed-point} satisfies
  \begin{align*}
    \sobonorm{\valuebar - \valueaux} + \Statnorm{\advFuncBar - \advFuncaux} \leq c  \Big\{ \inf_{\valfun \in \Fclass_v} \sobonorm{\valfun - \valueaux} +  \inf_{\advFunc \in \Fclass_q} \Statnorm{\advFunc - \advFuncaux} \Big\} + c \sqrt{\stepsize}.
  \end{align*}
\end{theorem}
\noindent See~\Cref{subsec:proof-lemma-projected-fixed-point-approximation} for the proof of this theorem. \Cref{lemma:projected-fixed-point-approximation} controls the $\soboone$-Sobolev norm error of the value function and the $\ltwospace$-norm error of the advantage function, and bound them by the best approximation errors within their respective function classes under the same norms. These norms are natural choices, as the gradient of the value function and the advantage function are used to derive the optimal policies~\cite{jia2023q,mou2025rl}.

The discretization error term of order $\sqrt{\stepsize}$ is incurred when we leverage continuous-time structures to analyze the discrete-time MDP. It is important direction of future work to deploy techniques from advanced numerical schemes to improve this error, as has been done in the policy evaluation setting~\cite{mou2024bellman}.

\Cref{lemma:projected-fixed-point-approximation} establishes the projected solutions $(\valuebar, \advFuncBar)$ as desirable targets to approximate. In the next two subsections, we provide convergence guarantees for both the population-level iterates and the data-driven algorithm towards these targets.

\subsection{Convergence of the population-level iterates}\label{subsec:population-level-convergence}
Let us present the convergence guarantees for the population-level iterative scheme~\eqref{eq:iterative-scheme-population}. We need the following upper bound on the learning rate $\learningrate$:
\begin{align}
  \learningrate \leq c_1 \min \Big\{ \frac{\lammin}{ \lammax^2}, \discount \Big\}, \label{eq:learning-rate-upper-bound}
\end{align}
for a sufficiently small constant $c_1 > 0$ depending on the regularity parameters. This upper bound ensures that the optimization updates are stable, and it allows the learning rate to be of constant order. Under this condition, we have the following linear convergence guarantee.
\begin{theorem}\label{thm:population-level-convergence}
   Under the setup of \Cref{lemma:projected-fixed-point-approximation}, let $\{(\valuefunc^{(t)}, \advFunc^{(t)})\}_{t = 0}^\infty$ be the iterates generated by the scheme~\eqref{eq:iterative-scheme-population} with learning rate satisfying \Cref{eq:learning-rate-upper-bound}, we have
   \begin{align*}
    \sobonorm{\valuefunc^{(N)} - \valuebar}^2 + \Statnorm{\advFunc^{(N)} - \advFuncBar}^2 
    \leq c  \exp \big( - \frac{\lammin \learningrate N}{4} \big) \sobonorm{\valuefunc^{(0)} - \valuebar}^2  + c \stepsize,
   \end{align*}
    where the constants $c, c_1 > 0$ depend on problem parameters.
\end{theorem}
\noindent See~\Cref{subsec:proof-thm-population-level-convergence} for the proof of this theorem. A few remarks are in order. First, \Cref{thm:population-level-convergence} ensure exponential contraction of the iterates towards an $O(\sqrt{\stepsize})$-neighbor of the projected fixed-point solution $(\valuebar, \advFuncBar)$. Combined with \Cref{lemma:projected-fixed-point-approximation}, this result establishes the population-level iterates as effective approximations to the true solutions $(\valueaux, \advFuncaux)$. The convergence speed is governed by the learning rate $\learningrate$ and the ellipticity constant $\lammin$. Second, the error bound only depends on the initial error of the value function but not the advantage function. This is because the value function update~\eqref{eq:iterative-scheme-value} uses a proximal step that stabilizes the optimization process, while the advantage function update~\eqref{eq:iterative-scheme-advantage} is a direct fixed-point iteration that does not depend heavily on the previous advantage function estimate.

\subsection{Convergence of the data-driven algorithm}\label{subsec:data-driven-convergence}
Finally, we present the convergence guarantees for the Sobolev-prox fitted $q$-learning algorithm introduced in \Cref{subsec:data-driven-algorithm}. To state the result, we need to introduce the notion of critical radii that capture the statistical complexity of the function classes $\Fclass_v$ and $\Fclass_q$. To start with, we define the localized function classes
\begin{align*}
  \Fclass_\valfun (r) & \mydefn \{ \valfun_1 - \valfun_2: \sobonorm{\valfun_1 - \valfun_2} \leq r, \valfun_1, \valfun_2 \in \Fclass_v \}, \quad  \Fclass_\advFunc (r)  \mydefn \{ \advFunc_1 - \advFunc_2: \Statnorm{\advFunc_1 - \advFunc_2} \leq r, \advFunc_1, \advFunc_2 \in \Fclass_q \}.
\end{align*}
For the advantage function class, we define the critical radii $r^*_\advFunc$ as the smallest positive solution to the inequality
\begin{subequations}
\begin{align}
  r^2 = \dudley_2 (\Fclass_\advFunc (r), \Stationary) \frac{\log^3 (n / \delta)}{\sqrt{\numobs}} + \dudley_1 (\Fclass_\advFunc, \mathbb{L}^\infty)  \frac{\log^4 (\numobs / \delta)}{n}.
\end{align}
Similarly, for the value function class, we define the critical radius $r^*_\valfun$ as the smallest positive solution to the inequality
\begin{align}
  r^2 = \dudley_2 (\Fclass_\valfun (r), \soboone) \frac{\log^3 (n / \delta)}{\sqrt{\numobs}} + \dudley_1 (\Fclass_\valfun, C^1)  \frac{\log^4 (\numobs / \delta)}{n}.
\end{align}
\end{subequations}
In these critical radii definitions, we use Dudley chaining integrals with exponent $1/2$ and $1$ to capture the complexity of the function classes under different norms. The $\dudley_2$ integrals are defined under the norms of interest, and they come from the variances of the empirical processes. The $\dudley_1$ integrals are defined under stronger uniform norms, and they arise from high-order terms in Bernstein-type concentration inequalities. The term involving $\dudley_2$ typically dominates when the sample size $\numobs$ is large. And in many concrete examples, such radii match the minimax optimal rates up to logarithmic factors.

In least-square regression, these critical radii characterize the statistical error of the empirical risk minimizers~\cite{Wainwright_nonasymptotic}. Here, they play a similar role in controlling the statistical error of the Sobolev-prox fitted $q$-learning algorithm. With these definitions, we have the following convergence guarantee. 
\begin{theorem}\label{thm:main}
  Under the setup of \Cref{thm:population-level-convergence}, let $\{(\valuehat^{(t)}, \advFuncHat^{(t)})\}_{t = 0}^N$ be the iterates generated by the data-driven scheme~\eqref{eq:empirical-iterative-scheme-advantage} and~\eqref{eq:empirical-iterative-scheme-value}. For any $\delta < \min (1/N, \stepsize)$, we have
  \begin{align*}
    &\sobonorm{\valuehat^{(N)} - \valueaux}^2 + \Statnorm{\advFuncHat^{(N)} - \advFuncaux}^2 \\
    &\leq c  \Big\{  \inf_{\valfun \in \Fclass_v} \sobonorm{\valfun - \valueaux}^2 +  \inf_{\advFunc \in \Fclass_q} \Statnorm{\advFunc - \advFuncaux}^2 +  \frac{(r^*_\valfun)^2}{\learningrate^2} + (r^*_\advFunc)^2 + e^{ - \tfrac{\lammin \learningrate N}{8}} \sobonorm{\valuehat^{(0)} - \valuebar}^2 + \stepsize \Big\}, 
  \end{align*}
  with probability at least $1 - \delta$. Here, the constant $c > 0$ depends on problem parameters.
\end{theorem}
\noindent See~\Cref{subsec:proof-thm-main} for the proof of this theorem. A few remarks are in order. First, the error bound consists of four components: the approximation errors from \Cref{lemma:projected-fixed-point-approximation}, the optimization error from \Cref{thm:population-level-convergence}, the statistical errors characterized by the critical radii, and an additional $O(\stepsize)$ discretization error. This gives an oracle inequality that decouples the approximation, optimization, and statistical errors. Such guarantees are standard in nonparametric regression and statistical learning theory~\cite{Wainwright_nonasymptotic}, but rarely exist in the RL literature with function approximation.

Second, the optimization error decays exponentially fast in the number of iterations $N$, and it suffices to run the algorithm for $N = O \big( \frac{1}{\learningrate} \log (1 / \varepsilon) \big)$ iterations to ensure that this error is at most $\varepsilon$. Consequently, the use of sample-splitting only results in a logarithmic factor loss in the final error bound, which is mild in practice.

The form of \Cref{thm:main} is comparable to previous results on policy evaluation~\cite{mou2025statistical} and diffusion model fine-tuning~\cite{mou2025rl}. However, we do not achieve the self-mitigating error phenomenon observed in these prior works, where the noise levels in the statistical error scales with the approximation error. It remains an important open question to investigate whether such phenomenon exists in the value learning setting.

\paragraph{Examples:} It is useful to instantiate the critical radii in \Cref{thm:main} for specific function classes. Here we present two cases.

\begin{itemize}
\item \textbf{Parametric class:} For parametric function classes, we can typically bound the covering number as
\begin{align*}
  \log N (\varepsilon; \Fclass_\valfun, C^1) \lesssim d_v \log (1 / \varepsilon)\quad \mbox{and} \quad\log N (\varepsilon; \Fclass_\advFunc, \mathbb{L}^\infty) \lesssim d_q \log (1 / \varepsilon)
\end{align*}
for some dimensions $d_v, d_q > 0$. Using these bounds, we can compute the critical radii as
\begin{align*}
  r^*_\valfun \lesssim \log^3 (\numobs / \delta) \sqrt{\frac{d_v }{\numobs}}, \quad r^*_\advFunc \lesssim \log^3 (\numobs / \delta) \sqrt{\frac{d_q}{\numobs}}, \qquad \mbox{for } \numobs \gtrsim d_v + d_q.
\end{align*}
\item \textbf{Nonparametric class:} For nonparametric function classes, suppose that the log covering numbers satisfy polynomial upper bounds
\begin{align*}
  \log N (\varepsilon; \Fclass_\valfun, C^1) \lesssim \varepsilon^{- \omega_v}, \quad \log N (\varepsilon; \Fclass_\advFunc, \mathbb{L}^\infty) \lesssim \varepsilon^{- \omega_q}.
\end{align*}
Suppose that $\omega_v, \omega_q \in (0, 1)$, we can compute the critical radii as
\begin{align*}
  r^*_\valfun \lesssim \log^3 (\numobs / \delta) \cdot \numobs^{- \tfrac{1}{2 + \omega_v}}, \quad r^*_\advFunc \lesssim \log^3 (\numobs / \delta) \cdot \numobs^{- \tfrac{1}{2 + \omega_q}},
\end{align*}
  for sufficiently large $\numobs$. For many common nonparametric classes such as RKHS, this rate matches the minimax optimal rate up to logarithmic factors~\cite{Wainwright_nonasymptotic}.

\end{itemize}

\section{Proofs}\label{sec:proofs}
We collect the proofs of the main results in this section, with several technical lemmas deferred to the appendix.

\subsection{Proof of \Cref{lemma:projected-fixed-point-approximation}}\label{subsec:proof-lemma-projected-fixed-point-approximation}
By first-order optimality condition of the projection~\eqref{eq:projected-fixed-point}, we have
\begin{align*}
  \statinprod{\BellOp_\valfun [\valuebar, \advFuncBar] - \valuebar}{\valfun - \valuebar} \leq 0 \quad \mbox{for any } \ValFun \in \Fclass_v,\\
  \Statinprod{\BellOp_\advFunc [\valuebar, \advFuncBar] - \advFuncBar}{\advFunc - \advFuncBar} \leq 0 \quad \mbox{for any } \advFunc \in \Fclass_q.
\end{align*}
Combining it with the Bellman equations~\eqref{eq:discretized-bellman-eq}, we have
\begin{align*}
    \statinprod{\BellOp_\valfun [\valuebar, \advFuncBar] - \BellOp_\valfun [\valueaux, \advFuncaux] - \valuebar + \valueaux}{\valfun - \valuebar} \leq 0 \quad \mbox{for any } \ValFun \in \Fclass_v,\\
  \Statinprod{\BellOp_\advFunc [\valuebar, \advFuncBar] - \BellOp_\advFunc [\valueaux, \advFuncaux] - \advFuncBar + \advFuncaux}{\advFunc - \advFuncBar} \leq 0 \quad \mbox{for any } \advFunc \in \Fclass_q.
\end{align*}
Rearranging yields
\begin{subequations}
\begin{align}
  &\statinprod{ \big(\valuefunc - \BellOp_\valfun [\ValFunc, \advFuncaux] \big) - \big( \valuebar - \BellOp_\valfun [\valuebar, \advFuncBar] \big)}{\valfun - \valuebar} \leq \statinprod{ \big(\valuefunc - \BellOp_\valfun [\ValFunc, \advFuncaux] \big) - \big( \valueaux - \BellOp_\valfun [\valueaux, \advFuncaux] \big)}{\valfun - \valuebar},\label{eq:fixed-pt-cond-in-approx-err-proof-valfunc} \\
   &\Statnorm{\advFunc - \advFuncBar}^2 \leq \Statinprod{\advFunc - \advFuncaux}{\advFunc - \advFuncBar} + \Statinprod{\BellOp_\advFunc [\valueaux, \advFuncaux] - \BellOp_\advFunc [\valuebar, \advFuncBar]}{\advFunc - \advFuncBar}.\label{eq:fixed-pt-cond-in-approx-err-proof-advfunc}
\end{align}
\end{subequations}
To analyze the system of inequalities, we use the following lemmas.
\begin{lemma}\label{lemma:bellman-positive-definite}
  Under the assumptions of \Cref{lemma:projected-fixed-point-approximation}, for any functions $\ValFun_1, \valfun_2 \in \Fclass_v $ and advantage functions $\advFunc_1, \advFunc_2 \in \Fclass_\advFunc \cup \{ \advFuncaux\}$, we have
  \begin{multline*}
   \stepsize^{-1} \statinprod{\big( \valfun_1 - \BellOp_\valuefunc [\valfun_1, \advFunc_1] \big) - \big( \valfun_2 - \BellOp_\valuefunc [\valfun_2, \advFunc_2] \big)}{\valfun_1 - \valfun_2}\\
    \geq \frac{\discount}{4} \statnorm{\valfun_1 - \valfun_2}^2 + \lammin \sobonorm{\valfun_1 - \valfun_2}^2 - \constcmp \statnorm{\valfun_1 - \valfun_2} \cdot \Statnorm{\advFunc_1 - \advFunc_2} - c \stepsize. 
  \end{multline*}
  where the constant $c > 0$ depends on problem parameters.
\end{lemma}
\noindent See Section~\ref{subsubsec:proof-lemma-bellman-positive-definite} for the proof of this lemma.

\begin{lemma}\label{lemma:bellman-bounded}
   Under the assumptions of \Cref{lemma:projected-fixed-point-approximation}, for any functions $\ValFun_1, \valfun_2 \in \Fclass_v \cup \{ \valueaux\}$, $\valfun \in \Fclass_v - \Fclass_v$, and an advantage functions $\advFunc_1, \advFunc_2 \in \Fclass_q \cup \{\advFuncaux\}$, we have
   \begin{multline*}
     \stepsize^{-1} \abss{ \statinprod{\big( \valfun_1 - \BellOp_\valuefunc [\valfun_1, \advFunc_1] \big) - \big( \valfun_2 - \BellOp_\valuefunc [\valfun_2, \advFunc_2] \big)}{\valfun} } \\
     \leq \lammax \sobonorm{\valfun_1 - \valfun_2} \sobonorm{\ValFun} + c \ctwofour \statnorm{\valfun_1 - \valfun_2}  \sobonorm{\ValFun} 
     + \constcmp \statnorm{\valfun_1 - \valfun_2} \Statnorm{\advFunc_1 - \advFunc_2} + c \stepsize,
   \end{multline*}
    where the constant $c > 0$ depends on problem parameters.
\end{lemma}
\noindent See Section~\ref{subsubsec:proof-lemma-bellman-bounded} for the proof of this lemma.

Finally, we have the following bound on the Bellman operator for the advantage function.
\begin{lemma}\label{lemma:semigroup-difference-bound}
  Under the assumptions of \Cref{lemma:projected-fixed-point-approximation}, for any value functions $\ValFun_1, \valfun_2 \in \Fclass_v \cup \{\valueaux\}$, advantage functions $\advFunc_1, \advFunc_2 \in \Fclass_q \cup \{\advFuncaux\}$, and $\advFunc \in \Fclass_\advFunc - \Fclass_\advFunc$, we have
  \begin{align*}
    \abss{\Statinprod{\BellOp_\advFunc [\ValueFunc_1, \advFunc_1] - \BellOp_\advFunc [\valuefunc_2, \advFunc_2]}{\advFunc}} \leq c \ctwofour \sobonorm{\valfun}  \Statnorm{\advFunc} + c_2 \stepsize.
  \end{align*}
\end{lemma}
\noindent See Section~\ref{subsubsec:proof-lemma-semigroup-difference-bound} for the proof of this lemma.

Taking these lemmas as given, we now proceed with the proof of \Cref{lemma:projected-fixed-point-approximation}. Define the projected functions
\begin{align}
  \valproj \mydefn \arg\min_{\valfun \in \Fclass_v} \sobonorm{\valfun - \valueaux}, \quad \advfuncproj \mydefn \arg\min_{\advFunc \in \Fclass_q} \Statnorm{\advFunc - \advFuncaux}.\label{eq:projection-definitions}
\end{align}
Note that \Cref{eq:fixed-pt-cond-in-approx-err-proof-valfunc} and \Cref{eq:fixed-pt-cond-in-approx-err-proof-advfunc} hold for any $\ValFun \in \Fclass_v$ and $\advFunc \in \Fclass_q$. In particular, we can substitute $\ValFun = \valproj$ and $\advFunc = \advfuncproj$ into the two inequalities.

Applying \Cref{lemma:bellman-positive-definite} to the left-hand-side of \Cref{eq:fixed-pt-cond-in-approx-err-proof-valfunc}, and \Cref{lemma:bellman-bounded} to the right-hand-side, we obtain
\begin{align}
  &\frac{\discount}{4} \statnorm{\valuebar - \valproj}^2 + \lammin \sobonorm{\valuebar - \valproj}^2 \nonumber\\ 
  &\leq \lammax \sobonorm{\valuebar - \valproj} \sobonorm{\valproj - \valueaux}
   + c \ctwofour \statnorm{\valuebar - \valproj}  \sobonorm{\valproj - \valueaux} \nonumber \\
   &\qquad 
  + \constcmp \statnorm{\valuebar - \valproj} \Statnorm{\advFuncBar - \advFuncaux} + c \stepsize \label{eq:approx-err-proof-valfunc-intermediate}
\end{align}
Similarly, applying \Cref{lemma:semigroup-difference-bound} to the second term on the right-hand-side of \Cref{eq:fixed-pt-cond-in-approx-err-proof-advfunc}, we obtain
\begin{align}
  \Statnorm{\advFuncBar - \advfuncproj}^2 \leq \Statnorm{\advFuncBar - \advfuncproj} \cdot \Statnorm{\advfuncproj - \advFuncaux} + c \ctwofour \sobonorm{\valproj - \valuebar} \cdot  \Statnorm{\advFuncBar - \advfuncproj} + c_2 \constFclass \stepsize .\label{eq:approx-err-proof-advfunc-intermediate}
\end{align}
Solving the inequality~\eqref{eq:approx-err-proof-advfunc-intermediate} for $\Statnorm{\advFuncBar - \advfuncproj}$ yields
\begin{align}
  \Statnorm{\advFuncBar - \advfuncproj} \leq \Statnorm{\advfuncproj - \advFuncaux} + c \ctwofour \sobonorm{\valproj - \valuebar} + c' \sqrt{\stepsize}.\label{eq:approx-err-proof-advfunc-bound}
\end{align}
Substituting this bound back into \Cref{eq:approx-err-proof-valfunc-intermediate}, we arrive at an inequality for the value function approximation error:
\begin{align*}
  &\frac{\discount}{4} \statnorm{\valuebar - \valproj}^2 + \lammin \sobonorm{\valuebar - \valproj}^2 \\
  &\leq \lammax \sobonorm{\valuebar - \valproj} \sobonorm{\valproj - \valueaux}
   + c \ctwofour  \statnorm{\valuebar - \valproj}  \sobonorm{\valproj - \valueaux} \\
   &\qquad 
  + \constcmp \statnorm{\valuebar - \valproj} \big( \Statnorm{\advfuncproj - \advFuncaux} + c \ctwofour \sobonorm{\valproj - \valuebar} + c' \sqrt{\stepsize} \big) + c \stepsize.
\end{align*}
When the discount factor satisfies the bound $
  \discount \geq \frac{8 c^2 \constcmp^2 \ctwofour^2}{\lammin}$, we have
\begin{align*}
  \constcmp \statnorm{\valuebar - \valproj} \cdot c \ctwofour \sobonorm{\valproj - \valuebar} \leq \frac{\discount}{8} \statnorm{\valuebar - \valproj}^2 + \frac{\lammin}{2} \sobonorm{\valproj - \valuebar}^2.
\end{align*}
Applying this inequality and solving for $ \sobonorm{\valuebar - \valproj}$, we obtain that
\begin{align*}
  \sobonorm{\valuebar - \valproj} & \leq c_1 \ctwofour \sobonorm{\valproj - \valueaux} + c_2 \Statnorm{\advfuncproj - \advFuncaux} + c_3 \sqrt{\stepsize},
\end{align*}
for some constants $c_1, c_2, c_3 > 0$ depending on problem parameters. Note that above inequality also implies a bound on $\Statnorm{\advFuncBar - \advfuncproj}$ via \Cref{eq:approx-err-proof-advfunc-bound}. Combining these two bounds with the triangle inequality, we arrive at the desired result.

\subsubsection{Some technical lemmas}\label{subsubsec:technical-lemmas}
Let us first collect some technical lemmas about the diffusion process and its infinitesimal generator, which will be used in the proof of \Cref{lemma:projected-fixed-point-approximation}.

\begin{lemma}\label{lemma:generator-positive-definite}
  Under the setup of \Cref{lemma:projected-fixed-point-approximation}, for any function $f \in \Fclass_v - \Fclass_v$, we have
  \begin{align*}
  \statinprod{f}{\big(\discount - \generator^{\behavpolicy} \big) f} \geq \frac{\discount}{2} \statnorm{f}^2 + \lammin \statnorm{\nabla f}^2 - c_0 \stepsize,
  \end{align*}
  for a constant $c_0$ depending on the parameters in the assumptions and $\couyang$ in \Cref{prop:ouyang}
\end{lemma}
\noindent See~\Cref{subsubsec:proof-lemma-generator-positive-definite} for the proof of this lemma.

\begin{lemma}\label{lemma:generator-upper-bound}
   Under the setup of \Cref{lemma:projected-fixed-point-approximation}, for any pair of functions $f, g$ and any action $\action \in \actionspace$, we have the bounds
  \begin{align*}
  \abss{\statinprod{f}{\generator^{\action}  g}} &\leq \lammax \sobonorm{f} \cdot \sobonorm{g} + c \vecnorm{f}{\lpspace{4} (\stationary)} \cdot \statnorm{\nabla g},\\
  \abss{\statinprod{f}{\generator^{\action}  g}} & \leq \lammax \sobonorm{f} \cdot \sobonorm{g} +  c  \statnorm{f} \cdot \vecnorm{\nabla g}{\lpspace{4} (\stationary)}\\
  \abss{\statinprod{f}{\generator^{\action} g}} &\leq c \statnorm{g} \cdot \vecnorm{f}{C^2}.
  \end{align*}
  for a constant $c$ depending on the parameters in the assumptions.
\end{lemma}
\noindent See~\Cref{subsubsec:proof-lemma-generator-upper-bound} for the proof of this lemma.

Note that since the generator $\generator^{\action}$ is linear in the drift function $\drift$, \Cref{lemma:generator-upper-bound} also holds when $\generator^{\action}$ is replaced by the operator $\generator^{\behavpolicy}$.

The following technical lemma controls the growth of the process $\{X_{t}\}_{t \geq 0}$ and $\{\widetilde{X}_t\}_{t \geq 0}$ using the stability condition in \Cref{assume:stability}.
\begin{lemma}\label{lemma:process-moment-bound}
  Under \Cref{assume:stability} and \Cref{assume:ellipticity}, for any $p \geq 2$, we have
  \begin{align*}
   \Big\{ \Exs \Big[\sup_{\leq t \leq T} \eucnorm{X_{t}}^p \Big] \Big\}^{1/p} \leq c_0 (1+T) \sqrt{p}, \quad \mbox{and} \quad
   \Big\{ \Exs \Big[\sup_{\leq t \leq T} \eucnorm{\widetilde{X}_{t}}^p \Big] \Big\}^{1/p} \leq c_0 (1+T) \sqrt{p},
  \end{align*}
  where the constant $c_0 > 0$ depends on the parameters in the assumptions.
\end{lemma}
\noindent See~\Cref{subsubsec:proof-lemma-process-moment-bound} for the proof of this lemma.

Finally, we have the following lemma on the gradient and Hessian of the log-density of the occupancy measure $\stationary$.
\begin{lemma}\label{lemma:occupancy-measure-gradient-log-density}
  Under the setup of \Cref{lemma:projected-fixed-point-approximation}, for $\discount \geq 2 \cmalliavin p$, we have
  \begin{align*}
    \Big\{ \int \eucnorm{\nabla_x \log \stationary (x)}^{p}\stationary (x) d x \Big\}^{1/p} \leq 2 e^{\cmalliavin} \sqrt{p},\quad \mbox{and} \quad \int \frac{\eucnorm{\nabla_x^2  \stationary (x)}^2}{\stationary (x)} d x \leq 2 e^{\cmalliavin}.
  \end{align*}
\end{lemma}
\noindent See~\Cref{app:subsec-proof-lemma-occupancy-measure-gradient-log-density} for the proof of this lemma.

\subsubsection{Proof of~\Cref{lemma:bellman-positive-definite}}\label{subsubsec:proof-lemma-bellman-positive-definite}
By definition, we have
\begin{multline*}
  \stepsize^{-1}\big( \valfun_1 - \BellOp_\valuefunc [\valfun_1, \advFunc_1] \big) - \stepsize^{-1} \big( \valfun_2 - \BellOp_\valuefunc [\valfun_2, \advFunc_2] \big) \\
  = \frac{ \IdMat - e^{- \discount \stepsize} \obstransition }{\stepsize} (\valfun_1 - \valfun_2) + e^{- \discount \stepsize} \obstransition \big[ \max_{\action' \in \actionspace} \advFunc_1 (\cdot, \action') - \max_{\action' \in \actionspace} \advFunc_2 (\cdot, \action') \big].
\end{multline*}
Define the terms
\begin{align*}
  \Term_1 &\mydefn \stepsize^{-1} \statinprod{\big( I - e^{- \discount \stepsize} \obstransition \big) (\valfun_1 - \valfun_2)}{\valfun_1 - \valfun_2}, \quad \mbox{and}\\
  \Term_2 &\mydefn e^{- \discount \stepsize} \statinprod{\obstransition \big[ \max_{\action' \in \actionspace} \advFunc_1 (\cdot, \action') - \max_{\action' \in \actionspace} \advFunc_2 (\cdot, \action') \big]}{\valfun_1 - \valfun_2}.
\end{align*}
In the following, we bound the terms $\Term_1$ and $\Term_2$ separately.

\paragraph{Lower bound for $\Term_1$:}
For notational convenience, we let $\valfun \mydefn \valfun_1 - \valfun_2$. Applying It\^{o}'s formula, we have
\begin{align*}
  \big( I - e^{- \discount \stepsize} \obstransition \big) \valuefunc (\state)& = \int_\actionspace  \big( I - e^{- \discount \stepsize} \semigroup_\stepsize^\action \big) \valuefunc (\state) \behavpolicy ( d\action \mid \state) \\
  &= \int_\actionspace \Exs_\action \Big[ \int_0^\stepsize e^{- \discount t} \big( \discount \valuefunc (\MyState_t) - \generator^\action \valuefunc (\MyState_t) \big) dt \mid \MyState_0 = \state \Big] \behavpolicy ( d\action \mid \state)\\
  &= \int_0^\stepsize \int_\actionspace e^{- \discount t}  \big( \discount  - \generator^\action  \big) \semigroup_t^\action \valuefunc (\state) \behavpolicy ( d\action \mid \state) dt\\
  &= (\discount - \generator^{\behavpolicy}) \valuefunc (\state) \cdot \stepsize + \int_0^\stepsize \int_\actionspace e^{- \discount t}  \big( \discount  - \generator^\action  \big) \big( \semigroup_t^\action - \IdMat \big) \valuefunc (\state) \behavpolicy ( d\action \mid \state) dt.
\end{align*}
By \Cref{lemma:generator-positive-definite}, when $\discount \geq 2 \couyang$, we have
\begin{align*}
  \statinprod{(\discount - \generator^{\behavpolicy}) \valuefunc}{\valuefunc} \geq \frac{\discount}{2} \statnorm{\valfun}^2 + \lammin \statnorm{\nabla \valfun}^2 - c \stepsize.
\end{align*}
It remains to bound the residual term. We note that
\begin{align*}
 &\abss{ \frac{1}{\stepsize} \int_\statespace \int_0^\stepsize \int_\actionspace e^{- \discount t}  \valfun (x) \cdot \big( \discount  - \generator^\action  \big) \big( \semigroup_t^\action - \IdMat \big) \valuefunc (\state) \behavpolicy ( d\action \mid \state) \stationary (\state) dt d \state}\\
 &\leq \sup_{t \in [0, \stepsize]} \sup_{\action \in \actionspace} \abss{ \statinprod{\valfun}{\big( \discount  - \generator^\action  \big) \big( \semigroup_t^\action - \IdMat \big) \valuefunc} }\\
 &\leq  \sup_{\action \in \actionspace} \int_0^\stepsize \abss{ \statinprod{\valfun}{(\discount - \generator^\action) \generator^\action \semigroup_t^\action \valuefunc} } dt\\
 &\leq \stepsize \sup_{\action \in \actionspace} \sup_{t \in [0, \stepsize]} \statnorm{\valfun} \cdot \statnorm{\big( \discount  - \generator^\action  \big) \big( \semigroup_t^\action - \IdMat \big) \valuefunc}.
\end{align*}
Since $\ValFunc \in \valfuncspace - \valfuncspace$, by \Cref{assume:function-class,assume:coefficient-regularity} and \Cref{lemma:process-moment-bound}, we have $\statnorm{\big( \discount  - \generator^\action  \big) \big( \semigroup_t^\action - \IdMat \big) \valuefunc} \leq c_1$ for some constant $c_1 > 0$ depending on the problem parameters. Putting the pieces together, we conclude that
\begin{align}
   \statinprod{\frac{ \IdMat - e^{- \discount \stepsize} \obstransition }{\stepsize} (\valfun_1 - \valfun_2)}{\valfun_1 - \valfun_2} \geq \frac{\discount}{2} \statnorm{\valfun_1 - \valfun_2}^2 + \lammin \statnorm{\nabla \valfun_1 - \nabla \valfun_2}^2 - c \stepsize.\label{eq:term-1-bound-in-lemma-bellman-positive-definite}
\end{align}
\paragraph{Upper bound for $|\Term_2|$:} Define the function
\begin{align*}
  h (\state) \mydefn \max_{\action' \in \actionspace} \advFunc_1 (\state, \action') - \max_{\action' \in \actionspace} \advFunc_2 (\state, \action').
\end{align*}
We have the decomposition
\begin{align*}
  |\Term_2| &\leq \abss{\statinprod{h}{\valfun_1 - \valfun_2}} + \abss{\statinprod{(\obstransition - \IdMat) h}{\valfun_1 - \valfun_2}}\\
  &\leq \statnorm{h} \cdot \statnorm{\valfun_1 - \valfun_2} + \int_0^\stepsize \abss{\statinprod{\generator^{\behavpolicy} \semigroup_t^{\behavpolicy} h}{\valfun_1 - \valfun_2}} dt.
\end{align*}
In order to bound the term $\statnorm{h}$, we apply \Cref{assume:coverage} to obtain
\begin{align*}
  \abss{h (\state)} \leq \max_{\action' \in \actionspace} \abss{ \big( \advFunc_1 - \advFunc_2 \big) (\state, \action') } \leq \constcmp \sqrt{\int_\actionspace \big(\advFunc_1 - \advFunc_2)^2 (\state, \action) d \behavpolicy (\action \mid \state)},
\end{align*}
and consequently, we have
\begin{align*}
  \statnorm{h} \leq \constcmp \Statnorm{\advFunc_1 - \advFunc_2}.
\end{align*}
It remains to bound the residual term. Invoking the last inequality of \Cref{lemma:generator-upper-bound}, we have
\begin{align*}
  \abss{\statinprod{\generator^{\behavpolicy} \semigroup_t^{\behavpolicy} h}{\valfun_1 - \valfun_2}} \leq c \statnorm{ \semigroup_t^{\behavpolicy} h} \cdot \vecnorm{\valfun_1 - \valfun_2}{C^2} \leq c \vecnorm{\valfun_1 - \valfun_2}{C^2} \cdot \vecnorm{\advFunc_1 - \advFunc_2}{\infty}.
\end{align*}
Since $\valfun_1, \valfun_2 \in \Fclass_v$ and $\advFunc_1, \advFunc_2 \in \Fclass_\advFunc \cup \{\advFuncaux\}$, invoking \Cref{assume:function-class}, we can bound the above term by $4 c \Fclass^2$. Putting the pieces together, we conclude that
\begin{align}
  |\Term_2| \leq \constcmp \statnorm{\valfun_1 - \valfun_2} \cdot \Statnorm{\advFunc_1 - \advFunc_2} + c \stepsize,\label{eq:term-2-bound-in-lemma-bellman-positive-definite}
\end{align}
Combining the bounds~\eqref{eq:term-1-bound-in-lemma-bellman-positive-definite} and~\eqref{eq:term-2-bound-in-lemma-bellman-positive-definite} concludes the proof.

\subsubsection{Proof of~\Cref{lemma:bellman-bounded}}\label{subsubsec:proof-lemma-bellman-bounded}
By definition, we have
\begin{multline*}
  \stepsize^{-1}\big( \valfun_1 - \BellOp_\valuefunc [\valfun_1, \advFunc_1] \big) - \stepsize^{-1} \big( \valfun_2 - \BellOp_\valuefunc [\valfun_2, \advFunc_2] \big)\\
  = \frac{ \IdMat - e^{- \discount \stepsize} \obstransition }{\stepsize} (\valfun_1 - \valfun_2) + e^{- \discount \stepsize} \obstransition \big[ \max_{\action' \in \actionspace} \advFunc_1 (\cdot, \action') - \max_{\action' \in \actionspace} \advFunc_2 (\cdot, \action') \big].
\end{multline*}
Define the two terms
\begin{align*}
  \Term_1 &\mydefn \stepsize^{-1} \statinprod{\big( I - e^{- \discount \stepsize} \obstransition \big) (\valfun_1 - \valfun_2)}{\valuefunc}, \quad \mbox{and}\\
  \Term_2 &\mydefn e^{- \discount \stepsize} \statinprod{\obstransition \big[ \max_{\action' \in \actionspace} \advFunc_1 (\cdot, \action') - \max_{\action' \in \actionspace} \advFunc_2 (\cdot, \action') \big]}{\valuefunc}.
\end{align*}
The term $\Term_2$ is exactly the same as that in the proof of \Cref{lemma:bellman-positive-definite}, and we have the bound
\begin{align*}
  |\Term_2| \leq \constcmp \statnorm{\valfun_1 - \valfun_2} \cdot \Statnorm{\advFunc_1 - \advFunc_2} + c \stepsize.
\end{align*}
It remains to bound the term $\Term_1$.
Similar to the proof of \Cref{lemma:bellman-positive-definite}, we use It\^{o}'s formula to write
\begin{align*}
    \frac{1}{\stepsize}  \big( I - e^{- \discount \stepsize} \obstransition \big) \valuefunc (\state) = (\discount - \generator^{\behavpolicy}) \valuefunc (\state)  + \frac{1}{\stepsize}\int_0^\stepsize \int_\actionspace e^{- \discount t}  \big( \discount  - \generator^\action  \big) \big( \semigroup_t^\action - \IdMat \big) \valuefunc (\state) \behavpolicy ( d\action \mid \state) dt.
\end{align*}
By \Cref{lemma:generator-upper-bound}, we have
\begin{align*}
  \abss{\statinprod{(\discount - \generator^{\behavpolicy}) (\valuefunc_1 - \valfun_2)}{\valuefunc}} 
  &\leq \lammax \sobonorm{\valfun_1 - \valfun_2} \cdot \sobonorm{\ValFun} + c \vecnorm{\valfun_1 - \valfun_2}{\stationary}  \cdot \vecnorm{\nabla \ValFun}{\lpspace{4} (\stationary)}\\
  &\leq \lammax \sobonorm{\valfun_1 - \valfun_2} \cdot \sobonorm{\ValFun} + c \ctwofour \statnorm{\valfun_1 - \valfun_2}  \cdot \sobonorm{\ValFun},
\end{align*}
for a constant $c_1 > 0$ depending on the problem parameters. By \Cref{assume:function-class}, we have $\vecnorm{\valfun_1 - \valfun_2}{\infty} \leq 2 \constFclass$.

For the residual term, we use the same argument as in the proof of \Cref{lemma:bellman-positive-definite} to obtain
\begin{align*}
  &\abss{ \frac{1}{\stepsize} \int_\statespace \int_0^\stepsize \int_\actionspace e^{- \discount t}  (\valfun_1 - \valfun_2) (\state) \cdot \big( \discount  - \generator^\action  \big) \big( \semigroup_t^\action - \IdMat \big) \valuefunc (\state) \behavpolicy ( d\action \mid \state) \stationary (\state) dt d \state} \\
  &\leq \sup_{\action \in \actionspace} \int_0^\stepsize \abss{ \statinprod{\valfun_1 - \valfun_2}{(\discount - \generator^\action) \generator^\action \semigroup_t^\action \valuefunc} } dt \leq c \stepsize,
\end{align*}
where the constant $c > 0$ depends on the regularity parameters in \Cref{assume:coefficient-regularity,assume:function-class}. Putting the pieces together, we conclude the desired bound.

\subsubsection{Proof of~\Cref{lemma:semigroup-difference-bound}}\label{subsubsec:proof-lemma-semigroup-difference-bound}
By definition, we have
\begin{align*}
  \BellOp_\advFunc [\ValueFunc_1, \advFunc_1] - \BellOp_\advFunc [\valuefunc_2, \advFunc_2] &= \frac{e^{- \discount \stepsize}}{\stepsize} \big( \semigroup_\stepsize^{\,\centerdot} - \obstransition \big) \big[  \valuefunc_1 - \valuefunc_2 + \stepsize \big(\max_{\action' \in \actionspace} \advFunc_1 (\cdot, \action') - \max_{\action' \in \actionspace} \advFunc_2 (\cdot, \action') \big) \big].
\end{align*}
We prove upper bounds for the following two terms separately:
\begin{align*}
  \Term_1 &\mydefn \Statinprod{\frac{\semigroup_\stepsize^{\,\centerdot} - \obstransition}{\stepsize} \big(  \valuefunc_1 - \valuefunc_2 \big)}{\advFunc}, \quad \mbox{and}\\
  \Term_2 &\mydefn \Statinprod{\big( \semigroup_\stepsize^{\,\centerdot} - \obstransition \big) \big(\max_{\action' \in \actionspace} \advFunc_1 (\cdot, \action') - \max_{\action' \in \actionspace} \advFunc_2 (\cdot, \action') \big)}{\advFunc}.
\end{align*}

\paragraph{Upper bound for $\Term_1$:} For notational convenience, we introduce $\valfun \mydefn \valfun_1 - \valfun_2$. Expanding the inner product, we have
\begin{align*}
  \Statinprod{\frac{\semigroup_\stepsize^{\,\centerdot} - \obstransition}{\stepsize} \ValFun}{\advFunc} &= \frac{1}{\stepsize} \int_\statespace \int_\actionspace \big( \semigroup_\stepsize^\action  - \obstransition  \big) \valuefunc (\state) \cdot \advFunc (\state, \action) \behavpolicy ( d\action \mid \state) \stationary (\state) d \state
\end{align*}
By It\^{o}'s formula, we have
\begin{align*}
  \frac{1}{\stepsize}\big(\semigroup_\stepsize^\action  - \obstransition \big) \valuefunc (\state) &= \frac{1}{\stepsize}\int_0^\stepsize \Big\{ \semigroup_t^\action \generator^\action \valuefunc (\state) - \int_\actionspace \semigroup_t^{\action'} \generator^{\action'} \valuefunc (\state) \behavpolicy ( d\action' \mid \state) \Big\} dt \\
  &= \big( \generator^\action - \generator^{\behavpolicy} \big) \valuefunc (\state) \behavpolicy  +  \frac{1}{\stepsize}\int_0^\stepsize \big(\semigroup_t^\action - \IdMat \big) \generator^\action \valuefunc (\state) dt - \frac{1}{\stepsize}\int_0^\stepsize \int_\actionspace \big(\semigroup_t^{\action'} - \IdMat \big) \generator^{\action'} \valuefunc (\state) \behavpolicy ( d\action' \mid \state) dt.
\end{align*}
For the first term, we note that
\begin{align*}
  \int_\statespace \int_\actionspace \big( \generator^\action - \generator^{\behavpolicy} \big) \valuefunc (\state) \cdot \advFunc (\state, \action) \behavpolicy ( d\action \mid \state) \stationary (\state) d \state = \int_\statespace \int_\actionspace \big(\drift^{\action} (\state) - \drift^{\behavpolicy} (\state) \big)^\top \nabla \valuefunc (\state) \cdot \advFunc (\state, \action) \Stationary ( d\state, d\action).
\end{align*}
Invoking H\"older's inequality and \Cref{assume:coefficient-regularity,assume:function-class}, we have
\begin{align*}
  \abss{ \int_\statespace \int_\actionspace \big( \generator^\action - \generator^{\behavpolicy} \big) \valuefunc (\state) \cdot \advFunc (\state, \action) \behavpolicy ( d\action \mid \state) \stationary (\state) d \state } \leq c \sobonorm{\valfun} \cdot \lpnorm{\advFunc}{4}{\Stationary} \leq c \ctwofour \sobonorm{\valfun} \cdot \Statnorm{\advFunc}.
\end{align*}
It remains to bound the two residual terms. We use a similar argument as in the proof of \Cref{lemma:bellman-positive-definite,lemma:bellman-bounded}. In particular, we note that
\begin{align*}
  (\semigroup_t^{\action} - \IdMat) f (\state) = \int_0^t \semigroup_s^{\action} \generator^{\action} f (\state) ds,
\end{align*}
for any function $f: \statespace \rightarrow \real$. Consequently, we have
\begin{multline*}
  \abss{ \frac{1}{\stepsize}\int_0^\stepsize \int_\statespace \int_\actionspace \big(\semigroup_t^{\action} - \IdMat \big) \generator^{\action} \valuefunc (\state) \cdot \advFunc (\state, \action) \behavpolicy ( d\action \mid \state) \stationary (\state) d \state dt }\\
  \leq \int_0^\stepsize \abss{ \int_{\statespace \times \actionspace}   \big(\generator^{\action} \big)^2 \semigroup_t^\action  \valuefunc (\state) \cdot \advFunc (\state, \action) \Stationary ( d\state, d\action)} dt \leq c_1 \stepsize \vecnorm{\valfun}{\infty} \cdot \sum_{|\alpha| \leq 4} \vecnorm{\partial_x^\alpha \advFunc}{\infty},
\end{multline*}
for some constant $c_1 > 0$ depending on the problem parameters. In the last step, we use the third inequality of \Cref{lemma:generator-upper-bound}, \Cref{assume:coefficient-regularity}, \Cref{lemma:process-moment-bound}, and the fact that $\semigroup_t^\action$ is a contraction operator in the supremum norm. 

Similarly, for the other residual term, we have
\begin{multline*}
  \abss{ \frac{1}{\stepsize}\int_0^\stepsize \int_\statespace \int_\actionspace \int_\actionspace \big(\semigroup_t^{\action'} - \IdMat \big) \generator^{\action'} \valuefunc (\state) \cdot \advFunc (\state, \action) \behavpolicy ( d\action' \mid \state) \behavpolicy ( d\action \mid \state) \stationary (\state) d \state dt }\\
  \leq \int_0^\stepsize \abss{ \int_{\statespace \times \actionspace}   \big(\generator^\action \big)^2 \semigroup_t^\action  \valuefunc (\state) \cdot \advFunc (\state, \action) \Stationary ( d\state, d\action)} dt \leq c_1 \stepsize \vecnorm{\valfun}{\infty} \cdot \sum_{|\alpha| \leq 4} \vecnorm{\partial_x^\alpha \advFunc}{\infty},
\end{multline*}
For $\advFunc \in \Fclass_\advFunc - \Fclass_\advFunc$, by \Cref{assume:function-class}, we have $\vecnorm{\partial_x^\alpha \advFunc}{\infty} \leq 2 \constFclass$ for any multi-index $\alpha$ with $|\alpha| \leq 4$. Putting them together, we conclude that
\begin{align}
  |T_1| \leq c \ctwofour \sobonorm{\valfun} \cdot \Statnorm{\advFunc} + c \stepsize \vecnorm{\valfun}{\infty} \cdot \vecnorm{\advFunc}{C^4} \leq c \ctwofour \sobonorm{\valfun} \cdot \Statnorm{\advFunc} + c \stepsize.\label{eq:term-1-bound-in-lemma-semigroup-difference-bound}
\end{align}

\paragraph{Upper bound for $\Term_2$:} Define the function
\begin{align*}
  h (\state) \mydefn \max_{\action' \in \actionspace} \advFunc_1 (\state, \action') - \max_{\action' \in \actionspace} \advFunc_2 (\state, \action').
\end{align*}
We can write the term $\Term_2$ in the following form:
\begin{align*}
  \Term_2 = \int_{\statespace \times \actionspace} \big( \semigroup_\stepsize^\action - \obstransition \big) h (\state) \cdot \advFunc (\state, \action) \Stationary ( d\state, d\action).
\end{align*}
Similar to the bound for $\Term_1$, we have
\begin{align*}
  \big(\semigroup_\stepsize^\action  - \obstransition \big) h (\state) = \int_0^\stepsize \Big\{ \generator^\action \semigroup_t^\action  h (\state) - \int_\actionspace \generator^{\action'} \semigroup_t^{\action'}  h (\state) \behavpolicy ( d\action' \mid \state) \Big\} dt
\end{align*}
By the third inequality of \Cref{lemma:generator-upper-bound}, for fixed pair $\action \in \actionspace$, we have
\begin{align*}
  \abss{ \int_{\statespace \times \actionspace}  \generator^\action \semigroup_t^\action  h (\state) \cdot \advFunc (\state, \action) \Stationary ( d\state, d\action)} \leq c \statnorm{\semigroup_t^\action  h} \cdot \vecnorm{\advFunc}{C^2} \leq c \vecnorm{h}{\infty} \cdot \vecnorm{\advFunc}{C^2}.
\end{align*}
This bound also applies toe the integral term over $\action'$, by simply replacing $\action$ with $\action'$. 
Since $\advFunc_1, \advFunc_2 \in \Fclass_\advFunc \cup \{\advFuncaux\}$, by \Cref{assume:function-class}, we have $\vecnorm{h}{\infty} \leq 2 \constFclass$. Putting the pieces together, we conclude that
\begin{align}
  |\Term_2| \leq c \stepsize \vecnorm{\advFunc}{C^2}.\label{eq:term-2-bound-in-lemma-semigroup-difference-bound}
\end{align} Combining the bounds~\eqref{eq:term-1-bound-in-lemma-semigroup-difference-bound} and~\eqref{eq:term-2-bound-in-lemma-semigroup-difference-bound} concludes the proof.

\subsection{Proof of \Cref{thm:population-level-convergence}}
\label{subsec:proof-thm-population-level-convergence}
We use the following lemmas that characterize the contraction properties of the one-step population-level updates.
\begin{lemma}\label{lemma:value-function-contraction}
  Under the setup of \Cref{thm:population-level-convergence}, we have
  \begin{multline*}
    \sobonorm{\valfun^{(t + 1)} - \ValBar}^2 + \frac{\sobonorm{\valfun^{(t + 1)} - \valfun^{(t)}}^2}{3} \\
  \leq \big(1 - \frac{\lammin \learningrate}{2} \big) \sobonorm{ \valfun^{(t)} - \ValBar}^2 - \frac{\discount}{8} \learningrate \statnorm{\valfun^{(t)} - \ValBar}^2 + 4 \learningrate \constcmp \Statnorm{\advFunc^{(t + 1)} - \advFuncBar} \cdot \statnorm{\valfun^{(t)} - \ValBar} + 5 c \learningrate \stepsize.
\end{multline*}
\end{lemma}
\noindent See~\Cref{subsubsec:proof-lemma-value-function-contraction} for the proof of this lemma.

\begin{lemma}\label{lemma:advantage-function-contraction}
  Under the setup of \Cref{thm:population-level-convergence}, we have
  \begin{align*}
   \Statnorm{\advFunc^{(t + 1)} - \advFuncBar} \leq c \sqrt{\ctwofour} \sobonorm{\valuefunc^{(t)} - \ValBar} + c \sqrt{ \stepsize},
  \end{align*}
  for a constant $c > 0$ depending on the problem parameters.
\end{lemma}
\noindent See ~\Cref{subsubsec:proof-lemma-advantage-function-contraction} for the proof of this lemma.

Taking these two lemmas as given for the moment, we now complete the proof of \Cref{thm:population-level-convergence}. Substituting the bound for $\Statnorm{\advFunc^{(t + 1)} - \advFuncBar}$ from \Cref{lemma:advantage-function-contraction} into the inequality of \Cref{lemma:value-function-contraction}, we obtain
\begin{align*}
   & \sobonorm{\valfun^{(t + 1)} - \ValBar}^2  \\
  &\leq \big(1 - \frac{\lammin \learningrate}{2} \big) \sobonorm{ \valfun^{(t)} - \ValBar}^2 - \frac{\discount}{8} \learningrate \statnorm{\valfun^{(t)} - \ValBar}^2 + 4 \constcmp c \learningrate \sqrt{\ctwofour} \sobonorm{\valuefunc^{(t)} - \ValBar} \cdot \statnorm{\valfun^{(t)} - \ValBar} \\
  &\qquad \qquad + 5 c \learningrate \stepsize + 4 c \learningrate \sqrt{\stepsize} \cdot \statnorm{\valfun^{(t)} - \ValBar}\\
& \leq \big(1 - \frac{\lammin \learningrate}{2} \big) \sobonorm{ \valfun^{(t)} - \ValBar}^2 - \frac{\discount}{8} \learningrate \statnorm{\valfun^{(t)} - \ValBar}^2 + 6 \constcmp c \learningrate \sqrt{\ctwofour} \sobonorm{\valuefunc^{(t)} - \ValBar} \cdot \statnorm{\valfun^{(t)} - \ValBar} + 7 c \learningrate \stepsize.
\end{align*}
When the discount rate $\discount$ satisfies the condition
\begin{align*}
  \discount \geq \frac{288 c^2 \constcmp^2}{\lammin} \ctwofour,
\end{align*}
we have
\begin{align*}
- \frac{\discount}{8} \learningrate \statnorm{\valfun^{(t)} - \ValBar}^2 + 6 \constcmp c \learningrate \sqrt{\ctwofour} \sobonorm{\valuefunc^{(t)} - \ValBar} \cdot \statnorm{\valfun^{(t)} - \ValBar} \leq \frac{\lammin \learningrate}{4} \sobonorm{\valfun^{(t)} - \ValBar}^2.
\end{align*}
Under such condition, the recursion simplifies to
\begin{align*}
  \sobonorm{\valfun^{(t + 1)} - \ValBar}^2 \leq \big(1 - \frac{\lammin \learningrate}{4} \big) \sobonorm{ \valfun^{(t)} - \ValBar}^2 + 7 c \learningrate \stepsize.
\end{align*}
Unrolling the recursion yields
\begin{align*}
  \sobonorm{\valfun^{(N)} - \ValBar} \leq  \exp \big( - \frac{\lammin \learningrate N}{8} \big) \sobonorm{ \valfun^{(0)} - \ValBar} + c' \sqrt{\stepsize}.
\end{align*}
For the advantage function, substituting the above bound into \Cref{lemma:advantage-function-contraction} yields
\begin{align*}
  \Statnorm{\advFunc^{(N)} - \advFuncBar} \leq c  \Big\{ \exp \big( - \frac{\lammin \learningrate N}{8} \big) \sobonorm{ \valfun^{(0)} - \ValBar} + c' \sqrt{\stepsize} \Big\} 
\end{align*}
This concludes the proof of \Cref{thm:population-level-convergence}.

\subsubsection{Proof of Lemma~\ref{lemma:value-function-contraction}}\label{subsubsec:proof-lemma-value-function-contraction}
By the optimality condition of the proximal operator in the updates~\eqref{eq:iterative-scheme-value}, for any $\gamma \in [0, 1]$, since $\gamma \ValFun^{(t + 1)} + (1 - \gamma)\ValBar \in \Fclass_\valfun$, we have
\begin{multline*}
  \sobonorm{\valfun^{(t + 1)} - \valfun^{(t)}}^2 + 2 \frac{\learningrate}{\stepsize} \statinprod{v^{(t)} - \BellOp_\ValFunc [\valuefunc^{(t)}, \advFunc^{(t + 1)}]}{\valfun^{(t + 1)} - \ValFunc^{(t)}} \\
  \leq \sobonorm{\gamma \valfun^{(t + 1)} + (1 - \gamma) \ValBar - \valfun^{(t)}}^2 \\
  + 2 \frac{\learningrate}{\stepsize} \statinprod{v^{(t)} - \BellOp_\ValFunc [\valuefunc^{(t)}, \advFunc^{(t + 1)}]}{\gamma \valfun^{(t + 1)} + (1 - \gamma) \ValBar - \ValFunc^{(t)}}
\end{multline*}
Taking the limit $\gamma \rightarrow 1$ yields
\begin{align*}
  \soboinprod{\valfun^{(t)} - \valfun^{(t + 1)}}{\ValBar - \valfun^{(t + 1)}} - \frac{\learningrate}{\stepsize} \statinprod{\valuefunc^{(t)} - \BellOp_\ValFunc [\valuefunc^{(t)}, \advFunc^{(t + 1)}]}{\ValBar - \valfun^{(t + 1)} } \leq 0.
\end{align*}
By the projected fixed-point condition~\eqref{eq:projected-fixed-point}, we have
\begin{align*}
   \statinprod{\valuebar - \BellOp_\valfun [\valuebar, \advFuncBar]}{\valuebar - \valfun^{(t + 1)}} \leq 0.
\end{align*}
Combining the two inequalities yields
\begin{align*}
\soboinprod{\valfun^{(t)} - \valfun^{(t + 1)}}{\ValBar - \valfun^{(t + 1)}} + \frac{\learningrate}{\stepsize} \statinprod{\big( \valuebar - \BellOp_\valfun [\valuebar, \advFuncBar] \big) - \big(\valuefunc^{(t)} - \BellOp_\ValFunc [\valuefunc^{(t)}, \advFunc^{(t + 1)}] \big)}{\ValBar - \valfun^{(t + 1)} } \leq 0.
\end{align*}
Rearranging the terms, we have
\begin{multline}
  \sobonorm{\valfun^{(t + 1)} - \ValBar}^2 - \sobonorm{\valfun^{(t)} - \ValBar}^2 + \sobonorm{\valfun^{(t + 1)} - \valfun^{(t)}}^2 \\
  \leq - 2 \frac{\learningrate}{\stepsize} \statinprod{\big( \valuebar - \BellOp_\valfun [\valuebar, \advFuncBar] \big) - \big(\valuefunc^{(t)} - \BellOp_\ValFunc [\valuefunc^{(t)}, \advFunc^{(t + 1)}] \big)}{\ValBar - \valfun^{(t)} }\\
   + 2 \frac{\learningrate}{\stepsize} \statinprod{\big( \valuebar - \BellOp_\valfun [\valuebar, \advFuncBar] \big) - \big(\valuefunc^{(t)} - \BellOp_\ValFunc [\valuefunc^{(t)}, \advFunc^{(t + 1)}] \big)}{\ValFunc^{(t + 1)} - \valfun^{(t)} }.\label{eq:value-function-recursion-population-level}
\end{multline}
Applying \Cref{lemma:bellman-positive-definite,lemma:bellman-bounded} to the terms in \Cref{eq:value-function-recursion-population-level}, note that
\begin{multline*}
 \frac{1}{\stepsize} \statinprod{\big( \valuebar - \BellOp_\valfun [\valuebar, \advFuncBar] \big) - \big(\valuefunc^{(t)} - \BellOp_\ValFunc [\valuefunc^{(t)}, \advFunc^{(t + 1)}] \big)}{\ValBar - \valfun^{(t)} } \\
 \geq \frac{\discount}{4} \statnorm{\valfun^{(t)} - \ValBar}^2 + \frac{\lammin}{2} \sobonorm{ \valfun^{(t)} - \ValBar}^2 - \constcmp \Statnorm{\advFunc^{(t + 1)} - \advFuncBar} \cdot \statnorm{\valfun^{(t)} - \ValBar} - c \stepsize,
\end{multline*}
and
\begin{multline*}
    \frac{1}{\stepsize} \statinprod{\big( \valuebar - \BellOp_\valfun [\valuebar, \advFuncBar] \big) - \big(\valuefunc^{(t)} - \BellOp_\ValFunc [\valuefunc^{(t)}, \advFunc^{(t + 1)}] \big)}{\ValFunc^{(t + 1)} - \valfun^{(t)} }\\
    \leq \lammax \sobonorm{\valfun^{(t)} - \ValBar} \sobonorm{\ValFunc^{(t + 1)} - \valfun^{(t)}} 
    + c \ctwofour \statnorm{\valfun^{(t)} - \ValBar} \sobonorm{\ValFunc^{(t + 1)} - \valfun^{(t)}} \\
    + \constcmp \statnorm{\valfun^{(t)} - \ValBar} \Statnorm{\advFunc^{(t + 1)} - \advFuncBar} + c \stepsize.
\end{multline*}
Combining these bounds with \Cref{eq:value-function-recursion-population-level} yields
\begin{align*}
  &\sobonorm{\valfun^{(t + 1)} - \ValBar}^2  + \sobonorm{\valfun^{(t + 1)} - \valfun^{(t)}}^2 \\
  &\leq \big(1 - \lammin \learningrate) \sobonorm{ \valfun^{(t)} - \ValBar}^2 - \frac{\discount \learningrate}{4} \statnorm{\valfun^{(t)} - \ValBar}^2 + 4 \learningrate \constcmp \Statnorm{\advFunc^{(t + 1)} - \advFuncBar} \cdot \statnorm{\valfun^{(t)} - \ValBar} + 4 c \learningrate \stepsize  \\
  &\quad + 2 \learningrate \Big(\lammax \sobonorm{\valfun^{(t)} - \ValBar} \sobonorm{\ValFunc^{(t + 1)} - \valfun^{(t)}} + c\ctwofour \statnorm{\valfun^{(t)} - \ValBar}  \sobonorm{\ValFunc^{(t + 1)} - \valfun^{(t)}} \Big).
\end{align*}
By Young's inequality, we have
\begin{align*}
   2 \learningrate \lammax \sobonorm{\valfun^{(t)} - \ValBar} \sobonorm{\ValFunc^{(t + 1)} - \valfun^{(t)}} - \frac{\sobonorm{\ValFunc^{(t + 1)} - \valfun^{(t)}}^2}{3} &\leq 3 \learningrate^2 \lammax^2 \sobonorm{\valfun^{(t)} - \ValBar}^2,\\
    2 c \learningrate \ctwofour \statnorm{\valfun^{(t)} - \ValBar}  \sobonorm{\ValFunc^{(t + 1)} - \valfun^{(t)}} - \frac{ \sobonorm{\ValFunc^{(t + 1)} - \valfun^{(t)}}^2}{3} &\leq 3 \learningrate^2 c^2 \ctwofour^2 \statnorm{\valfun^{(t)} - \ValBar}^2.
\end{align*}
Combining these pieces, when the learning rate $\learningrate$ satisfies
\begin{align*}
  \learningrate \leq \min \Big\{ \frac{\lammin}{12 \lammax^2}, \frac{\discount}{24 c^2 \ctwofour^2} \Big\},
\end{align*}
we have
\begin{multline*}
  \sobonorm{\valfun^{(t + 1)} - \ValBar}^2 + \frac{\sobonorm{\valfun^{(t + 1)} - \valfun^{(t)}}^2}{3} \\
  \leq \big(1 - \frac{\lammin \learningrate}{2} \big) \sobonorm{ \valfun^{(t)} - \ValBar}^2 - \frac{\discount}{8} \learningrate \statnorm{\valfun^{(t)} - \ValBar}^2 + 4 \learningrate \constcmp \Statnorm{\advFunc^{(t + 1)} - \advFuncBar} \cdot \statnorm{\valfun^{(t)} - \ValBar} + 5 c \learningrate \stepsize,
\end{multline*}
which concludes the proof.

\subsubsection{Proof of \Cref{lemma:advantage-function-contraction}}\label{subsubsec:proof-lemma-advantage-function-contraction}

Similarly, by the optimality condition of the proximal operator in the updates~\eqref{eq:iterative-scheme-advantage}, we can derive the first-order optimality condition
\begin{align*}
  \Statinprod{\BellOp_\advFunc [\valuefunc^{(t)}, \advFunc^{(t)}] - \advFunc^{(t + 1)}}{\advFuncBar - \advFunc^{(t + 1)}} \leq 0.
\end{align*}
By the projected fixed-point condition~\eqref{eq:projected-fixed-point}, we have
\begin{align*}
   \Statinprod{\advFuncBar - \BellOp_\advFunc [\valuebar, \advFuncBar]}{\advFuncBar - \advFunc^{(t + 1)}} \leq 0.
\end{align*}
Combining the two inequalities yields
\begin{align}
  \Statnorm{\advFuncBar - \advFunc^{(t + 1)}}^2 \leq \Statinprod{\advFuncBar - \advFunc^{(t + 1)}}{  \BellOp_\advFunc [\valuebar, \advFuncBar]  -   \BellOp_\advFunc [\valuefunc^{(t)}, \advFunc^{(t)}] \big)}.\label{eq:advantage-function-recursion-population-level}
\end{align}
Applying \Cref{lemma:semigroup-difference-bound} to the term in \Cref{eq:advantage-function-recursion-population-level}, we have
\begin{align*}
  \Statinprod{\advFuncBar - \advFunc^{(t + 1)}}{\BellOp_\advFunc [\valuebar, \advFuncBar] - \BellOp_\advFunc [\valuefunc^{(t)}, \advFunc^{(t)}] } \leq 
  c \ctwofour \sobonorm{\valuefunc^{(t)} - \ValBar} \cdot \Statnorm{\advFunc^{(t + 1)} - \advFuncBar}  + c_2 \stepsize.
\end{align*}
Substituting this bound into \Cref{eq:advantage-function-recursion-population-level} yields
\begin{align*}
  \Statnorm{\advFunc^{(t + 1)} - \advFuncBar}^2 \leq c \ctwofour \sobonorm{\valuefunc^{(t)} - \ValBar}  \Statnorm{\advFunc^{(t + 1)} - \advFuncBar} + c_2  \stepsize.  
\end{align*}
Noting that $\stepsize < 1/2$. Some algebra yields
\begin{align*}
  \Statnorm{\advFunc^{(t + 1)} - \advFuncBar} \leq c \sqrt{\ctwofour} \sobonorm{\valuefunc^{(t)} - \ValBar} + c' \stepsize^{1/2},
\end{align*}
which concludes the proof.

\subsection{Proof of \Cref{thm:main}}\label{subsec:proof-thm-main}
The proof involves bounding the statistical errors between the empirical updates and the population-level updates. To proceed, let us first define some notations.

the following empirical process suprema for a given radius $r > 0$:
\begin{align*}
  \ErrTermSqVal (r) &\mydefn \sup_{f \in \Fclass_\valfun \cap \ball_{\soboone} (r)}\abss{ \empsoboinprod{f}{f}^2 - \sobonorm{f}^2 },\\
  \ErrTermMainVal (r; \valuefunc, \advFunc) &\mydefn \frac{1}{\stepsize} \sup_{f \in \Fclass_\valfun \cap \ball_{\soboone} (r)} \abss{ \empbilinear{\valfun}{f}{\advFunc} - \statinprod{\valfun - \BellOp_\valfun [\valfun, \advFunc]}{f} }.
\end{align*}

The advantage function update is performing a least-square regression in the function class $\Fclass_\advFunc$. We characterize its statistical error in the following lemma.
\begin{lemma}\label{lemma:advantage-function-regression-error}
  Under the setup of \Cref{thm:main}, for any fixed value function $\valuefunc \in \Fclass_\valfun \cup \{\valueaux\}$ and $\advFunc \in \Fclass_\advFunc \cup \{\advFuncaux\}$, with probability $1 - \delta$, we have
  \begin{align*}
    \Statnorm{\empbelladv (\valuefunc, \advFunc) - \projectto{\Fclass_\advFunc}^{\Stationary} \circ \BellOp_\advFunc [\valuefunc, \advFunc]} \leq r^*_\advFunc.
  \end{align*}
\end{lemma}
\noindent See~\Cref{subsubsec:proof-lemma-advantage-function-regression-error} for the proof of this lemma.

Additionally, we need uniform concentration bounds for the empirical Sobolev norm and the empirical Bellman operator. These results are summarized in the following two lemmas.
\begin{lemma}\label{lemma:uniform-concentration-sobolev-norm}
  Under the setup of \Cref{thm:main}, with probability at least $1 - \delta$, we have
  \begin{align*}
    \ErrTermSqVal (r) \leq c \dudley_2 (\Fclass_\valfun (r), \soboone) \frac{\log (n / \delta)}{\sqrt{\numobs}} + c \dudley_1 (\Fclass_\valfun, C^1) \frac{\log^3 (\numobs / \delta)}{\numobs},
  \end{align*}
  for a constant $c > 0$ depending on the problem parameters.
\end{lemma}
\noindent See~\Cref{subsubsec:proof-lemma-uniform-concentration-sobolev-norm} for the proof of this lemma.

\begin{lemma}\label{lemma:uniform-concentration-bellman-operator}
  Under the setup of \Cref{thm:main}, given $\valfun \in \Fclass_\valfun$ and $\advFunc \in \Fclass_\advFunc$ fixed, with probability at least $1 - \delta$, we have
  \begin{align*}
    \ErrTermMainVal (r; \valuefunc, \advFunc) \leq c \dudley_2 (\Fclass_\valfun (r), \stationary) \frac{\log^3 (n / \delta)}{\sqrt{\numobs}} + c \dudley_1 (\Fclass_\valfun, \vecnorm{\cdot}{\infty}) \frac{\log^4 (\numobs / \delta)}{\numobs},
  \end{align*}
\end{lemma}
\noindent See~\Cref{subsubsec:proof-lemma-uniform-concentration-bellman-operator} for the proof of this lemma.

Underlying the proof of these lemmas is the following general technical tool that bounds the empirical process suprema using Bernstein-type inequalities and chaining arguments. We state the result here for completeness.
\begin{lemma}\label{lemma:empirical-process-tool}
  Given a function class $\funcClass$ and $\mathrm{i.i.d.}$ processes $(\varepsilon_i (f))_{f \in \funcClass}$ satisfying $\Exs[\varepsilon_i (f)] = 0$ for all $f \in \funcClass$. Suppose that we have
  \begin{align*}
    \Exs \big[ |\varepsilon_i (f_1) - \varepsilon_i (f_2)|^2 \big] \leq d_1 (f_1, f_2)^2 , \quad \vecnorm{\varepsilon_i (f_1) - \varepsilon_i (f_2)}{\psi_1} \leq d_2 (f_1, f_2),
  \end{align*}
  for any $f_1, f_2 \in \funcClass$ and some semi-metrics $d_1, d_2$ on $\funcClass$. Assume furthermore that $0 \in \funcClass$ and $\varepsilon_i (0) \equiv 0$. Then for any $\delta \in (0, 1)$, with probability $1 - \delta$, we have
  \begin{align*}
  \sup_{f \in \funcClass} \abss{ \frac{1}{\numobs} \sum_{i = 1}^\numobs \varepsilon_i (f) } \leq c \frac{\dudley_2 (\funcClass, d_1) + \diameter_{d_1} (\funcClass)\sqrt{\log (1/ \delta)}}{\sqrt{n}} + c \log \numobs  \frac{\dudley_1 (\funcClass, d_2) + \diameter_{d_2} (\funcClass)\log (1 / \delta)}{n},
\end{align*}
  for a universal constant $c > 0$.
\end{lemma}
\noindent See~\Cref{subsubsec:proof-lemma-empirical-process-tool} for the proof of this lemma.

Taking these lemmas as given for the moment, we now complete the proof of \Cref{thm:main}.
By the update scheme~\eqref{eq:empirical-iterative-scheme-value}, we have the first-order optimality condition
\begin{align*}
  \empsoboinprod{\valuehat^{(t + 1)} - \valuehat^{(t)}}{\ValBar - \valuehat^{(t + 1)}} - \frac{\learningrate}{\stepsize} \empbilinear{\valuehat^{(t)}}{\ValBar - \valuehat^{(t + 1)}}{\advFuncHat^{(t + 1)}} \leq 0.
\end{align*}
Applying polarization identity to the empirical Sobolev inner product yields
\begin{multline*}
  \empsoboinprod{\valuehat^{(t + 1)} - \ValBar}{\valuehat^{(t + 1)} - \ValBar} - \empsoboinprod{\valuehat^{(t)} - \ValBar}{\valuehat^{(t)} - \ValBar} + \empsoboinprod{\valuehat^{(t + 1)} - \valuehat^{(t)}}{\valuehat^{(t + 1)} - \valuehat^{(t)}} \\
  \leq 2 \frac{\learningrate}{\stepsize} \empbilinear{\valuehat^{(t)}}{\valuebar - \valuehat^{(t)}}{\advFuncHat^{(t + 1)}} + 2 \frac{\learningrate}{\stepsize} \empbilinear{\valuehat^{(t)}}{\valuehat^{(t + 1)} - \valuehat^{(t)}}{\advFuncHat^{(t + 1)}}.
\end{multline*}
Introducing the empirical process suprema terms $\ErrTermSqVal$ and $\ErrTermMainVal$, we can extract the population-level counterparts to obtain
\begin{align*}
  &\sobonorm{\valuehat^{(t + 1)} - \ValBar}^2 - \sobonorm{\valuehat^{(t)} - \ValBar}^2 + \sobonorm{\valuehat^{(t + 1)} - \valuehat^{(t)}}^2 \\
 & \leq 2 \frac{\learningrate}{\stepsize} \statinprod{\valuehat^{(t)} - \BellOp_\valfun [\valuehat^{(t)}, \advFuncHat^{(t + 1)}]}{\valuebar - \valuehat^{(t + 1)}} + 2 \learningrate \ErrTermMainVal (\sobonorm{\valuehat^{(t + 1)} - \ValBar}; \valuehat^{(t)}, \advFuncHat^{(t + 1)}) \\
 & \qquad \qquad + \ErrTermSqVal (\sobonorm{\valuehat^{(t + 1)} - \ValBar}) +  \ErrTermSqVal (\sobonorm{\valuehat^{(t)} - \ValBar}) +  \ErrTermSqVal (\sobonorm{\valuehat^{(t + 1)} - \valuehat^{(t)}}).
\end{align*}
By the first-order condition of the projected fixed-point condition~\eqref{eq:projected-fixed-point} for $\valuebar$, since $\valuehat^{(t + 1)} \in \Fclass_\valfun$, we have
\begin{align*}
   \statinprod{\valuebar - \BellOp_\valfun [\valuebar, \advFuncBar]}{\valuebar - \valuehat^{(t + 1)}} \leq 0.
\end{align*}
Combining with the inequality above, let $\Delhat^{(t)} \mydefn \valuehat^{(t)} - \ValBar$ for notation simplicity, we have
\begin{align*}
  &\sobonorm{\Delhat^{(t + 1)}}^2 - \sobonorm{\Delhat^{(t)}}^2 + \sobonorm{\Delhat^{(t + 1)} - \Delhat^{(t)}}^2 \\
 & \leq - 2 \frac{\learningrate}{\stepsize} \statinprod{ \big( \valuebar - \BellOp_\valfun [\valuebar, \advFuncBar] \big) - \big(\valuehat^{(t)} - \BellOp_\valfun [\valuehat^{(t)}, \advFuncHat^{(t + 1)}] \big)}{\valuebar - \valuehat^{(t)}} \\
 &\qquad+ 2 \frac{\learningrate}{\stepsize} \statinprod{ \big( \valuebar - \BellOp_\valfun [\valuebar, \advFuncBar] \big) - \big(\valuehat^{(t)} - \BellOp_\valfun [\valuehat^{(t)}, \advFuncHat^{(t + 1)}] \big)}{\valuehat^{(t + 1)} - \valuehat^{(t)}} \\
 &\qquad \qquad + 2 \learningrate \ErrTermMainVal (\sobonorm{\Delhat^{(t + 1)}}; \valuehat^{(t)}, \advFuncHat^{(t + 1)}) + 3\ErrTermSqVal (\sobonorm{\Delhat^{(t + 1)}}) + 3 \ErrTermSqVal (\sobonorm{\Delhat^{(t)}}) 
\end{align*}
where we have used the the fact that
\begin{align*}
  \sobonorm{\valuehat^{(t + 1)} - \valuehat^{(t)}} \leq \sobonorm{\Delhat^{(t + 1)}} + \sobonorm{\Delhat^{(t)}} \leq 2 \max \{ \sobonorm{\Delhat^{(t + 1)}}, \sobonorm{\Delhat^{(t)}} \}.
\end{align*}
and consequently,
\begin{align*}
  \ErrTermSqVal (\sobonorm{\valuehat^{(t + 1)} - \valuehat^{(t)}}) \leq 2 \max \{ \ErrTermSqVal (\sobonorm{\Delhat^{(t + 1)}}), \ErrTermSqVal (\sobonorm{\Delhat^{(t)}}) \}.
\end{align*}
Following the proof of \Cref{lemma:value-function-contraction} in \Cref{subsubsec:proof-lemma-value-function-contraction}, we can apply \Cref{lemma:bellman-positive-definite,lemma:bellman-bounded} to the first two terms on the right-hand side to obtain
\begin{align}
  \sobonorm{\Delhat^{(t + 1)}}^2
   &\leq \big(1 - \lammin \learningrate \big) \sobonorm{\Delhat^{(t)}}^2 - \frac{\discount \learningrate}{4} \statnorm{\Delhat^{(t)}}^2 
  + 4 \learningrate \constcmp \Statnorm{\advFuncHat^{(t + 1)} - \advFuncBar} \cdot \statnorm{\Delhat^{(t)}} + 4 c \learningrate \stepsize \nonumber \\
  &\quad + 2 \learningrate \ErrTermMainVal (\sobonorm{\Delhat^{(t + 1)}}; \valuehat^{(t)}, \advFuncHat^{(t + 1)}) + 3\ErrTermSqVal (\sobonorm{\Delhat^{(t + 1)}}) + 3 \ErrTermSqVal (\sobonorm{\Delhat^{(t)}}).\label{eq:value-function-recursion-empirical-level}
\end{align}
On the other hand, invoking \Cref{lemma:advantage-function-regression-error}, conditionally on the past iterates, with probability $1 - \delta$, we have
\begin{align}
  \Statnorm{\advFuncHat^{(t + 1)} - \projectto{\Fclass_\advFunc}^{\Stationary} \circ \BellOp_\advFunc \big[ \valuehat^{(t)}, \advFunc^{(t)} \big] } \leq c r^*_\advFunc.\label{eq:advantage-function-regression-error-application}
\end{align}
In order to relate this local regression error bound to the global error $\Statnorm{\advFuncHat^{(t + 1)} - \advFuncBar}$, we invoke \Cref{lemma:semigroup-difference-bound} with the test function $\projectto{\Fclass_\advFunc}^{\Stationary} \circ \BellOp_\advFunc \big[ \valuehat^{(t)}, \advFunc^{(t)} \big] - \advFuncBar$, which yields
\begin{align*}
  \Statnorm{\BellOp_\advFunc \big[ \valuehat^{(t)}, \advFunc^{(t)} \big]  - \advFuncBar} = \Statnorm{\BellOp_\advFunc \big[ \valuehat^{(t)}, \advFunc^{(t)} \big] - \BellOp_\advFunc [\valuebar, \advFuncBar]} \leq c  \sobonorm{\Delhat^{(t)}} + c' \stepsize^{1/2}.
\end{align*}
Combining this bound with \Cref{eq:advantage-function-regression-error-application} via the triangle inequality yields
\begin{align}
  \Statnorm{\advFuncHat^{(t + 1)} - \advFuncBar} \leq c \sobonorm{\Delhat^{(t)}} + c' \stepsize^{1/2} + c r^*_\advFunc.\label{eq:advantage-function-error-one-step}
\end{align}
Now we can substitute the bound~\eqref{eq:advantage-function-error-one-step} into \Cref{eq:value-function-recursion-empirical-level} to obtain
\begin{align*}
  \sobonorm{\Delhat^{(t + 1)}}^2
   &\leq \big(1 - \lammin \learningrate \big) \sobonorm{\Delhat^{(t)}}^2 - \frac{\discount \learningrate}{4} \statnorm{\Delhat^{(t)}}^2 \\
   &\quad + 4 c \learningrate \constcmp \big(\sobonorm{\Delhat^{(t)}} + \stepsize^{1/2} + r^*_\advFunc \big) \cdot \statnorm{\Delhat^{(t)}} + 4 c \learningrate \stepsize \nonumber \\
  &\quad + 2 \learningrate \ErrTermMainVal (\sobonorm{\Delhat^{(t + 1)}}; \valuehat^{(t)}, \advFuncHat^{(t + 1)}) + 3\ErrTermSqVal (\sobonorm{\Delhat^{(t + 1)}}) + 3 \ErrTermSqVal (\sobonorm{\Delhat^{(t)}}).
\end{align*}
Under the assumption~\eqref{eq:discount-rate-lower-bound} on the discount factor, by Young's inequality, we can absorb the third term on the right-hand side into the negative term involving $\statnorm{\Delhat^{(t)}}^2$. This leads to the recursion
\begin{align*}
  \sobonorm{\Delhat^{(t + 1)}}^2
   &\leq \big(1 - \frac{\lammin \learningrate}{2} \big) \sobonorm{\Delhat^{(t)}}^2 + c \learningrate \stepsize + c \learningrate (r^*_\advFunc)^2 \\
  &\quad + 2 \learningrate \ErrTermMainVal (\sobonorm{\Delhat^{(t + 1)}}; \valuehat^{(t)}, \advFuncHat^{(t + 1)}) + 3\ErrTermSqVal (\sobonorm{\Delhat^{(t + 1)}}) + 3 \ErrTermSqVal (\sobonorm{\Delhat^{(t)}}).
\end{align*}
To deal with the empirical process suprema terms, we note that by convexity of the set $\Fclass_\valfun$, we have $\valuehat^{(t)} \in \Fclass_\valfun$, the functions 
\begin{align*}
   r \mapsto \frac{\ErrTermMainVal (r; \valuehat^{(t)}, \advFuncHat^{(t + 1)})}{r}, \quad \mbox{and} \quad r \mapsto \frac{\ErrTermSqVal (r)}{r^2},
\end{align*}
are both non-increasing. Therefore, we have
\begin{align*}
  \ErrTermMainVal (\sobonorm{\Delhat^{(t + 1)}}; \valuehat^{(t)}, \advFuncHat^{(t + 1)}) \leq \begin{cases}
    \ErrTermMainVal (r^*_\valfun; \valuehat^{(t)}, \advFuncHat^{(t + 1)}) , & \mbox{if } \sobonorm{\Delhat^{(t + 1)}} < r^*_\valfun,\\
    \ErrTermMainVal (r^*_\valfun; \valuehat^{(t)}, \advFuncHat^{(t + 1)}) \cdot \frac{\sobonorm{\Delhat^{(t + 1)}}}{r^*_\valfun}, & \mbox{if } \sobonorm{\Delhat^{(t + 1)}} \geq r^*_\valfun,
  \end{cases}
\end{align*}
We therefore have the bound
\begin{align*}
  \ErrTermMainVal (\sobonorm{\Delhat^{(t + 1)}}; \valuehat^{(t)}, \advFuncHat^{(t + 1)}) &\leq \ErrTermMainVal (r^*_\valfun; \valuehat^{(t)}, \advFuncHat^{(t + 1)}) \cdot  \Big\{ 1 + \frac{\sobonorm{\Delhat^{(t + 1)}}}{r^*_\valfun} \Big\}.
\end{align*}
Similarly, given a constant $c_0 > 0$ to be chosen later, we have
\begin{align*}
  \ErrTermSqVal (\sobonorm{\Delhat^{(t + 1)}}) &\leq \begin{cases}
    \ErrTermSqVal (c_0 r^*_\valfun) , & \mbox{if } \sobonorm{\Delhat^{(t + 1)}} < c_0 r^*_\valfun,\\
    \ErrTermSqVal (c_0 r^*_\valfun) \cdot \frac{\sobonorm{\Delhat^{(t + 1)}}^2}{(c_0 r^*_\valfun)^2}, & \mbox{if } \sobonorm{\Delhat^{(t + 1)}} \geq c_0 r^*_\valfun,
  \end{cases}
\end{align*}
which yields the bounds
\begin{align*}
  \ErrTermSqVal (\sobonorm{\Delhat^{(t + 1)}}) &\leq \ErrTermSqVal (c_0 r^*_\valfun) \cdot  \Big\{ 1 + \frac{\sobonorm{\Delhat^{(t + 1)}}^2}{(c_0 r^*_\valfun)^2} \Big\},\\
  \ErrTermSqVal (\sobonorm{\Delhat^{(t)}}) &\leq \ErrTermSqVal (c_0 r^*_\valfun) \cdot  \Big\{ 1 + \frac{\sobonorm{\Delhat^{(t)}}^2}{(c_0 r^*_\valfun)^2} \Big\}.
\end{align*}
Invoking \Cref{lemma:uniform-concentration-sobolev-norm,lemma:uniform-concentration-bellman-operator} to bound these empirical process suprema terms, with probability at least $1 - \delta$, we have
\begin{align*}
  \ErrTermMainVal (r^*_\valfun; \valuehat^{(t)}, \advFuncHat^{(t + 1)}) &\leq c \dudley_2 (\Fclass_\valfun (r^*_\valfun), \stationary) \frac{\log^3 (n / \delta)}{\sqrt{\numobs}} + c \dudley_1 (\Fclass_\valfun, \vecnorm{\cdot}{\infty}) \frac{\log^4 (\numobs / \delta)}{\numobs} \leq  c (r^*_\valfun)^2,\\
  \ErrTermSqVal (c_0 r^*_\valfun) &\leq c \dudley_2 (\Fclass_\valfun (c_0 r^*_\valfun), \soboone) \frac{\log (n / \delta)}{\sqrt{\numobs}} + c \dudley_1 (\Fclass_\valfun, C^1) \frac{\log^3 (\numobs / \delta)}{\numobs} \leq c c_0 (r^*_\valfun)^2,
\end{align*}
where we have used the definitions of the critical radii $r^*_\valfun$ and sub-linearity of the Dudley entropy integrals. On the event that these bounds hold, substituting them back into the recursion yields
\begin{align*}
  &\sobonorm{\Delhat^{(t + 1)}}^2
   \leq \big(1 - \frac{\lammin \learningrate}{2} \big) \sobonorm{\Delhat^{(t)}}^2 + c \learningrate \stepsize + c \learningrate (r^*_\advFunc)^2 \\
  &\qquad + 2 \learningrate c \Big\{ (r^*_\valfun)^2 + {r^*_\valfun} {\sobonorm{\Delhat^{(t + 1)}}} \Big\} + 3 c \Big\{ c_0 (r^*_\valfun)^2 + \frac{1}{c_0}\sobonorm{\Delhat^{(t + 1)}}^2 \Big\} + 3 c \Big\{ c_0 (r^*_\valfun)^2 + \frac{1}{c_0}\sobonorm{\Delhat^{(t)}}^2 \Big\}.
\end{align*}
Choose $c_0 \mydefn \frac{48 c}{\lammin \learningrate}$, we can absorb the terms $\frac{3c}{c_0} \sobonorm{\Delhat^{(t + 1)}}^2$ on the right-hand side into the left-hand side, and the term $\frac{3c}{c_0} \sobonorm{\Delhat^{(t)}}^2$ into the contraction term. We further apply Young's inequality to the cross term to obtain
\begin{align*}
  2 \learningrate c {r^*_\valfun} {\sobonorm{\Delhat^{(t + 1)}}} \leq \frac{\lammin \learningrate}{8} \sobonorm{\Delhat^{(t + 1)}}^2 + \frac{8 c^2}{\lammin} \learningrate (r^*_\valfun)^2.
\end{align*}
Combining these bounds yields the recursion
\begin{align*}
  \sobonorm{\Delhat^{(t + 1)}}^2
   &\leq \big(1 - \frac{\lammin \learningrate}{8} \big) \sobonorm{\Delhat^{(t)}}^2 + c \learningrate \stepsize + c \learningrate (r^*_\advFunc)^2 + \frac{c}{\learningrate} (r^*_\valfun)^2,
\end{align*}
which, conditionally on the previous iterates $(\valuehat^{(i)}, \advFuncHat^{(i)})_{i = 0}^t$, holds with probability at least $1 - \delta$. Taking union bound over all iterations $t = 0, 1, \ldots, N - 1$, we have that with probability at least $1 - N \delta$, the above recursion holds for all $t = 0, 1, \ldots, N - 1$. When $\delta < 1 / N$, the impact of the union bound is absorbed into the logarithmic factors in the definitions of the critical radii $(r^*_\valfun, r^*_\advFunc)$.

Therefore, solving the recursion leads to
\begin{align*}
  \sobonorm{\Delhat^{(N)}}^2 &\leq \exp \big(- \frac{\lammin \learningrate N}{8} \big) \sobonorm{\Delhat^{(0)}}^2 + c \stepsize + c (r^*_\advFunc)^2 + \frac{c}{\learningrate^2} (r^*_\valfun)^2,
\end{align*}
holding with probability at least $1 - \delta$.

Invoking \Cref{eq:advantage-function-error-one-step}, we also have the bound for the advantage function error
\begin{align*}
  \Statnorm{\advFuncHat^{(N)} - \advFuncBar} &\leq c \sobonorm{\Delhat^{(N)}} + c' \stepsize^{1/2} + c r^*_\advFunc,
\end{align*}
with probability at least $1 - \delta$.

Putting together these two bounds concludes the proof of \Cref{thm:main}.

\newcommand{\gbar}{\widebar{g}}
\newcommand{\ghat}{\widehat{g}}

\subsubsection{Proof of \Cref{lemma:advantage-function-regression-error}}\label{subsubsec:proof-lemma-advantage-function-regression-error}
To simplify the notation, we define
\begin{align*}
  g^* \mydefn \BellOp_\advFunc [\valuefunc, \advFunc], \quad \widebar{g} \mydefn \projectto{\Fclass_\advFunc}^{\Stationary} (g^*), \quad \widehat{g} \mydefn \empbelladv (\valuefunc, \advFunc).
\end{align*}
By the first-order optimality condition of the regression problem in the definition of $\empbelladv$, since $\widebar{g} \in \Fclass_\advFunc$, we have
\begin{multline*}
 \sum_{(i, k) \in \Dset} \Big\{  \randreward^{(i)}_{k \stepsize}  + \frac{e^{- \discount \stepsize}}{\stepsize} \big(\valfun (\State_{(k + 1) \stepsize}^{(i)}) - \valfun (\State_{k \stepsize}^{(i)}) \big) + e^{ - \discount \stepsize} \max_{\action' \in \actionspace} \advFunc(\State_{(k + 1) \stepsize}^{(i)}, \action') - \widehat{g} (\State_{k \stepsize}^{(i)}, \Action_{k \stepsize}^{(i)})  \Big\}\\
\times  \big( \widebar{g} - \widehat{g} \big) (\State_{k \stepsize}^{(i)}, \Action_{k \stepsize}^{(i)})
 \leq 0,
\end{multline*}
which leads to the basic inequality
\begin{multline*}
   \sum_{(i, k) \in \Dset} \big( \widehat{g} - \widebar{g} \big)^2 (\State_{k \stepsize}^{(i)}, \Action_{k \stepsize}^{(i)})\\
   \leq  \sum_{(i, k) \in \Dset} \Big\{  \randreward^{(i)}_{k \stepsize}  + \frac{e^{- \discount \stepsize}}{\stepsize} \big(\valfun (\State_{(k + 1) \stepsize}^{(i)}) - \valfun (\State_{k \stepsize}^{(i)}) \big)  + e^{ - \discount \stepsize} \max_{\action' \in \actionspace} \advFunc(\State_{(k + 1) \stepsize}^{(i)}, \action') - \widebar{g} (\State_{k \stepsize}^{(i)}, \Action_{k \stepsize}^{(i)})  \Big\} \\
   \times \big( \widehat{g} - \widebar{g}\big) (\State_{k \stepsize}^{(i)}, \Action_{k \stepsize}^{(i)}).
\end{multline*}
Define the random variables
\begin{align*}
  \xi_{i, k} \mydefn  \randreward^{(i)}_{k \stepsize}  + \frac{e^{- \discount \stepsize}}{\stepsize} \big(\valfun (\State_{(k + 1) \stepsize}^{(i)}) - \valfun (\State_{k \stepsize}^{(i)}) \big) + e^{ - \discount \stepsize} \max_{\action' \in \actionspace} \advFunc(\State_{(k + 1) \stepsize}^{(i)}, \action') - g^* (\State_{k \stepsize}^{(i)}, \Action_{k \stepsize}^{(i)}),
\end{align*}
for each $(i, k) \in \Dset$. The basic inequality can be rewritten as
\begin{align}
    \frac{1 - e^{- \discount \stepsize}}{\numobs} \sum_{(i, k) \in \Dset} \big( \widehat{g} - \widebar{g} \big)^2 (\State_{k \stepsize}^{(i)}, \Action_{k \stepsize}^{(i)}) \leq  \frac{1 - e^{- \discount \stepsize}}{\numobs} \sum_{(i, k) \in \Dset} \xi_{i, k} \cdot \big( \widehat{g} - \widebar{g}\big) (\State_{k \stepsize}^{(i)}, \Action_{k \stepsize}^{(i)}).\label{eq:advantage-regression-basic-inequality}
\end{align}
Define the empirical process suprema terms
\begin{align*}
    \ErrTermSqAdv (r) &\mydefn \sup_{f \in \Fclass_\advFunc (r)} \abss{ \frac{1 - e^{- \discount \stepsize}}{\numobs} \sum_{(i, k) \in \Dset} f^2 (\State_{k \stepsize}^{(i)}, \Action_{k \stepsize}^{(i)}) - \Statnorm{f}^2 },\\
    \ErrTermMainAdv (r) &\mydefn \sup_{f \in \Fclass_\advFunc (r)} \abss{\frac{1 - e^{- \discount \stepsize}}{\numobs}  \sum_{(i, k) \in \Dset} \xi_{i, k} f (\State_{k \stepsize}^{(i)}, \Action_{k \stepsize}^{(i)}) - \Exs \Big[ (1 - e^{- \discount \stepsize})  \sum_{k \geq 0} \xi_{i, k} f (\State_{k \stepsize}^{(i)}, \Action_{k \stepsize}^{(i)}) \Big] }.
\end{align*}

The following two lemmas characterize the behavior of these empirical process terms. They are analogous to \Cref{lemma:uniform-concentration-sobolev-norm,lemma:uniform-concentration-bellman-operator} respectively.
\begin{lemma}\label{lemma:uniform-concentration-norm-advantage}
  Under the setup of \Cref{thm:main}, for any fixed $r > 0$, with probability $1 - \delta$, we have
  \begin{align*}
    \ErrTermSqAdv (r) &\leq c \dudley_2 (\Fclass_\advFunc (r), \Stationary) \frac{\log (n / \delta)}{\sqrt{\numobs}} + c \dudley_1 (\Fclass_\advFunc, \mathbb{L}^\infty)  \frac{\log^3 (\numobs / \delta)}{n},
  \end{align*}
  for a constant $c > 0$ depending on the problem parameters.
\end{lemma}
\noindent See~\Cref{app:subsec-proof-lemma-uniform-concentration-norm-advantage} for the proof of this lemma.

\begin{lemma}\label{lemma:uniform-concentration-bellman-advantage}
  Under the setup of \Cref{thm:main}, for any fixed $r > 0$, with probability $1 - \delta$, we have
  \begin{align*}
    \ErrTermMainAdv (r) &\leq c \dudley_2 (\Fclass_\advFunc (r), \Stationary) \frac{\log^{3} (n / \delta)}{\sqrt{\numobs}} + c \dudley_1 (\Fclass_\advFunc, \mathbb{L}^\infty)  \frac{\log^4 (\numobs / \delta)}{n},
  \end{align*}
  for a constant $c > 0$ depending on the problem parameters.
\end{lemma}
\noindent See~\Cref{app:subsec-proof-lemma-uniform-concentration-bellman-advantage} for the proof of this lemma.

Taking these two lemmas as given, we proceed to complete the proof of \Cref{lemma:advantage-function-regression-error}.
To start with, we note that for any function $g \in \Fclass_\advFunc - \gbar$, we have
\begin{multline*}
  \Exs \Big[  (1 - e^{- \discount \stepsize}) \sum_{k \geq 0} \xi_{i, k} \cdot \big( g - \widebar{g}\big) (\State_{k \stepsize}^{(i)}, \Action_{k \stepsize}^{(i)}) \Big] \\
   = \Statinprod{\BellOp_\advFunc [\valfun, \advFunc] - \gbar }{g - \gbar} + \frac{e^{- \discount \stepsize}}{\stepsize} \Statinprod{ \semigroup_\stepsize^{\behavpolicy} \valfun - \valfun }{g - \gbar} = \Statinprod{\BellOp_\advFunc [\valfun, \advFunc] - \gbar }{g - \gbar}  \leq 0,
\end{multline*}
where in the last step we use the zero-mean property~\eqref{eq:advantage-mean-zero} of the class $\Fclass_\advFunc$, as well as the definition of the projection $\gbar$.

Consequently, the basic inequality~\eqref{eq:advantage-regression-basic-inequality} implies the self-bounding inequality
\begin{align*}
  \Statnorm{\ghat - \gbar}^2 &\leq \ErrTermMainAdv \big( \Statnorm{\ghat - \gbar} \big) + \ErrTermSqAdv \big( \Statnorm{\ghat - \gbar} \big) + \sup_{g \in \Fclass_\advFunc - \gbar}\Exs \Big[  (1 - e^{- \discount \stepsize}) \sum_{k \geq 0} \xi_{i, k} \cdot \big( g - \widebar{g}\big) (\State_{k \stepsize}^{(i)}, \Action_{k \stepsize}^{(i)}) \Big]\\
  &\leq \ErrTermMainAdv \big( \Statnorm{\ghat - \gbar} \big) + \ErrTermSqAdv \big( \Statnorm{\ghat - \gbar} \big).
\end{align*}
Since the set $\Fclass_\advFunc$ is convex, the functions
\begin{align*}
  r \mapsto \frac{\ErrTermMainAdv (r)}{r}, \quad \mbox{and} \quad r \mapsto \frac{\ErrTermSqAdv (r)}{r^2},
\end{align*}
are both non-increasing.

Given a constant $c_0 > 0$ to be specified later, on the event that $\Statnorm{\ghat - \gbar} \geq c_0 r^*_\advFunc$, we have
\begin{align*}
  \Statnorm{\ghat - \gbar}^2 &\leq  \ErrTermMainAdv \big( \Statnorm{\ghat - \gbar} \big) + \ErrTermSqAdv \big( \Statnorm{\ghat - \gbar} \big)\\
  &\leq \frac{\Statnorm{\ghat - \gbar}}{c_0 r^*_\advFunc} \ErrTermMainAdv (c_0 r^*_\advFunc) + \frac{\Statnorm{\ghat - \gbar}^2}{(c_0 r^*_\advFunc)^2} \ErrTermSqAdv (c_0 r^*_\advFunc).
\end{align*}
By \Cref{lemma:uniform-concentration-norm-advantage,lemma:uniform-concentration-bellman-advantage}, with probability at least $1 - \delta$, we have
\begin{align*}
  \ErrTermMainAdv (c_0 r^*_\advFunc) &\leq c \dudley_2 (\Fclass_\advFunc (c_0 r^*_\advFunc), \Stationary) \frac{\log (n / \delta)}{\sqrt{\numobs}} + c \dudley_1 (\Fclass_\advFunc, \mathbb{L}^\infty)  \frac{\log^3 (\numobs / \delta)}{n} \leq c c_0 (r^*_\advFunc)^2,\\
  \ErrTermSqAdv (c_0 r^*_\advFunc) &\leq c \dudley_2 (\Fclass_\advFunc (c_0 r^*_\advFunc), \Stationary) \frac{\log^{3} (n / \delta)}{\sqrt{\numobs}} + c \dudley_1 (\Fclass_\advFunc, \mathbb{L}^\infty)  \frac{\log^4 (\numobs / \delta)}{n} \leq c c_0 (r^*_\advFunc)^2.
\end{align*}
On the event that both inequalities hold but $\Statnorm{\ghat - \gbar} \geq c_0 r^*_\advFunc$, we have
\begin{align*}
  \Statnorm{\ghat - \gbar}^2 &\leq c \Statnorm{\ghat - \gbar} r^*_\advFunc + \frac{c}{c_0} \Statnorm{\ghat - \gbar}^2.
\end{align*}
Taking $c_0 = 3 c$ yields a contradiction, which implies that with probability at least $1 - \delta$, we have
\begin{align*}
  \Statnorm{\ghat - \gbar} \leq c_0 r^*_\advFunc,
\end{align*}
which concludes the proof.

\subsubsection{Proof of \Cref{lemma:uniform-concentration-sobolev-norm}}\label{subsubsec:proof-lemma-uniform-concentration-sobolev-norm}
Define the random functionals
\begin{align*}
  \zeta_i (f) \mydefn (1 - e^{- \discount \stepsize}) \sum_{k = 0}^{\lfloor T_i / \stepsize \rfloor} \Big\{ f^2 (\State_{k \stepsize}^{(i)}) + \eucnorm{\nabla f (\State_{k \stepsize}^{(i)})}^2 \Big\} , \quad \mbox{for } i = 1, 2, \ldots, \numobs,
\end{align*}
and we also define $\widetilde{\zeta}_i (f) \mydefn \zeta_i \bm{1}_{T_i \leq \Tmax}$ for some fixed $\Tmax > 0$ to be specified later.

Given $f$, $\{\zeta_i (f)\}_{i = 1}^\numobs$ are $\mathrm{i.i.d.}$ random variables with mean $\sobonorm{f}^2$. The empirical process supremum $\ErrTermSqVal (r)$ can be rewritten as
\begin{align*}
  \ErrTermSqVal (r) = \sup_{f \in \Fclass_\valfun \cap \ball_{\soboone} (r)} \abss{ \frac{1}{\numobs} \sum_{i = 1}^{\numobs} \zeta_i (f) - \Exs[\zeta_i (f)] }.
\end{align*}
To control this term, we first condition on the event $\Event \mydefn \{ T_i \leq \Tmax, \mbox{ for all } i = 1, 2, \ldots, \numobs \}$. Under this event, we have $\zeta_i (f) = \widetilde{\zeta}_i (f)$ for all $i = 1, 2, \ldots, \numobs$. Since $T_i \sim \mathrm{Exp} (\discount)$, by tail bound of exponential distribution and union bound, by taking $\Tmax = c \discount^{-1}\log (\numobs / \delta)$, we have
\begin{align*}
  \Prob (\Event^c) \leq \delta / 2, \quad \abss{\Exs [\zeta_i (f) ]- \Exs [\widetilde{\zeta}_i (f)]} \leq \delta / \numobs^2.
\end{align*}
Thus, it suffices to control the empirical process supremum for the truncated functionals. By \Cref{lemma:empirical-process-tool}, we need to verify the Bernstein-type conditions for the processes $\widetilde{\zeta}_i (f)$. For any pair $f_1, f_2 \in \Fclass_\valfun$, we have
\begin{align*}
  &\Exs \big[ |\widetilde{\zeta}_i (f_1) - \widetilde{\zeta}_i (f_2)|^2 \big] 
  \\
  &\leq 2 (\stepsize \discount)^2 \Exs \Big[ \Big( \sum_{k = 0}^{\lfloor T_i / \stepsize \rfloor}  f_1^2 (\State_{k \stepsize}^{(i)}) + \eucnorm{\nabla f_1 (\State_{k \stepsize}^{(i)})}^2 - f_2^2 (\State_{k \stepsize}^{(i)}) - \eucnorm{\nabla f_2 (\State_{k \stepsize}^{(i)})}^2  \Big)^2 \bm{1}_{T_i \leq \Tmax} \Big]\\
  &\leq 2 (\stepsize \discount)^2  \Tmax  \Exs \Big[  \sum_{k = 0}^{\lfloor T_i / \stepsize \rfloor}  \Big\{ f_1^2 (\State_{k \stepsize}^{(i)}) + \eucnorm{\nabla f_1 (\State_{k \stepsize}^{(i)})}^2 - f_2^2 (\State_{k \stepsize}^{(i)}) - \eucnorm{\nabla f_2 (\State_{k \stepsize}^{(i)})}^2 \Big\}^2  \Big]\\
  &\leq 8 (\stepsize \discount)^2  \Tmax \Exs \Big[ \sum_{k = 0}^{\lfloor T_i / \stepsize \rfloor} \Big\{ \big( f_1 - f_2\big)^2 (\State_{k \stepsize}^{(i)})  \big( f_1+ f_2 \big)^2 (\State_{k \stepsize}^{(i)})  +  \big| \nabla (f_1  - f_2) (\State_{k \stepsize}^{(i)}) \big|^2 \big| \nabla (f_1 +  f_2) (\State_{k \stepsize}^{(i)}) \big|^2 \Big\}\Big]\\
  &\leq   8 (\stepsize \discount)^2  \Tmax \constFclass^2 \Exs \Big[ \sum_{k = 0}^{\lfloor T_i / \stepsize \rfloor} \Big\{ \big( f_1 - f_2\big)^2 (\State_{k \stepsize}^{(i)})  +  \big| \nabla (f_1  - f_2) (\State_{k \stepsize}^{(i)}) \big|^2  \Big\}\Big]\\
  &\leq 8 \constFclass^2 \log (n / \delta)\sobonorm{f_1 - f_2}^2.
\end{align*}
Additionally, we have
\begin{align*}
  |\widetilde{\zeta}_i (f_1) - \widetilde{\zeta}_i (f_2)|
  &\leq \sum_{k = 0}^{\Tmax}  \abss{(f_1 - f_2) (\State_{k \stepsize}^{(i)})}  \abss{(f_1 + f_2) (\State_{k \stepsize}^{(i)})} + \eucnorm{\nabla (f_1 - f_2) (\State_{k \stepsize}^{(i)})}  \eucnorm{\nabla (f_1 + f_2) (\State_{k \stepsize}^{(i)})}\\
  &\leq 2 \constFclass \log (n / \delta) \vecnorm
  {f_1 - f_2}{C^1}.
\end{align*}
Applying \Cref{lemma:empirical-process-tool} with the above two bounds, and combining the probability of the complement event $\Event^c$, we conclude the inequality with probability at least $1 - \delta$,
\begin{align*}
  \ErrTermSqVal (r) &\leq c \Big\{\dudley_2 (\Fclass_\valfun (r), \soboone) + r \sqrt{\log (1/ \delta)} \Big\} \sqrt{\frac{\log (n / \delta)}{n}} + c \log^2 (\numobs / \delta) \frac{\dudley_1 (\Fclass_\valfun, C^1) + \diameter_{C_1} (\Fclass_\valfun) \log (1 / \delta)}{n}\\
&\leq c \dudley_2 (\Fclass_\valfun (r), \soboone) \frac{\log (n / \delta)}{\sqrt{\numobs}} + c \dudley_1 (\Fclass_\valfun, C^1) \frac{\log^3 (\numobs / \delta)}{\numobs},
\end{align*}
which concludes the proof.

\subsubsection{Proof of \Cref{lemma:uniform-concentration-bellman-operator}}\label{subsubsec:proof-lemma-uniform-concentration-bellman-operator}
Define the random functionals
\begin{align*}
  \zeta_i (f) \mydefn \frac{1 - e^{- \discount \stepsize}}{\stepsize} \sum_{k = 0}^{\lfloor T_i / \stepsize \rfloor} \Big\{ \valfun (\State_{k \stepsize}^{(i)}) - \stepsize \randreward^{(i)}_{k \stepsize} - e^{- \discount \stepsize} \valfun (\State_{(k + 1) \stepsize}^{(i)}) - \stepsize e^{- \discount \stepsize} \max_{\action' \in \actionspace} \advFunc (\State_{(k + 1) \stepsize}^{(i)}, \action')  \Big\} f (\State_{k \stepsize}^{(i)}),
\end{align*}
and we also define $\widetilde{\zeta}_i (f) \mydefn \zeta_i \bm{1}_{T_i \leq \Tmax}$ for $\Tmax = c \discount^{-1} \log (\numobs / \delta)$. Similar to the proof of \Cref{lemma:uniform-concentration-sobolev-norm}, for the event $\Event \mydefn \{ T_i \leq \Tmax, \mbox{ for all } i = 1, 2, \ldots, \numobs \}$, we have $\Prob (\Event^c) \leq \delta / 2$. Furthermore, we note that
\begin{align*}
  &\abss{\Exs [\zeta_i (f)] - \Exs [\widetilde{\zeta}_i (f)]}\\
  &\leq \discount \Exs \Big[ \sum_{\lfloor \Tmax / \stepsize \rfloor + 1}^{\lfloor T_i / \stepsize \rfloor} \Big\{ \valfun (\State_{k \stepsize}^{(i)}) - \stepsize \randreward^{(i)}_{k \stepsize} - e^{- \discount \stepsize} \valfun (\State_{(k + 1) \stepsize}^{(i)}) - \stepsize e^{- \discount \stepsize} \max_{\action' \in \actionspace} \advFunc (\State_{(k + 1) \stepsize}^{(i)}, \action')  \Big\} f (\State_{k \stepsize}^{(i)}) \bm{1}_{T_i > \Tmax} \Big]\\
  &\leq \frac{4 \discount \constFclass^2 }{\stepsize} \Exs \big[( T_i - \Tmax) \bm{1}_{T_i > \Tmax} \big] \leq \frac{\delta}{\numobs^2},
\end{align*}
whenever we choose $\delta \leq \stepsize$. So it suffices to control the empirical process supremum for the truncated functionals. Similar to the proof of \Cref{lemma:uniform-concentration-sobolev-norm}, we bound the variance and the Orlicz norm of the functional. For any pair $f_1, f_2 \in \Fclass_\valfun$, by Young's inequality, we can decompose the variance as
\begin{align*}
  &\Exs \big[ |\widetilde{\zeta}_i (f_1) - \widetilde{\zeta}_i (f_2)|^2 \big]\\
  &\leq 2\discount^2 \Exs \Big[ \Big\{\sum_{k = 0}^{\lfloor T_i / \stepsize \rfloor} \Big\{\stepsize \randreward^{(i)}_{k \stepsize} + \stepsize e^{- \discount \stepsize} \max_{\action' \in \actionspace} \advFunc (\State_{(k + 1) \stepsize}^{(i)}, \action') \Big\} (f_1 - f_2) (\State_{k \stepsize}^{(i)}) \Big\}^2 \Big]\\
   &\qquad+  2\discount^2 \Exs \Big[ \Big\{\sum_{k = 0}^{\lfloor T_i / \stepsize \rfloor} \Big\{\valfun (\State_{k \stepsize}^{(i)})  - e^{- \discount \stepsize} \valfun (\State_{(k + 1) \stepsize}^{(i)})  \Big\}(f_1 - f_2) (\State_{k \stepsize}^{(i)}) \Big\}^2 \Big] 
\end{align*}
For the first term, we apply Cauchy--Schwarz inequality to obtain
\begin{align*}
   &2\discount^2 \Exs \Big[ \Big\{\sum_{k = 0}^{\lfloor T_i / \stepsize \rfloor} \Big\{\stepsize \randreward^{(i)}_{k \stepsize} + \stepsize e^{- \discount \stepsize} \max_{\action' \in \actionspace} \advFunc (\State_{(k + 1) \stepsize}^{(i)}, \action') \Big\} (f_1 - f_2) (\State_{k \stepsize}^{(i)}) \Big\}^2 \Big]\\
  &\leq \frac{\discount^2 \Tmax}{\stepsize} \Exs \Big[ \sum_{k = 0}^{\lfloor T_i / \stepsize \rfloor} \Big\{  \stepsize \randreward^{(i)}_{k \stepsize} + \stepsize e^{- \discount \stepsize} \max_{\action' \in \actionspace} \advFunc (\State_{(k + 1) \stepsize}^{(i)}, \action')  \Big\}^2 (f_1 - f_2)^2 (\State_{k \stepsize}^{(i)}) \Big]\\
  &\leq  {2 \discount^2 \Tmax}{\stepsize} \sum_{k = 0}^{+ \infty}  \Exs \Big[  \big(\randreward^{(i)}_{k \stepsize} + e^{- \discount \stepsize} \max_{\action' \in \actionspace} \advFunc (\State_{(k + 1) \stepsize}^{(i)}, \action') \big)^2 (f_1 - f_2)^2 (\State_{k \stepsize}^{(i)}) \bm{1}_{T_i \geq k \stepsize} \Big].
\end{align*}
Note that by boundedness of the reward function and the advantage function in $\Fclass_\advFunc$, we have
\begin{multline*}
  {2 \discount^2 \Tmax}{\stepsize} \sum_{k = 0}^{+ \infty}   \Exs \Big[  \big(\randreward^{(i)}_{k \stepsize} + e^{- \discount \stepsize} \max_{\action' \in \actionspace} \advFunc (\State_{(k + 1) \stepsize}^{(i)}, \action') \big)^2 (f_1 - f_2)^2 (\State_{k \stepsize}^{(i)}) \bm{1}_{T_i \geq k \stepsize} \Big] \\
  \leq {2 \discount^2 \Tmax}{\stepsize} \sum_{k = 0}^{+ \infty}   c  \Exs \big[ (f_1 - f_2)^2 (\State_{k \stepsize}^{(i)}) \bm{1}_{T_i \geq k} \big] \leq \frac{2 \discount^2 \Tmax \stepsize c}{1 - e^{- \discount \stepsize}} \Statnorm{f_1 - f_2}^2 \leq c' \log (n / \delta) \Statnorm{f_1 - f_2}^2.
\end{multline*}
For the Orlicz norm, we can apply a similar decomposition
\begin{align*}
  \vecnorm{\widetilde{\zeta}_i (f_1) - \widetilde{\zeta}_i (f_2)}{\psi_1} &\leq 2 \discount \vecnorm{ \sum_{k = 0}^{\lfloor T_i / \stepsize \rfloor} \Big\{\stepsize \randreward^{(i)}_{k \stepsize} + \stepsize e^{- \discount \stepsize} \max_{\action' \in \actionspace} \advFunc (\State_{(k + 1) \stepsize}^{(i)}, \action') \Big\} (f_1 - f_2) (\State_{k \stepsize}^{(i)}) \bm_{T_i \leq \Tmax}}{\psi_1}\\
  &\qquad+ 2 \discount \vecnorm{ \sum_{k = 0}^{\lfloor T_i / \stepsize \rfloor} \Big\{\valfun (\State_{k \stepsize}^{(i)})  - e^{- \discount \stepsize} \valfun (\State_{(k + 1) \stepsize}^{(i)})  \Big\}(f_1 - f_2) (\State_{k \stepsize}^{(i)}) \bm{1}_{T_i \leq \Tmax}}{\psi_1}.
\end{align*}
The first term is controlled by the boundedness of the reward function and the advantage function in $\Fclass_\advFunc$. In particular, we have
\begin{multline*}
  \Big| 2 \discount \sum_{k = 0}^{\lfloor T_i / \stepsize \rfloor} \Big\{\stepsize \randreward^{(i)}_{k \stepsize} + \stepsize e^{- \discount \stepsize} \max_{\action' \in \actionspace} \advFunc (\State_{(k + 1) \stepsize}^{(i)}, \action') \Big\} (f_1 - f_2) (\State_{k \stepsize}^{(i)}) \bm_{T_i \leq \Tmax} \Big|\\
  \leq c \discount \stepsize \Tmax \constFclass \vecnorm{f_1 - f_2}{\infty} \leq c' \log (n / \delta)\vecnorm{f_1 - f_2}{\infty}.
\end{multline*}

It suffices to bound the variance and the Orlicz norm of the term involving differences in the value function. This is accomplished by the following lemma.
\begin{lemma}\label{lemma:bellman-variance-control-in-emp-proc-valfunc-proof}
  Under the setup of \Cref{lemma:uniform-concentration-bellman-operator}, for any function $f \in \Fclass_\valfun - \Fclass_\valfun$, we have
  \begin{align*}
    \Exs \Big[ \Big\{\sum_{k = 0}^{\lfloor T_i / \stepsize \rfloor} \Big(\valfun (\State_{k \stepsize}^{(i)})  - e^{- \discount \stepsize} \valfun (\State_{(k + 1) \stepsize}^{(i)})  \Big)  f (\State_{k \stepsize}^{(i)}) \Big\}^2 \bm{1}_{T_i \leq \Tmax} \Big]  &\leq c \log^3 (n / \delta) \statnorm{f}^2 + \frac{\vecnorm{f}{\infty}^2}{\numobs},\\
    \orlicznorm{\Big\{\sum_{k = 0}^{\lfloor T_i / \stepsize \rfloor} \Big(\valfun (\State_{k \stepsize}^{(i)})  - e^{- \discount \stepsize} \valfun (\State_{(k + 1) \stepsize}^{(i)})  \Big)  f (\State_{k \stepsize}^{(i)}) \Big\}^2 \bm{1}_{T_i \leq \Tmax}}{1} &\leq c' \log^2 (n / \delta) \vecnorm{f}{\infty}.
  \end{align*}
\end{lemma}
\noindent See~\Cref{app:subsec-proof-lemma-bellman-variance-control-in-emp-proc-valfunc-proof} for the proof of this lemma.

Combining \Cref{lemma:bellman-variance-control-in-emp-proc-valfunc-proof} with the previous bounds, we have verified the Bernstein-type conditions
\begin{align*}
  \Exs \big[ |\widetilde{\zeta}_i (f_1) - \widetilde{\zeta}_i (f_2)|^2 \big] &\leq c \log^3 (n / \delta) \Statnorm{f_1 - f_2}^2 + \frac{\vecnorm{f_1 - f_2}{\infty}^2}{\numobs},\\
  \vecnorm{\widetilde{\zeta}_i (f_1) - \widetilde{\zeta}_i (f_2)}{\psi_1} &\leq c' \log^2 (n / \delta) \vecnorm{f_1 - f_2}{\infty}.
\end{align*}
Applying \Cref{lemma:empirical-process-tool} with the above two bounds, and combining the probability of the complement event $\Event^c$, we conclude the inequality
\begin{align*}
  \ErrTermSqVal (r) &\leq c \Big\{\dudley_2 (\Fclass_\valfun (r), \stationary) + r \sqrt{\log (1/ \delta)} \Big\} \sqrt{\frac{\log^4 (n / \delta)}{n}} + c \Big\{\dudley_2 (\Fclass_\valfun (r), \vecnorm{\cdot}{\infty}) + \constFclass \sqrt{\log (1/ \delta)} \Big\} \frac{\sqrt{\log (n / \delta)}}{n^{3/2}} \\
  &\qquad + c \log^3 (\numobs / \delta) \frac{\dudley_1 (\Fclass_\valfun, \vecnorm{\cdot}{\infty}) + \diameter_{\infty} (\Fclass_\valfun) \log (1 / \delta)}{n}\\
  &\leq c \dudley_2 (\Fclass_\valfun (r), \stationary) \frac{\log^3 (n / \delta)}{\sqrt{\numobs}} + c \dudley_1 (\Fclass_\valfun, \vecnorm{\cdot}{\infty}) \frac{\log^4 (\numobs / \delta)}{\numobs},
\end{align*}
with probability $1 - \delta$. This concludes the proof of \Cref{lemma:uniform-concentration-bellman-operator}.

\subsubsection{Proof of \Cref{lemma:empirical-process-tool}}\label{subsubsec:proof-lemma-empirical-process-tool}
To prove this lemma, we use the following known results from literature

\begin{proposition}[Theorem 4 of~\cite{adamczak2008tail}, simplified]
  \label{prop:concentration-adamczak}
 Given a countable class $\funcClass$ of measurable functions, and let
 $X_1, \cdots X_n$ be independent random variables. Assuming that
 $\Exs f (X_i) = 0$ for any $f \in \funcClass$, define $Z \mydefn
 \sup_{f \in \funcClass} \abss{\sum_{i = 1}^n f (X_i)}$ and $v^2
 \mydefn \sup_{f \in \funcClass} \sum_{i = 1}^n \Exs f (X_i)^2$. There
 exists a universal constant $c > 0$, such that for any $t > 0$ and
 $\alpha \geq 1$, we have:
\begin{align*}
\Prob \big( Z > 2 \Exs (Z) + t \big) \leq \exp \Big( \frac{-
  t^2}{4 v^2} \Big) + 3 \exp \Big\{ - \Big( \frac{t}{c
  \vecnorm{\max_i \sup_{f \in \funcClass} |f (X_i)| }{\psi_{1
      /\alpha}}} \Big)^{1 / \alpha} \Big\},
    \end{align*}
\end{proposition}
\begin{proposition}[Theorem 3.5 of~\cite{dirksen2015tail}, simplified]
\label{prop:dirksen}
Given a separable stochastic process $(Y_t)_{t \in T}$ and a pair
$(d_1, d_2)$ of metrics over the index set $T$, satisfying the
following mixed-tail assumption for any pair $s, t \in T$:
\begin{subequations}
\begin{align}
  \Prob \Big( \abss{Y_{s} - Y_{t}} \geq \sqrt{u} d_1 (s, t) + u d_2
  (s, t) \Big) \leq 2 e^{- u}, \quad \mbox{for any } u >
  0,\label{eq:mixed-tail-condition}
\end{align}
suppose that there exists $t_0 \in T$ such that $Y_{t_0} \equiv 0$, for any $\delta \in (0, 1)$, with probability $1 - \delta$, we have
\begin{align}
  \sup_{t \in T} \abss{Y_t} \leq c \Big\{ \dudley_2 (T, d_1) + \dudley_1 (T, d_2) \Big\} + c \Big( \sqrt{\log (1 / \delta)} \sup_{t \in T} d_1 (t, t_0) + \log (1 / \delta) \sup_{t \in T} d_2 (t, t_0) \Big),
\end{align}
\end{subequations}
for some universal constant $c > 0$.
\end{proposition}
We first apply \Cref{prop:concentration-adamczak} with the function class being a singleton to get a pointwise Bernstein inequality. In particular, for any pair $f_1, f_2 \in \funcClass$, we note that
\begin{align*}
  \vecnorm{\max_i |\varepsilon_i (f_1) - \varepsilon_i (f_2)|}{\psi_1} \leq \log \numobs \cdot \vecnorm{\varepsilon_i (f_1) - \varepsilon_i (f_2)}{\psi_1} \leq \log \numobs \cdot d_2 (f_1, f_2).
\end{align*}
 Invoking \Cref{prop:concentration-adamczak} yields that for any $u > 0$,
\begin{align*}
  \Prob \Big\{ \Big|\frac{1}{n} \sum_{i = 1}^{n} \big(  \varepsilon_i (f_1) - \varepsilon_i (f_2)\big)\Big| \geq  c d_1 (f_1, f_2)\sqrt{\frac{u}{\numobs}} + c d_2 (f_1, f_2) \log \numobs \frac{u}{\numobs}  \Big\} \leq e^{- u}
\end{align*}
Then by applying \Cref{prop:dirksen} to the stochastic process $Y_f \mydefn \frac{1}{\numobs} \sum_{i = 1}^{\numobs} \varepsilon_i (f)$ indexed by $f \in \funcClass$, we obtain
\begin{align*}
  \sup_{f \in \funcClass} \abss{ \frac{1}{\numobs} \sum_{i = 1}^\numobs \varepsilon_i (f) } \leq c \frac{\dudley_2 (\funcClass, d_1) + \diameter_{d_1} (\funcClass)\sqrt{\log (1/ \delta)}}{\sqrt{n}} + c \log \numobs  \frac{\dudley_1 (\funcClass, d_2) + \diameter_{d_2} (\funcClass)\log (1 / \delta)}{n},
\end{align*}
with probability at least $1 - \delta$. This concludes the proof.

\section{Discussion}\label{sec:discussion}
In this paper, we have designed and analyzed a new class of model-free off-policy algorithms for continuous-time RL with general function approximation. Building upon new geometric insights into the structure of the underlying Bellman operators, we have established non-asymptotic oracle inequalities for the proposed algorithms without structural assumptions on the underlying MDP such as Bellman completeness. The new algorithms and the structural results open up several interesting directions for future research.
\begin{itemize}
  \item First, while we focus on the off-policy learning problem in this paper, it would be interesting to incorporate exploration into the learning algorithms. In particular, it is important design and analyze model-free algorithms that can provably balance exploration and exploitation in continuous-time RL with general function approximation. We conjecture that the Hilbert space structure induced by ellipticity condition may also be useful in this context.
  \item Second, as the Sobolev-prox updates adapts to the geometry better than standard fitted Q iteration, it is important to extend it to more practical settings. For example, stochastic approximation variants of the proposed algorithms would be more practical in large-scale problems, and the study on non-convex parametric function approximation will shed light on the practical implementations of the proposed algorithms using neural networks.
  \item Finally, beyond the elliptic diffusion setting considered in this paper, it is of great interest to extend the idea of ellipticity-induced geometry to other RL problems, including jump processes, degenerate diffusions, and structured discrete-space MDPs.
\end{itemize}

\section*{Acknowledgements}
This work was partially supported by NSERC grant RGPIN-2024-05092 and a Connaught New Researcher Award to WM. WM thanks Du Ouyang for helpful discussion on approximation error by aggregated dynamics.

\bibliographystyle{alpha}
\bibliography{references}

\appendix

\section{Proof of technical lemmas in \Cref{subsubsec:technical-lemmas}}\label{app:proofs-of-technical-lemmas}
In this appendix, we collect the proofs of several technical lemmas that were used in the proof of \Cref{lemma:projected-fixed-point-approximation}.

\subsection{Proof of \Cref{lemma:generator-positive-definite}}\label{subsubsec:proof-lemma-generator-positive-definite}
By definition, we have
\begin{align*}
   \statinprod{f}{\generator^{\behavpolicy} f} 
   &= \int f (\state) \Big(  \drift^{\behavpolicy} (\state)^\top \nabla f (\state) + \frac{1}{2} \mathrm{Tr} \big( \covMat (\state) \cdot \nabla^2 f (\state) \big) \Big) \stationary (\state) d \state.
\end{align*}
Applying integration by parts to the first term, we have
\begin{align*}
  - \int f (\state)  \drift^{\behavpolicy} (\state)^\top \nabla f (\state) \stationary (\state)  d \state = \frac{1}{2} \int f^2 (x) \nabla \cdot \big( \drift^{\behavpolicy} \stationary \big) (\state) d \state,
\end{align*}
and for the second term, we have
\begin{align*}
  - \int f (\state) \mathrm{Tr} \big( \covMat (\state) \cdot \nabla^2 f (\state) \big) \stationary (\state) d \state = \int \nabla f (\state)^\top \covMat (\state) \nabla f (\state) \stationary (\state) d \state + \int f (\state) \nabla f (\state)^\top \big( \nabla \cdot ( \covMat \stationary ) \big) (\state) d \state.
\end{align*}
Applying integration by parts again to the last term yields
\begin{align*}
  \int f (\state) \nabla f (\state)^\top \big( \nabla \cdot ( \covMat \stationary ) \big) (\state) d \state = - \frac{1}{2} \int f^2 (\state) \nabla^2 \cdot ( \covMat \stationary )(\state) d \state.
\end{align*}
Putting the pieces together, we have
\begin{align*}
  - \statinprod{f}{\generator^{\behavpolicy} f} = \frac{1}{2} \int \nabla f (\state)^\top \covMat (\state) \nabla f (\state) \stationary (\state) d \state - \frac{1}{2} \int f^2 (\state) \cdot (\generator^{\behavpolicy})^* \stationary (\state) d \state.
\end{align*}
By the uniform ellipticity condition in \Cref{assume:ellipticity}, we have
\begin{align*}
  \int \nabla f (\state)^\top \covMat (\state) \nabla f (\state) \stationary (\state) d \state \geq \lammin \statnorm{\nabla f}^2.
\end{align*}
It remains to bound the term involving $(\generator^{\behavpolicy})^* \stationary$. In doing so, we introduce an auxiliary occupancy measure, which is the small-stepsize limit of $\stationary$:
\begin{align*}
  \widetilde{\stationary} \mydefn \discount \int_0^\infty e^{- \discount t}  (\semigroup_t^{\behavpolicy})^* \initDistr dt.
\end{align*}
Applying integration-by-parts formula in the time domain, we have
\begin{multline*}
  \widetilde{\stationary} = - e^{- \discount t} (\semigroup_t^{\behavpolicy})^* \initDistr  \Big|_{t = 0}^{t = \infty} + \int_0^\infty e^{- \discount t}  (\partial_t \semigroup_t^{\behavpolicy})^* \initDistr dt\\
  = \initDistr + \int_0^\infty e^{- \discount t}  (\generator^{\behavpolicy})^* (\semigroup_t^{\behavpolicy})^* \initDistr dt = \initDistr + \frac{1}{\discount} (\generator^{\behavpolicy})^* \widetilde{\stationary}.
\end{multline*}
So we have $(\generator^{\behavpolicy})^* \widetilde{\stationary} = \discount (\widetilde{\stationary} - \initDistr)$, and consequently,
\begin{align*}
  - \frac{1}{2} \int f^2 (\state) \cdot (\generator^{\behavpolicy})^* \widetilde{\stationary} (\state) d \state = \frac{\discount}{2} \int f^2 (\state) \big( \initDistr (\state) - \widetilde{\stationary} (\state) \big) d \state = \frac{\discount}{2} \vecnorm{f}{\initDistr}^2 - \frac{\discount}{2} \vecnorm{f}{\widetilde{\stationary}}^2.
\end{align*}
Combining the pieces yields
\begin{align*}
  &\statinprod{f}{(\discount - \generator^{\behavpolicy}) f}  \\
  &\geq \lammin \statnorm{\nabla f}^2 + \frac{\discount}{2} \statnorm{f}^2 + \frac{\discount}{2} \vecnorm{f}{\initDistr}^2 - \frac{\discount}{2} \abss{ \vecnorm{f}{\widetilde{\stationary}}^2 - \vecnorm{f}{\stationary}^2 } - \frac{1}{2} \abss{ \int f^2 (\state) \cdot  (\generator^{\behavpolicy})^* \big(\stationary - \widetilde{\stationary} \big) (\state)  d \state }\\
  &\geq \lammin \statnorm{\nabla f}^2 + \frac{\discount}{2} \statnorm{f}^2 + \frac{\discount}{2} \vecnorm{f}{\initDistr}^2 - \frac{\discount}{2} \abss{ \vecnorm{f}{\widetilde{\stationary}}^2 - \vecnorm{f}{\stationary}^2 } - \frac{1}{2} \abss{ \int \generator^{\behavpolicy} (f^2) (\state) \cdot  \big(\stationary - \widetilde{\stationary} \big) (\state)  d \state }.
\end{align*}
It remains to bound the last two terms, which involves the error induced by approximating $\widetilde{\stationary}$ with $\stationary$. We use the following lemma to control such approximation error.
\begin{lemma}\label{lemma:occupancy-measure-approximation}
  Under the setup of \Cref{lemma:projected-fixed-point-approximation}, given $\discount \geq 2 \couyang$, for any test function $g \in \Clin{4}$, we have
  \begin{align*}
    \abss{ \int g (\state) \big( \stationary - \widetilde{\stationary} \big) (\state) d \state } \leq c \stepsize \vecnorm{g}{\Clin{4}},
  \end{align*}
  where the constant $c > 0$ depends on the regularity constants in \Cref{assume:coefficient-regularity} and $\couyang$ defined in \Cref{prop:ouyang}.
\end{lemma}
\noindent See~\Cref{subsec:app-proof-lemma-occupancy-measure-approximation} for the proof of this lemma.

Applying \Cref{lemma:occupancy-measure-approximation},we have the error bounds
\begin{align*}
  &\abss{ \vecnorm{f}{\widetilde{\stationary}}^2 - \vecnorm{f}{\stationary}^2 }  \leq c \stepsize \vecnorm{f^2}{\Clin{4}} \leq c \stepsize \constFclass^2, \qquad \mbox{and}\\
  &\abss{ \int \generator^{\behavpolicy} (f^2) (\state) \cdot  \big(\stationary - \widetilde{\stationary} \big) (\state)  d \state }  \leq c  \stepsize \vecnorm{\generator^{\behavpolicy} (f^2)}{\Clin{4}} \leq c  \stepsize \constFclass^2,
\end{align*}
where the constant $c > 0$ is from \Cref{lemma:occupancy-measure-approximation} and the last step follows from \Cref{assume:function-class,assume:coefficient-regularity}.
Putting the pieces together yields the desired result.

\subsubsection{Proof of \Cref{lemma:occupancy-measure-approximation}}\label{subsec:app-proof-lemma-occupancy-measure-approximation}
By definition, we have
\begin{align*}
  \widetilde{\stationary} - \stationary = (1 - e^{- \discount \stepsize}) \sum_{k = 0}^{+ \infty} \int_{k \stepsize}^{(k + 1) \stepsize} e^{- \discount t} \Big\{ (\semigroup_t^{\behavpolicy})^* \initDistr - [(\obstransition)^*]^k \initDistr \Big\} dt.
\end{align*}

Recall that $(\State_t)_{t \geq 0}$ and $(\widetilde{\State}_t)_{t \geq 0}$ are the actual observation process and the idealized aggregated dynamics, respectively. Invoking \Cref{prop:ouyang}, we have
\begin{align*}
  \abss{\Exs \big[ g (\State_t) \big] - \Exs \big[ g (\widetilde{\State}_t) \big]} \leq \exp \big( \couyang (1 + t) \big) \stepsize \vecnorm{g}{\Clin{4}}.
\end{align*}
On the other hand, for $t \in [k \stepsize, (k + 1) \stepsize]$, by It\^{o}'s formula, we have
\begin{align*}
   \abss{\Exs [g (X_t)] - \Exs [g (X_{k \stepsize})]} = \abss{\int_{k \stepsize}^t \Exs [\generator^{\behavpolicy} g (X_s)] ds} \leq c_1 \stepsize \vecnorm{g}{\Clin{2}},
\end{align*}
where the constants $c_1$ depends on the bounds in \Cref{assume:coefficient-regularity}.

Putting them together, we have
\begin{align*}
  \abss{ \Exs [g (X_{k \stepsize})] - \Exs [g (\widetilde{X}_t)] } \leq  \stepsize \big\{ c_1 + e^{(1 + t) \couyang}\big\} \vecnorm{g}{\Clin{4}}.
\end{align*}
Integrating over $t$ with exponential discounting, we have
\begin{align*}
  \abss{ \int g (\state) \big( \stationary - \widetilde{\stationary} \big) (\state) d \state } &\leq (1 - e^{- \discount \stepsize}) \sum_{k = 0}^{+ \infty} \int_{k \stepsize}^{(k + 1) \stepsize} e^{- \discount t} \abss{ \Exs [g (X_{k \stepsize})] - \Exs [g (\widetilde{X}_t)] } dt\\
  &\leq \discount  \vecnorm{g}{\Clin{4}} \cdot  \stepsize \int_0^{+ \infty} e^{- \discount t}  \big\{ c_1 + e^{(1 + t) \couyang}\big\} dt .
\end{align*}
When we have $\discount \geq 2 \couyang$, we can further bound the integral as
\begin{align*}
  \discount \int_0^{+ \infty} e^{- \discount t}  \big\{ c_1 + e^{(1 + t) \couyang}\big\} dt \leq c_1 + 2 e^{\couyang}.
\end{align*}
Putting the pieces together yields the desired result. 

\subsection{Proof of \Cref{lemma:generator-upper-bound}}\label{subsubsec:proof-lemma-generator-upper-bound}
Similar to the proof of \Cref{lemma:generator-positive-definite}, by applying integration by parts, we have
\begin{multline*}
  - \statinprod{f}{\generator^{\action} g} = 
  - \int f (\state)  \drift^{\action} (\state)^\top \nabla g (\state) \stationary (\state)  d  \state + \frac{1}{2} \int \nabla f (\state)^\top \covMat (\state) \nabla g (\state)\stationary (\state) d \state \\
  + \frac{1}{2} \int f (\state) \nabla g (\state)^\top \big( \nabla \cdot \covMat \big) (\state)\stationary  (\state) d \state + \frac{1}{2} \int f (\state) \nabla g (\state)^\top \covMat (\state) \nabla \log \stationary (\state) \cdot\stationary  (\state) d \state 
\end{multline*}
By Cauchy--Schwarz inequality, we can bound the second and third terms as
\begin{align*}
  \int \nabla f (\state)^\top \covMat (\state) \nabla g (\state)\stationary (\state) d \state &\leq \lammax \sobonorm{f} \cdot \sobonorm{g}, \qquad \mbox{and}\\
  \int f (\state) \nabla g (\state)^\top \big(\nabla \cdot \covMat \big) (\state)\stationary  (\state) d \state &\leq \creg \statnorm{f} \cdot \sobonorm{g}.
\end{align*}

In order to bound the terms involving the drift, we note that \Cref{assume:coefficient-regularity} and \Cref{lemma:process-moment-bound} together imply
\begin{align*}
  \Big\{ \Exs \big[ \abss{\drift^{\action} (\State_t)}^{p} \big] \Big\}^{1/p} \leq \creg c_0 (1 + t) \sqrt{p}.
\end{align*}
And consequently, we can bound the moment under the the occupancy measure $\stationary$ as
\begin{align*}
   \int \abss{\drift^{\action} (\state)}^p \stationary (\state) d \state &\leq (1 - e^{- \discount \stepsize}) \sum_{k = 0}^\infty e^{- k \discount \stepsize} \Exs \big[ \abss{\drift^{\action} (\State_{k \stepsize})}^p \big]\\
   &\leq \discount \int_0^\infty e^{- \discount t} \Big\{ \creg c_0 (1 + t) \sqrt{p} \Big\}^p dt \leq (c'p)^{p/2},
\end{align*}
for a constant $c'$ depending on the problem parameters. This implies that $\drift^{\behavpolicy} (\State)$ is sub-Gaussian for $\State \sim \stationary$, with Orlicz norm bounded by a constant depending on the problem parameters. Invoking H\"{o}lder's inequality, we have
\begin{align*}
  \abss{ \int f (\state)  \drift^{\action} (\state)^\top \nabla g (\state) \stationary (\state)  d  \state } \leq \vecnorm{f}{\lpspace{4} (\stationary)} \cdot \vecnorm{\drift^{\action}}{\lpspace{4} (\stationary)} \cdot \vecnorm{\nabla g}{\stationary} \leq c \vecnorm{f}{\lpspace{4} (\stationary)} \cdot \vecnorm{\nabla g}{\stationary},
\end{align*}
and similarly, by swapping the roles of $f$ and $\nabla g$, we have
\begin{align*}
  \abss{ \int f (\state)  \drift^{\action} (\state)^\top \nabla g (\state) \stationary (\state)  d  \state } \leq c \vecnorm{f}{\stationary} \cdot \vecnorm{\nabla g}{\lpspace{4} (\stationary)},
\end{align*}
which proves the first two bounds.

In order to prove the last bound, we apply integration-by-parts formula once more to write
\begin{multline*}
   \statinprod{f}{\generator^{\action} g} = 
  - \int g (\state) \nabla^\top \big( f (\state)  \drift^{\action} (\state)^\top   \stationary (\state) \big) d  \state + \frac{1}{2} \int g (\state) \nabla^\top \big( \covMat (\state) \nabla f (\state) \stationary (\state) \big) d \state \\
  + \frac{1}{2} \int   g (\state) \nabla^\top \big(  f \stationary  \nabla \cdot \covMat \big) (\state)  d \state + \frac{1}{2} \int g (\state) \nabla^\top \big( f (\state)  \covMat (\state) \nabla \stationary (\state) \big) d \state.
\end{multline*}
By applying Cauchy--Schwarz inequality to each term in the decomposition, and invoking \Cref{assume:coefficient-regularity,assume:function-class}, as well as \Cref{lemma:process-moment-bound,lemma:occupancy-measure-gradient-log-density}, we can conclude that
\begin{align*}
  \abss{ \statinprod{f}{\generator^{\action} g} } \leq c \statnorm{g} \cdot \vecnorm{f}{C^2},
\end{align*}
for a constant $c > 0$ depending on the problem parameters. This completes the proof.

\subsection{Proof of \Cref{lemma:process-moment-bound}}\label{subsubsec:proof-lemma-process-moment-bound}
For notation simplicity, we drop the dependence on the action $\Action_{k \stepsize}$ in the drift function. By It\^{o}'s formula, for any $t$, we have
\begin{align*}
  \eucnorm{X_t}^2 &= 2 \int_0^t X_s^\top \drift (X_s) ds + \int_0^t \mathrm{Tr}(\covMat(X_s)) ds + 2 \int_0^t X_s^\top \covMat^{1/2} (X_s) d \BM_s\\
  &\leq (2 \constbd + \lammax d) t + 2 \int_0^t X_s^\top \covMat^{1/2} (X_s) d \BM_s,
\end{align*}
Taking supremum over $t \in [0, T]$, we have
\begin{align*}
  \sup_{t \in [0, T]} \eucnorm{X_t}^2 &\leq (2 \constbd + \lammax d)  T + 2 \sup_{t \in [0, T]} \abss{ \int_0^t X_s^\top \covMat^{1/2} (X_s) d \BM_s }.
\end{align*}
By Burkholder--Davis--Gundy inequality, there exists a universal constant $c > 0$ such that for any $p \geq 2$, we have
\begin{align*}
  \Exs \Big[ \sup_{t \in [0, T]} \abss{ \int_0^t X_s^\top \covMat^{1/2} (X_s) d \BM_s }^p \Big] &\leq c \Exs \Big[ \Big( \int_0^T X_s^\top \covMat (X_s) X_s ds \Big)^{p/2} \Big] \leq c \lammax^{p/2} T^{p/2 - 1} \int_0^T \Exs \big[ \eucnorm{X_s}^p \big] ds.
\end{align*}
So it suffices to control the $p$-th moment of $\eucnorm{X_t}$. By It\^{o}'s formula again, we have
\begin{align*}
 d \Exs \big[\eucnorm{X_t}^p \big] &= p \Exs \big[\eucnorm{X_t}^{p-2} X_t^\top \drift^{\action_k}(X_t) \big] dt + \frac{p(p-2)}{2} \Exs \big[\eucnorm{X_t}^{p-4} X_t^\top \covMat(X_t) X_t \big] dt + \frac{p}{2} \Exs \big[\eucnorm{X_t}^{p-2} \mathrm{Tr}(\covMat(X_t)) \big] dt\\
 &\leq p \constbd \Exs \big[\eucnorm{X_t}^{p-2} \big] dt + \frac{p(p-1)}{2} \lammax \Exs \big[\eucnorm{X_t}^{p-2} \big] dt\\
  &\leq \Big( p \constbd + \frac{p(p-1)}{2} \lammax \Big) \Big\{\Exs \big[\eucnorm{X_t}^{p} \big] \Big\}^{\frac{p-2}{p}} dt.
\end{align*}
Some algebra yields
\begin{align*}
  \frac{d}{dt} \Big\{ \Exs \big[\eucnorm{X_t}^{p} \big]^{2/p} \Big\} \leq 2 \constbd + (p-1) \lammax, \quad \mbox{and} \quad \frac{d}{dt} \Big\{ \Exs \big[\eucnorm{\widetilde{X}_t}^{p} \big]^{2/p} \Big\} \leq 2 \constbd + (p-1) \lammax,
\end{align*}
for any $t \geq 0$. Integrating over $t$ twice yields
\begin{align*}
  \int_0^T \Exs \big[\eucnorm{X_t}^{p} \big] dt \leq T \Exs \big[ \eucnorm{X_0}^p \big] + (2 \constbd + p \lammax )^{p/2} T^{p/2 + 1}.
\end{align*}
Substituting back into the previous bound, we have
\begin{align*}  \Exs \Big[ \sup_{t \in [0, T]} \abss{ \int_0^t X_s^\top \covMat^{1/2} (X_s) d \BM_s }^p \Big] &\leq   (1 + T)^{p} \Big\{c \lammax(2 \constbd + p \lammax ) \Big\}^{p/2} 
\end{align*}
Consequently, we have
\begin{align*}
  \Big\{ \Exs \big[ \sup_{t \in [0, T]} \eucnorm{X_t}^p \big] \Big\}^{1/p} &\leq \sqrt{(2 \constbd + \lammax d) T} + c' (1 + T) \sqrt{\lammax (2 \constbd + p \lammax) }\\
  &\leq c_0 (1 + T) \sqrt{p},
\end{align*}
Note that the above differential inequality is independent of the choice of action $\action_k$, and therefore holds for all $t \geq 0$, and consequently, for the process $\{\widetilde{X}_t\}_{t \geq 0}$ as well. This completes the proof.

\subsection{Proof of \Cref{lemma:occupancy-measure-gradient-log-density}}\label{app:subsec-proof-lemma-occupancy-measure-gradient-log-density}
Note that for any $p \geq 1$, the map
$(u,v) \longmapsto \frac{|v|^p}{u^{p-1}}$
is the perspective of the convex function $z \mapsto |z|^p$, and therefore jointly convex (see \cite{Rockafellar}). Consequently, the functional
\begin{align*}
\mu \mapsto \int_{\mathbb{R}^d} |\nabla \log \mu(x)|^p \mu (x) dx
\end{align*}
is convex in the density $\mu$.

By Jensen's inequality, we have
\begin{align*}
  \int \rho (x) |\nabla \log \rho (x)|^p dx \leq (1 - e^{- \discount \stepsize}) \sum_{k = 0}^\infty e^{- k \discount \stepsize} \int \mu_{k \stepsize} (x) |\nabla \log \mu_{k \stepsize} (x)|^p dx,
\end{align*}
where $\mu_t$ is the density of the process $\{X_t\}_{t \geq 0}$ at time $t$. By \Cref{assume:density-regularity}, we have
\begin{align*}
  \int \mu_{k \stepsize} (x) |\nabla \log \mu_{k \stepsize} (x)|^p dx \leq  p^{p/2} \exp(\cmalliavin p (1 + k \stepsize)).
\end{align*}
Substituting back into the previous bound, when $\discount \geq 2 \cmalliavin p$, we have
\begin{align*}
  \int \rho (x) |\nabla \log \rho (x)|^p dx \leq \discount e^{p \cmalliavin} p^{p/2} \int_0^\infty e^{- \discount t / 2}  dt = 2 e^{p \cmalliavin} p^{p/2}.
\end{align*}
which proves the bound for the first derivative.

For the second derivative, we apply similar argument to obtain the convexity of the functional
\begin{align*}
\mu \mapsto \int_{\mathbb{R}^d} \frac{\eucnorm{\nabla^2 \mu (x)}^2}{\mu (x)} dx.
\end{align*}
By Jensen's inequality, we have
\begin{align*}
  \int \rho (x) \frac{\eucnorm{\nabla^2 \rho (x)}^2}{\rho (x)} dx \leq (1 - e^{- \discount \stepsize}) \sum_{k = 0}^\infty e^{- k \discount \stepsize} \int \mu_{k \stepsize} (x) \frac{\eucnorm{\nabla^2 \mu_{k \stepsize} (x)}^2}{\mu_{k \stepsize} (x)} dx.
\end{align*}
Invoking \Cref{assume:density-regularity} again, we have
\begin{align*}
  \int_{\mathbb{R}^d} \frac{\eucnorm{\nabla^2 \mu_{k \stepsize} (x)}^2}{\mu_{k \stepsize} (x)} dx \leq \exp ( \cmalliavin (1 + k \stepsize)), \qquad \mbox{for any } k \geq 0.
\end{align*}
Substituting back, for $\discount \geq 2 \cmalliavin$, we have
\begin{align*}
  \int \rho (x) \frac{\eucnorm{\nabla^2 \rho (x)}^2}{\rho (x)} dx \leq 2 e^{\cmalliavin} .
\end{align*}
This completes the proof.

\section{Proof of technical lemmas in \Cref{subsec:proof-thm-main}}\label{app:subsec-proof-technical-lemmas-data-driven}
We collect the proofs of technical lemmas used in the statistical error analysis in \Cref{thm:main}.

\subsection{Proof of \Cref{lemma:bellman-variance-control-in-emp-proc-valfunc-proof}}\label{app:subsec-proof-lemma-bellman-variance-control-in-emp-proc-valfunc-proof}
By It\^{o}'s formula, we have
\begin{align*}
  &\valfun (\State_{k \stepsize}^{(i)}) - e^{- \discount \stepsize} \valfun (\State_{(k + 1) \stepsize}^{(i)}) \\
  &= \int_{k \stepsize}^{(k + 1) \stepsize} e^{- \discount (t - k \stepsize)} \big( \discount - \generator^{\Action_{k \stepsize}^{(i)}} \big) \valfun (\State_t^{(i)}) dt + \int_{k \stepsize}^{(k + 1) \stepsize} e^{- \discount (t - k \stepsize)} \nabla \valfun (\State_t^{(i)})^\top \covMat^{1/2} (\State_t^{(i)}) d \BM_t.
\end{align*}
Define the terms
\begin{align*}
  I_1 &\mydefn \sum_{k = 0}^{\lfloor T_i / \stepsize \rfloor} f(\State_{k \stepsize}^{(i)}) \cdot \int_{k \stepsize}^{(k + 1) \stepsize} e^{- \discount (t - k \stepsize)} \big( \discount - \generator^{\Action_{k \stepsize}^{(i)}} \big) \valfun (\State_t^{(i)}) dt ,\\
  I_2 &\mydefn \sum_{k = 0}^{\lfloor T_i / \stepsize \rfloor} f(\State_{k \stepsize}^{(i)}) \cdot \int_{k \stepsize}^{(k + 1) \stepsize} e^{- \discount (t - k \stepsize)} \nabla \valfun (\State_t^{(i)})^\top \covMat^{1/2} (\State_t^{(i)}) d \BM_t.
\end{align*}

\paragraph{Second moment bounds:}
For the term $I_1$, by Cauchy--Schwarz inequality, we have
\begin{align*}
 \Exs \big[I_1^2 \bm{1}_{T_i \leq \Tmax} \big] &\leq \frac{\Tmax}{\stepsize} \Exs \Big[ \sum_{k = 0}^{\lfloor T_i / \stepsize \rfloor} \Big\{ f(\State_{k \stepsize}^{(i)}) \cdot \int_{k \stepsize}^{(k + 1) \stepsize} e^{- \discount (t - k \stepsize)} \big( \discount - \generator^{\Action_{k \stepsize}^{(i)}} \big) \valfun (\State_t^{(i)}) dt \Big\}^2 \Big]\\
 &\leq c \stepsize \Exs \Big[ \sum_{k = 0}^{\lfloor T_i / \stepsize \rfloor} f (\State_{k \stepsize}^{(i)})^2 \cdot \Big\{1 + \sup_{0 \leq t \leq \Tmax} \eucnorm{X_t^{(i)}}^2 \Big\} \Big],
\end{align*}
where we have used the regularity assumption on the value function $\valfun$ and the coefficients of the diffusion process. The dependence on the norm of the process $\{X_t^{(i)}\}_{t \geq 0}$ comes from the linear growth condition on the drift term (\Cref{assume:coefficient-regularity}).

Define the event
\begin{align*}
  \Event (R) \mydefn \Big\{ \sup_{t \in [0, \Tmax]} \eucnorm{X_t^{(i)}} \leq R \Big\},
\end{align*}
where $R$ is a parameter to be chosen later. We have the decomposition
\begin{align*}
  \Exs \big[I_1^2 \bm{1}_{T_i \leq \Tmax} \big] &\leq c \stepsize (1 + R^2) \Exs \Big[ \sum_{k = 0}^{\lfloor T_i / \stepsize \rfloor} f (\State_{k \stepsize}^{(i)})^2 \Big] + c \stepsize \Exs \Big[ \sum_{k = 0}^{\lfloor T_i / \stepsize \rfloor} f (\State_{k \stepsize}^{(i)})^2  \Big\{1 + \sup_{0 \leq t \leq \Tmax} \eucnorm{X_t^{(i)}}^2 \Big\} \bm{1}_{\Event(R)^c} \Big]\\
  &\leq \frac{c}{\discount} (1 + R^2) \statnorm{f}^2 + \frac{c}{\discount} \vecnorm{f}{\infty}^2 \Exs \Big[\Big\{ 1 + \sup_{0 \leq t \leq \Tmax} \eucnorm{X_t^{(i)}}^2  \Big\} \bm{1}_{\Event(R)^c} \Big].
\end{align*}
By \Cref{lemma:process-moment-bound}, the supremum of squared norm of the process $\{X_t^{(i)}\}_{t \geq 0}$ has sub-exponential tail. By standard tail bound for sub-exponential random variables, choosing
\begin{align*}
  R = c' (1 + \Tmax) \sqrt{\log n},
\end{align*}
we have the bound
\begin{align*}
  \Exs \Big[\Big\{ 1 + \sup_{0 \leq t \leq \Tmax} \eucnorm{X_t^{(i)}}^2  \Big\} \bm{1}_{\Event(R)^c} \Big] \leq \frac{1}{n}.
\end{align*}
Substituting back, we have
\begin{align*}
  \Exs \big[I_1^2 \bm{1}_{T_i \leq \Tmax} \big] &\leq c_1 \big( 1 + \Tmax^2 \log n \big) \statnorm{f}^2 + \frac{c_1}{\discount n} \vecnorm{f}{\infty}^2.
\end{align*}

Note that the term $I_2$ is a discrete-time martingale with respect to the natural filtration generated by the Brownian motion and the randomly sampled action sequence, stopped at the random time $\lfloor T_i / \stepsize \rfloor$. Since the stopping time is independent of the natural filtration, we have
\begin{align*}
  \Exs \big[ I_2^2 \big] &= \sum_{k = 0}^{\infty} \Exs \Big[ \abss{f(\State_{k \stepsize}^{(i)}) \cdot \int_{k \stepsize}^{(k + 1) \stepsize} e^{- \discount (t - k \stepsize)} \nabla \valfun (\State_t^{(i)})^\top \covMat^{1/2} (\State_t^{(i)}) d \BM_t}^2 \Big] \cdot \Prob \big( \lfloor T_i / \stepsize \rfloor \geq k \big)\\
  &\leq \constFclass^2 \lammax \stepsize \sum_{k = 0}^{\infty} \Exs \big[ \eucnorm{f (\State_{k \stepsize}^{(i)})}^2 \big] e^{- k \discount \stepsize } \leq \frac{\constFclass^2 \lammax }{\discount} \statnorm{f}^2.
\end{align*}
Combining the bounds for the terms $I_1$ and $I_2$ yields the second moment bound.

\paragraph{Orlicz norm bounds:} We now turn to the Orlicz norm bounds. For the term $I_1$, we have
\begin{align*}
  \abss{I_1 \bm{1}_{T_i \leq \Tmax}} &\leq \sum_{k = 0}^{\Tmax} \abss{ f(\State_{k \stepsize}^{(i)}) \cdot \int_{k \stepsize}^{(k + 1) \stepsize} e^{- \discount (t - k \stepsize)} \big( \discount - \generator^{\Action_{k \stepsize}^{(i)}} \big) \valfun (\State_t^{(i)}) dt }\\
  &\leq c \Tmax \vecnorm{f}{\infty} \cdot \Big\{ 1 + \sup_{0 \leq t \leq \Tmax} \eucnorm{X_t^{(i)}} \Big\}.
\end{align*}
By \Cref{lemma:process-moment-bound}, the supremum of the norm of the process $\{X_t^{(i)}\}_{t \geq 0}$ has sub-Gaussian tail, and therefore, we have 
\begin{align*}
  \orlicznorm{\sup_{0 \leq t \leq \Tmax} \eucnorm{X_t^{(i)}}}{2} \leq c' (1 + \Tmax).
\end{align*}
Substituting back, we have
\begin{align*}
  \orlicznorm{I_1 \bm{1}_{T_i \leq \Tmax}}{1} \leq \orlicznorm{I_1 \bm{1}_{T_i \leq \Tmax}}{2}  \leq c_2 \log^2 (n / \delta) \vecnorm{f}{\infty}.
\end{align*}
As for the term $I_2$, by Burkholder--Davis--Gundy inequality, for any $p \geq 2$, we have
\begin{align*}
  \Exs \big[ |I_2|^p \bm{1}_{T_i \leq \Tmax} \big] &\leq c^p p^{p/2} \Exs \Big[ \Big( \sum_{k = 0}^{\lfloor \Tmax / \stepsize \rfloor} f(\State_{k \stepsize}^{(i)})^2 \cdot \int_{k \stepsize}^{(k + 1) \stepsize} e^{- 2\discount (t - k \stepsize)} \eucnorm{\nabla \valfun (\State_t^{(i)})^\top \covMat^{1/2} (\State_t^{(i)})}^2 dt \Big)^{p/2} \Big]\\
  &\leq \Big\{c^2 p \Tmax \vecnorm{f}{\infty}^2 c'\Big\}^{p/2},
\end{align*}
which implies the Orlicz norm bound
\begin{align*}
  \orlicznorm{I_2 \bm{1}_{T_i \leq \Tmax}}{1} \leq \orlicznorm{I_2 \bm{1}_{T_i \leq \Tmax}}{2}  \leq c_3 \sqrt{\log (n / \delta)}\vecnorm{f}{\infty}.
\end{align*}
Combining the bounds for the terms $I_1$ and $I_2$ yields the Orlicz norm bound, which completes the proof.

\subsection{Proof of \Cref{lemma:uniform-concentration-norm-advantage}}\label{app:subsec-proof-lemma-uniform-concentration-norm-advantage}
The proof is similar to that of \Cref{lemma:uniform-concentration-sobolev-norm}. We define the random functionals
\begin{align*}
  \zeta_i (f) \mydefn (1 - e^{- \discount \stepsize}) \sum_{k = 0}^{\lfloor T_i / \stepsize \rfloor - 1} f (\State_{k \stepsize}^{(i)}, \Action_{k \stepsize}^{(i)})^2, \quad \mbox{for } i = 1, 2, \ldots,\numobs,
\end{align*}
as well as the truncated version $\widetilde{\zeta}_i (f) \mydefn \zeta_i (f) \bm{1}_{T_i \leq \Tmax}$. On the event $\mathcal{E} = \{ T_i \leq \Tmax, \mbox{ for all } i = 1, 2, \ldots, \numobs \}$, we have $\zeta_i (f) = \widetilde{\zeta}_i (f)$ for all $i$. Following the proof of \Cref{lemma:uniform-concentration-sobolev-norm}, we can choose $\Tmax = $$c_0 \discount^{-1} \log (\numobs / \delta)$ for a sufficiently large constant $c_0 > 0$ such that $\Prob (\mathcal{E}^c) \leq \delta / 2$. Furthermore, the truncation error in the expectation can be controlled as
\begin{align*}
  \abss{ \Exs [\zeta_i (f)] - \Exs [\widetilde{\zeta}_i (f)] } \leq \constFclass \Exs \big[ \discount T_i \bm{1}_{T_i \geq \Tmax} \big] \leq \frac{\delta}{\numobs^2}.
\end{align*}
So it suffices to control the uniform concentration of the truncated functionals $\widetilde{\zeta}_i (f)$.

Given a pair of functions $f_1, f_2 \in \Fclass_\advFunc$, we apply Cauchy--Schwarz inequality to obtain
\begin{align*}
 \Exs \big[ \big| \widetilde{\zeta}_i (f_1) - \widetilde{\zeta}_i (f_2) \big|^2 \big] &\leq \stepsize \discount^2 \Tmax \Exs \Big[ \sum_{k = 0}^{\lceil T_i / \stepsize \rceil - 1} \big| f_1 (\State_{k \stepsize}^{(i)}, \Action_{k \stepsize}^{(i)})^2 - f_2 (\State_{k \stepsize}^{(i)}, \Action_{k \stepsize}^{(i)})^2 \big|^2  \Big]\\
 &\leq 4 \stepsize \discount^2 \Tmax \constFclass^2 \Exs \Big[ \sum_{k = 0}^{\lceil T_i / \stepsize \rceil - 1} \big| \big(f_1 - f_2 \big) (\State_{k \stepsize}^{(i)}, \Action_{k \stepsize}^{(i)}) \big|^2  \Big]\\
  &\leq 4 c_0 \constFclass^2 \log (n / \delta) \Statnorm{f_1 - f_2}^2.
\end{align*}
We also have the almost sure bound
\begin{align*}
  \abss{ \widetilde{\zeta}_i (f_1) - \widetilde{\zeta}_i (f_2) } \leq 2 \constFclass \discount \Tmax \vecnorm{f_1 - f_2}\infty \leq c_0' \constFclass \log (n / \delta) \vecnorm{f_1 - f_2}\infty.
\end{align*}
Invoking \Cref{lemma:empirical-process-tool}, and combining with the probability of the truncation event, we have with probability at least $1 - \delta$,
\begin{align*}
  \ErrTermSqAdv &\leq  c \Big\{\dudley_2 (\Fclass_\advFunc (r), \Stationary) + r \sqrt{\log (1/ \delta)} \Big\} \sqrt{\frac{\log (n / \delta)}{n}} + c \log^2 (\numobs / \delta) \frac{\dudley_1 (\Fclass_\advFunc, \mathbb{L}^\infty) + \diameter_{\infty} (\Fclass_\advFunc) \log (1 / \delta)}{n}\\
  &\leq  c \dudley_2 (\Fclass_\advFunc (r), \Stationary) \frac{\log (n / \delta)}{\sqrt{\numobs}} + c \dudley_1 (\Fclass_\advFunc, \mathbb{L}^\infty)  \frac{\log^3 (\numobs / \delta)}{n},
\end{align*}
which completes the proof.

\subsection{Proof of \Cref{lemma:uniform-concentration-bellman-advantage}}\label{app:subsec-proof-lemma-uniform-concentration-bellman-advantage}
The proof is similar to that of \Cref{lemma:uniform-concentration-bellman-operator}. We define the random functionals
\begin{align*}
  \zeta_i (f) \mydefn (1 - e^{- \discount \stepsize}) \sum_{k = 0}^{\lfloor T_i / \stepsize \rfloor - 1} f (\State_{k \stepsize}^{(i)}, \Action_{k \stepsize}^{(i)}) \cdot \xi_{i, k}, \quad \mbox{for } i = 1, 2, \ldots,\numobs,
\end{align*}
as well as the truncated version $\widetilde{\zeta}_i (f) \mydefn \zeta_i (f) \bm{1}_{T_i \leq \Tmax}$. On the event $\mathcal{E} = \{ T_i \leq \Tmax, \mbox{ for all } i = 1, 2, \ldots, \numobs \}$, we have $\zeta_i (f) = \widetilde{\zeta}_i (f)$ for all $i$. Following the proof of \Cref{lemma:uniform-concentration-sobolev-norm}, we can choose $\Tmax = c_0 \discount^{-1} \log (\numobs / \delta)$ for a sufficiently large constant $c_0 > 0$ such that $\Prob (\mathcal{E}^c) \leq \delta / 2$. Furthermore, the truncation error in the expectation can be controlled as
\begin{align*}
  \abss{ \Exs [\zeta_i (f)] - \Exs [\widetilde{\zeta}_i (f)] } \leq \discount \stepsize \constFclass \Exs \Big[ \sum_{k = 0}^{\lfloor T_i / \stepsize \rfloor} |\xi_{i, k}| \bm{1}_{T_i > \Tmax} \Big] \leq c' \discount \Exs \Big[ \frac{T_i}{\stepsize} \bm{1}_{T_i > \Tmax} \Big] \leq \frac{\delta}{\numobs^2},
\end{align*}
when we choose the constant $c_0$ sufficiently large and $\delta \leq \stepsize$. So it suffices to control the uniform concentration of the truncated functionals $\widetilde{\zeta}_i (f)$.

In order to apply \Cref{lemma:empirical-process-tool}, let us now verify the Bernstein-type condition. For notational simplicity, we split the random variable $\widetilde{\zeta}_i (f)$ into two parts as $\widetilde{\zeta}_i (f) = \widetilde{\zeta}_i^{(1)} (f) + \widetilde{\zeta}_i^{(2)} (f)$, where
\begin{align*}
  \widetilde{\zeta}_i^{(1)} (f) &\mydefn (1 - e^{- \discount \stepsize}) \sum_{k = 0}^{\lfloor T_i / \stepsize \rfloor - 1} f (\State_{k \stepsize}^{(i)}, \Action_{k \stepsize}^{(i)}) \cdot \Big\{
  \randreward^{(i)}_{k \stepsize} + e^{ - \discount \stepsize} \max_{\action' \in \actionspace} \advFunc(\State_{(k + 1) \stepsize}^{(i)}, \action') - g^* (\State_{k \stepsize}^{(i)}, \Action_{k \stepsize}^{(i)}) \Big\} \bm{1}_{T_i \leq \Tmax},\\
  \widetilde{\zeta}_i^{(2)} (f) &\mydefn (1 - e^{- \discount \stepsize}) e^{- \discount \stepsize} \sum_{k = 0}^{\lfloor T_i / \stepsize \rfloor - 1} \frac{e^{- \discount \stepsize}}{\stepsize} f (\State_{k \stepsize}^{(i)}, \Action_{k \stepsize}^{(i)})  \big(\valfun (\State_{(k + 1) \stepsize}^{(i)}) - \valfun (\State_{k \stepsize}^{(i)}) \big) \bm{1}_{T_i \leq \Tmax}.
\end{align*}

\paragraph{Bounds for the term $\widetilde{\zeta}_i^{(1)} (f)$:}
Given a pair of functions $f_1, f_2 \in \Fclass_\advFunc$, by Cauchy--Schwarz inequality, we have
\begin{align*}
  \Exs \big[ \big| \widetilde{\zeta}_i^{(1)} (f_1) - \widetilde{\zeta}_i^{(1)} (f_2) \big|^2 \big] &\leq c \stepsize \constFclass^2 \discount^2 \Tmax \Exs \Big[ \sum_{k = 0}^{\lceil T_i / \stepsize \rceil - 1} \big| f_1 (\State_{k \stepsize}^{(i)}, \Action_{k \stepsize}^{(i)}) - f_2 (\State_{k \stepsize}^{(i)}, \Action_{k \stepsize}^{(i)}) \big|^2  \Big]\\
  &\leq c_1 \constFclass^2 \log (n / \delta) \Statnorm{f_1 - f_2}^2.
\end{align*}
We also have the almost sure bound
\begin{align*}
  \abss{ \widetilde{\zeta}_i^{(1)} (f_1) - \widetilde{\zeta}_i^{(1)} (f_2) } \leq 2 \constFclass \discount \Tmax \vecnorm{f_1 - f_2}\infty \leq c_2 \constFclass^2 \log (n / \delta) \vecnorm{f_1 - f_2}\infty.
\end{align*}

\paragraph{Bounds for the term $\widetilde{\zeta}_i^{(2)} (f)$:} Similar to \Cref{lemma:bellman-variance-control-in-emp-proc-valfunc-proof}, we use the following lemma to control the variance and the Orlicz norm of the term $\widetilde{\zeta}_i^{(2)} (f)$.
\begin{lemma}\label{lemma:advantage-variance-control-in-emp-proc-valfunc-proof}
Under the conditions of \Cref{lemma:uniform-concentration-bellman-advantage}, for any function $f \in \Fclass_\advFunc$, we have
\begin{align*}
  \Exs \big[ |\widetilde{\zeta}_i^{(2)} (f)|^2 \big] &\leq c \log^3 (n / \delta) \statnorm{f}^2 + \frac{\vecnorm{f}{\infty}^2}{\numobs},\\
  \orlicznorm{\widetilde{\zeta}_i^{(2)} (f)} {1} &\leq c \log^2 (\numobs / \delta) \vecnorm{f}{\infty},
\end{align*}
where $c > 0$ is a constant that only depends on the problem parameters.
\end{lemma}
\noindent See \Cref{app:subsubsec-proof-lemma-advantage-variance-control-in-emp-proc-valfunc-proof} for the proof of this lemma.

Combining the bounds for the terms $\widetilde{\zeta}_i^{(1)} (f)$ and $\widetilde{\zeta}_i^{(2)} (f)$, we have verified the Bernstein-type condition. Invoking \Cref{lemma:empirical-process-tool}, and combining with the probability of the truncation event, we have with probability at least $1 - \delta$,
\begin{align*}
  \ErrTermMainAdv &\leq  c \Big\{\dudley_2 (\Fclass_\advFunc (r), \Stationary) + r \sqrt{\log (1/ \delta)} \Big\} \sqrt{\frac{\log^3 (n / \delta)}{n}} + c \Big\{\dudley_2 (\Fclass_\advFunc, \vecnorm{\cdot}{\infty}) + \constFclass \sqrt{\log (1/ \delta)} \Big\} \frac{1}{\numobs^{3/2}}  \\
  &\qquad+ c \log^2 (\numobs / \delta) \frac{\dudley_1 (\Fclass_\advFunc, \mathbb{L}^\infty) + \diameter_{\infty} (\Fclass_\advFunc) \log (1 / \delta)}{n}\\
  &\leq c \dudley_2 (\Fclass_\advFunc (r), \Stationary) \frac{\log^{3} (n / \delta)}{\sqrt{\numobs}} + c \dudley_1 (\Fclass_\advFunc, \mathbb{L}^\infty)  \frac{\log^4 (\numobs / \delta)}{n},
\end{align*}
which completes the proof.

\subsubsection{Proof of \Cref{lemma:advantage-variance-control-in-emp-proc-valfunc-proof}}\label{app:subsubsec-proof-lemma-advantage-variance-control-in-emp-proc-valfunc-proof}
The proof is similar to that of \Cref{lemma:bellman-variance-control-in-emp-proc-valfunc-proof}. By It\^{o}'s formula, we have
\begin{align*}
  \valfun (\State_{(k + 1) \stepsize}^{(i)}) - \valfun (\State_{k \stepsize}^{(i)}) &= \int_{k \stepsize}^{(k + 1) \stepsize} \generator^{\Action_{k \stepsize}^{(i)}} \valfun (\State_t^{(i)}) dt + \int_{k \stepsize}^{(k + 1) \stepsize} \nabla \valfun (\State_t^{(i)})^\top \covMat^{1/2} (\State_t^{(i)}) d \BM_t.
\end{align*}
Define the terms
\begin{align*}
  I_1 &\mydefn \sum_{k = 0}^{\lfloor T_i / \stepsize \rfloor - 1} f(\State_{k \stepsize}^{(i)}, \Action_{k \stepsize}^{(i)}) \cdot \int_{k \stepsize}^{(k + 1) \stepsize} \generator^{\Action_{k \stepsize}^{(i)}} \valfun (\State_t^{(i)}) dt ,\\
  I_2 &\mydefn \sum_{k = 0}^{\lfloor T_i / \stepsize \rfloor - 1} f(\State_{k \stepsize}^{(i)}, \Action_{k \stepsize}^{(i)}) \cdot \int_{k \stepsize}^{(k + 1) \stepsize} \nabla \valfun (\State_t^{(i)})^\top \covMat^{1/2} (\State_t^{(i)}) d \BM_t.
\end{align*}
Note that the structure of the terms $I_1$ and $I_2$ are the same as those in \Cref{lemma:bellman-variance-control-in-emp-proc-valfunc-proof}, except that the functions are now defined on the state-action space. The second moment and Orlicz norm bounds for the term $I_1$ can be established in the same way as in \Cref{lemma:bellman-variance-control-in-emp-proc-valfunc-proof}, which yield
\begin{align*}
  \Exs \big[I_1^2 \bm{1}_{T_i \leq \Tmax} \big] &\leq \big( 1 + \Tmax^2 \log n \big) \Statnorm{f}^2 + \frac{c_1}{\discount n} \vecnorm{f}{\infty}^2, \quad \mbox{and}\\
   \orlicznorm{I_1 \bm{1}_{T_i \leq \Tmax}}{1} &\leq c_2 \log^2 (n / \delta) \vecnorm{f}{\infty}.
\end{align*}
As for the term $I_2$, we note that it is also a discrete-time martingale with respect to the natural filtration generated by the Brownian motion and the randomly sampled action sequence, stopped at the random time $\lfloor T_i / \stepsize \rfloor$. Following the proof of \Cref{lemma:bellman-variance-control-in-emp-proc-valfunc-proof}, we have
\begin{align*}
  \Exs \big[ I_2^2 \bm{1}_{T_i \leq \Tmax} \big] &\leq \frac{\constFclass^2 \lammax}{\discount} \Statnorm{f}^2, \quad \mbox{and}\\
  \orlicznorm{I_2 \bm{1}_{T_i \leq \Tmax}}{1} &\leq c_3 \sqrt{\log (n / \delta)}\vecnorm{f}{\infty}.
\end{align*}
Combining the bounds for the terms $I_1$ and $I_2$ yields the desired results.

\end{document}